\pdfoutput=1
\documentclass[runningheads]{llncs}

\usepackage{eccv}

\usepackage{eccvabbrv}

\usepackage{graphicx}
\usepackage{booktabs}
\usepackage{array}

\usepackage[accsupp]{axessibility}
\usepackage{hyperref}

\usepackage{orcidlink}

\newcommand{\ours}{\textsc{StyleFields}\xspace}

\begin{document}

\title{StyleFields: Multi-Scale AdaIN-Modulated Implicit SDFs for Coarse-to-Fine 3D Shape Reconstruction and Editing}

\author{
Ehsan Garaaghaji\inst{1} \and
Nicolas Talabot\inst{2} \and
Pascal Fua\inst{2} \and
Doruk Oner\inst{1}
}

\authorrunning{E.~Garaaghaji et al.}
\titlerunning{StyleFields}

\institute{
Bilkent University, Ankara, Türkiye
\and
École Polytechnique Fédérale de Lausanne (EPFL), Lausanne, Switzerland
}

\maketitle

\begin{abstract}
We introduce \ours, a DeepSDF-based architecture for high-fidelity 3D reconstruction that enables controllable geometric style mixing: the coarse structure of one object can be combined with the fine-scale details of another. The core idea is depth-aware modulation: instead of a single global code, we inject latents via multi-level Adaptive Instance Normalization at several decoder depths, and supervise matching auxiliary heads with a coarse-to-fine schedule while gradually growing network depth. This aligns early layers with global shape and later layers with high-frequency detail, achieving content–style decoupling without part labels or adversarial training. \ours delivers faithful reconstructions, convincing cross-instance hybrids, and consistent gains in ablations over injection depth and supervision granularity. We further demonstrate a practical application in automotive aerodynamics: a learned surrogate drag predictor serves as a differentiable objective to optimize reconstructed cars, allowing targeted edits of global form or surface details by freezing the complementary latent stream. \ours offers a simple, effective recipe for controllable implicit reconstruction and downstream performance-driven design.

\end{abstract}

\section{Introduction}
\label{sec:introduction}

\begin{figure*}[t]
    \centering
    \includegraphics[width=0.99\linewidth]{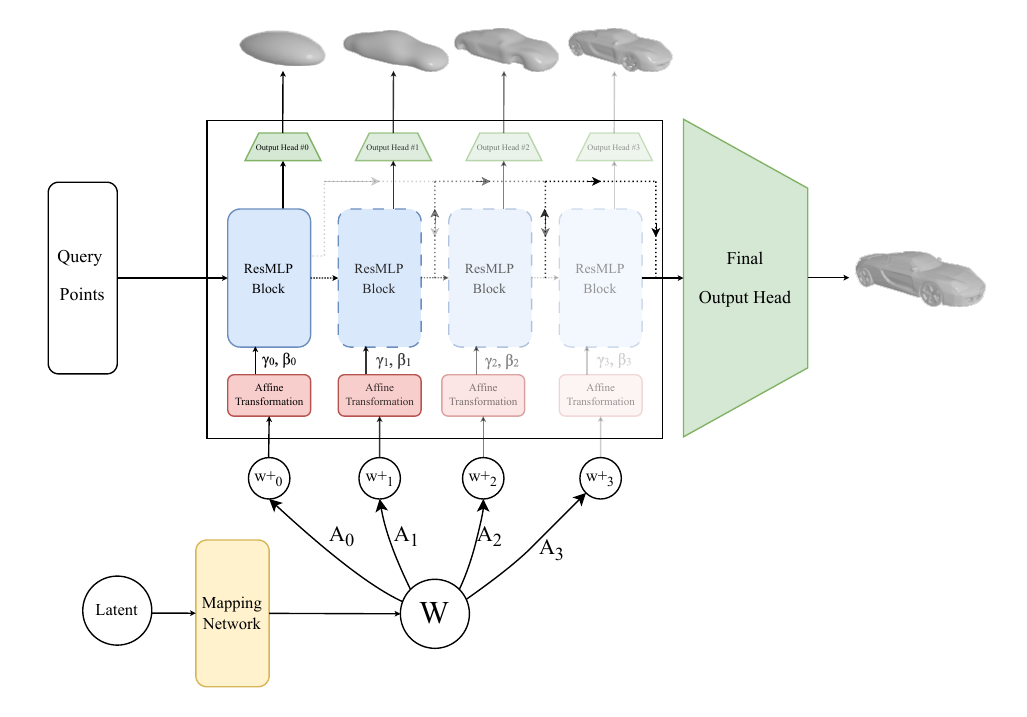}
    \caption{\ours architecture with depth-aware AdaIN and progressive decoder growth. Fading colors indicate the curriculum on depth. Earlier \textsc{ResMLP} blocks model coarse structure, while later (faded) ones are gradually activated for fine detail.}
    \label{fig:architecture}
\end{figure*}

High-fidelity 3D reconstruction and controllable shape manipulation are key to applications in robotics, AR/VR content and digital twin creation, cultural-heritage capture, medical modeling, autonomous driving, and industrial design. Neural implicit signed distance fields provide a powerful foundation for this by  representing geometry continuously with sub-voxel precision without topological constraints and being easy to integrate with gradient-based optimization pipelines. However, existing implicit representations offer only limited control over different levels of geometric detail—a limitation we address in this work.

Neural implicit methods such as DeepSDF~\cite{park2019deepsdf,duan2020curriculum} achieve high-fidelity instance reconstruction by conditioning a decoder on a single latent code. However, this global code entangles coarse structure and fine geometric detail, so modifying one inevitably affects the other. Part-based formulations offer more localized control by assigning separate latents to semantic components~\cite{talabot2025partsdf}, but they require part annotations and often introduce discontinuities at boundaries. StyleGAN-derived 3D generators~\cite{chan2022efficient,gao2022get3d} provide rich latent manipulations, yet they model distributions rather than target instances and therefore struggle with precise recovery and fine-grained, instance-specific control. 
More recent neural-field representations such as 3DShape2VecSet and Dora~\cite{zhang20233dshape2vecset,chen2025dora} replace the single global latent of DeepSDF by vector sets. However, while powerful, these sets are not explicitly organized by \emph{geometric scale}, making operations such as “freeze coarse, edit fine” or predictable cross-instance coarse--fine exchange ill-defined.

In this paper, we introduce \ours, a neural implicit representation that preserves high-fidelity reconstruction while enabling independent control over coarse and fine geometry. Instead of a single global latent, \ours{} uses multiple style codes, each modulating a different depth range of the decoder through a StyleGAN-inspired latent injection mechanism. A combination of depth-aware modulation, hierarchical multi-scale supervision, and a curriculum over decoder depth and target smoothness encourages early layers to capture global structure while deeper layers specialize in fine detail. As a result, different style codes specialize to different levels of geometric detail, yielding a controllable, granularity-aware implicit representation.

We demonstrate that \ours delivers more precise control than earlier methods for two different purposes.
\begin{enumerate}

 \item {\bf Shape Manipulation.} \ours{} enables controlled editing across geometric scales: users can refine surface detail while preserving global structure, alter global shape while keeping fine features intact, or transfer coarse attributes from one instance to another without disrupting local detail. Unlike single-latent implicit models, which only support global interpolation, \ours provides explicit, multi-level control and cross-instance style mixing by combining style codes across depth ranges. This is achieved with a single, lightweight network without the need for adversarial training or part annotations.

 \item  {\bf Design Optimization.} \ours also supports differentiable design optimization by enabling scale-selective latent updates. After training on car shapes, we couple the implicit SDF to a differentiable surrogate that predicts aerodynamic drag and optimize the shape through gradient descent. By freezing either the coarse-level or fine-level style codes, we can target global form or surface detail independently, preserving design intent while improving aerodynamic performance. This level of selective control is difficult to achieve with single-latent implicit models, which tie all geometric scales to one vector.

\end{enumerate}

In short, our contributions are as follows. 
\begin{enumerate}
    \item A multi-level, granularity-aware implicit representation that organizes
    geometry across coarse and fine scales and enables structured manipulation at
    each level, including cross-instance style mixing where different scales of
    geometry can be selectively exchanged between shapes.
    \item A curriculum-guided, coarse-to-fine training strategy combining depth-aware modulation and hierarchical multi-scale supervision, which induces a consistent coarse–fine organization across decoder layers.
    \item A differentiable design-optimization pipeline that leverages scale-selective control to refine shape attributes while preserving high-level design intent, demonstrated on aerodynamic drag reduction.

\end{enumerate}

We evaluate \ours on standard benchmarks of cars, chairs, and airplanes, demonstrating state-of-the-art reconstruction accuracy and controllable coarse-to-fine shape editing beyond what previous INR models support.

\section{Related Work}
\label{sec:relatedwork}

We briefly review some of the main current approaches to implicit surface modeling and to providing control over level of detail. We argue that they are not designed for automated and controlled editing across geometric scales, which can also be used for design-optimization, the way \ours{} is.

\paragraph{Global-Latent INR Models.}
The original Implicit Neural Representations (INRs) model 3D shapes in terms of continuous functions over $\mathbb{R}^3$ that implicitly encode object surfaces~\cite{park2019deepsdf,mescheder2019occupancy,chen2019learning,xu2019disn}. More recent methods improve reconstruction fidelity and scalability through richer parameterizations, including periodic or multi-resolution coordinate networks \cite{sitzmann2020implicit,takikawa2021neural}, dense or hybrid feature volumes \cite{yan2022shapeformer,mittal2022autosdf}, irregular grids \cite{zhang20223dilg}, and unordered latent sets \cite{zhang20233dshape2vecset}. However, when applied to instance-conditioned reconstruction, all these approaches still represent each shape using a single global latent code, causing coarse structure and fine detail to be entangled in one vector and preventing scale-specific manipulation.

Among these, DeepSDF-style~\cite{park2019deepsdf} decoders use a single-resolution MLP conditioned on this global latent, offering no explicit factorization across geometric level of details or network depths. This remains true of more recent and more sophisticated methods, such as~\cite{zhang20233dshape2vecset}. Curriculum and progressive growing strategies \cite{duan2020curriculum} encourage coarse-to-fine learning during training but do not change the underlying representation: a single latent code continues to control all scales simultaneously. As a result, most INR-based reconstruction methods inherit this coupling of coarse and fine geometry and provide no mechanism for disentangled, multi-level control, the limitation we target.

\paragraph{Part-Aware Representation.}
To address the limited controllability of global-latent INRs, several works introduce part-aware decompositions. Representations that assume a known part structure, such as PQ-Net~\cite{wu20pqnet}, ProGRIP~\cite{deng22progrip}, PASTA~\cite{li2024pasta} and PartSDF~\cite{talabot2025partsdf}, associate a separate latent code to each component or rely on the global part structure, enabling explicit part-wise editing but requiring part annotations and additional constraints to prevent seams or inconsistencies at part boundaries. Complementarily, some unsupervised approaches~\cite{chen2019baenet,chen2023daenet} perform part-based co-segmentation to infer decompositions directly from data, therefore reducing manual labeling. Other methods, such as RIM-Net~\cite{niu22rimnet}, SPAGHETTI~\cite{hertz2022spaghetti}, and PartNeRF~\cite{tertikas23partnerf}, learn and leverage this decomposition to create part hierarchy or to add controllability to the shape. However, the discovered parts do not necessarily align with semantic components, limiting interpretability and precise control. Very recent diffusion-based methods push part-aware modeling further by directly \emph{synthesizing} structured components rather than post-hoc decomposing a fused shape: PartCrafter~\cite{lin2025partcrafter} jointly denoises multiple mesh parts from a single image via a compositional latent space and cross-part attention; Dual Volume Packing~\cite{tang2025efficient} generates an arbitrary number of complete parts by packing them into complementary volumes for end-to-end assembly; and OmniPart~\cite{yang2025omnipart} decouples controllable structure planning (a variable-length set of 3D part boxes guided by 2D part masks) from spatially conditioned part synthesis adapted from a holistic 3D generator. Relatedly, ARTDECO~\cite{li2025artdeco} emphasizes \emph{structured} representations for efficient high-fidelity reconstruction by coupling foundation-model priors with a hierarchical 3D Gaussian scene representation for monocular video. In contrast to such explicit part generators, our method retains a single continuous implicit field and attains fine-grained control by distributing latent information across multiple layer-wise modulations, enabling smooth localized edits without requiring part masks or part-level supervision, while remaining complementary to part-structured diffusion pipelines.

\paragraph{Shape Detailing.}
A separate line of methods focuses on detailing shapes, that is, upsampling coarse voxelized inputs into high-resolution geometry conditioned on a style example~\cite{chen2021decor,chen2023shaddr,chen2024decollage}. Using conditional generators and patch-level discriminators, these approaches can synthesize rich surface detail and offer limited bi-level control by modifying either the coarse input or the style exemplar. However, their style spaces are tied to the GAN’s training distribution, restricting generalization to unseen styles; their adversarial formulation introduces training instability and high computational cost; and their control is fundamentally limited to a coarse vs.\ fine hierarchy, without exposing intermediate levels of geometric detail.

\paragraph{Style-Based Generation.}
Style-based 3D generators such as SDF-StyleGAN~\cite{zheng2022sdf} and StyleSDF~\cite{or2022stylesdf} synthesize shapes by decoding multi-resolution voxel or hybrid feature volumes modulated by per-resolution latents. This architecture provides coarse-to-fine style control within the generated volume, and latent inversion enables reconstruction and completion. Related adversarial 3D generators such as EG3D~\cite{chan2022efficient} and GET3D~\cite{gao2022get3d} offer similar style manipulation capabilities in alternative 3D parameterizations. While these models achieve strong visual realism and editable latent spaces, they are optimized primarily for unconditional generation: inversion remains approximate, instance-level fidelity is limited, and the reliance on large voxelized feature grids and adversarial training introduces substantial computational overhead and stability challenges. As a result, these approaches serve as powerful generative priors but are not well suited as lightweight, reconstruction-centric implicit representations with precise, instance-conditioned geometric control.

\section{Methodology}
\label{sec:methodology}

\begin{figure}[t]
	\centering
	
	\newcommand{\carA}{0}
	\newcommand{\carB}{4}
	\newcommand{\chairA}{5}
	\newcommand{\chairB}{44}
	\newcommand{\planeA}{5}
	\newcommand{\planeB}{3}
	
	\newlength{\qualwidth}
	\setlength{\qualwidth}{0.118\linewidth}

	\setlength{\tabcolsep}{1pt}
	\renewcommand{\arraystretch}{0.92}
	
	\newlength{\rowlabw}
	\setlength{\rowlabw}{0.20\linewidth}
	
	\scriptsize

    \newcommand{\CH}[1]{\parbox[c]{\qualwidth}{\centering\tiny\textsc{#1}}}
    \newcommand{\CHtwo}[2]{\parbox[c]{\qualwidth}{\centering\tiny\textsc{#1}\\\textsc{#2}}}
    
	\begin{tabular}{@{}
		>{\centering\arraybackslash}m{\rowlabw}
		m{\qualwidth} m{\qualwidth} m{\qualwidth} m{\qualwidth} m{\qualwidth} m{\qualwidth}
	@{}}
		& \CH{GT}
        & \CH{Base}
        & \CH{Curr.}
        & \CHtwo{Curr.}{+Style}
        & \CH{Decoll.}
        & \CH{Ours} \\
		\cmidrule(lr){2-2}\cmidrule(lr){3-3}\cmidrule(lr){4-4}\cmidrule(lr){5-5}\cmidrule(lr){6-6}\cmidrule(lr){7-7}
		
		\textsc{Cars}~S1
		& \includegraphics[width=\qualwidth]{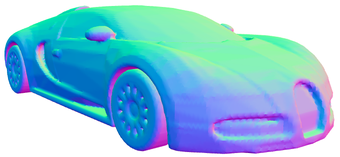}
		& \includegraphics[width=\qualwidth]{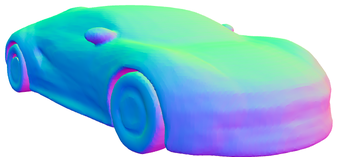}
		& \includegraphics[width=\qualwidth]{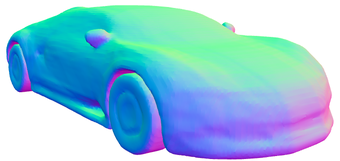}
		& \includegraphics[width=\qualwidth]{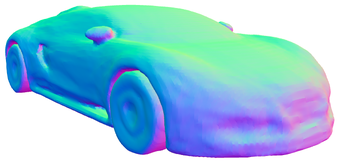}
		& \includegraphics[width=\qualwidth]{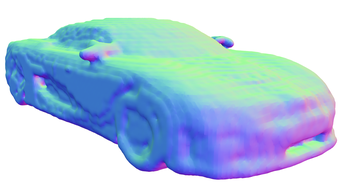}
		& \includegraphics[width=\qualwidth]{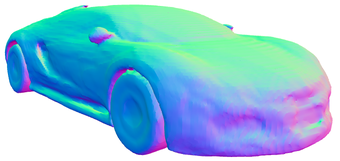} \\
		
		\textsc{Cars}~S2
		& \includegraphics[width=\qualwidth]{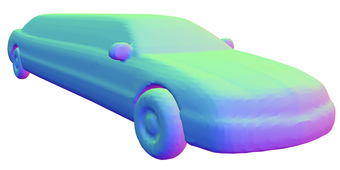}
		& \includegraphics[width=\qualwidth]{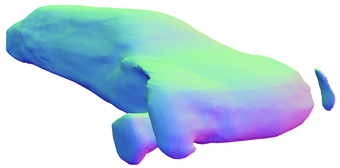}
		& \includegraphics[width=\qualwidth]{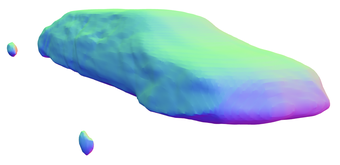}
		& \includegraphics[width=\qualwidth]{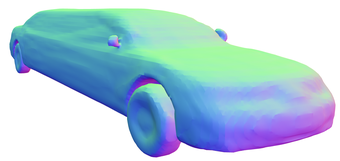}
		& \includegraphics[width=\qualwidth]{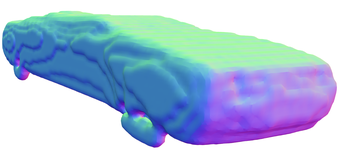}
		& \includegraphics[width=\qualwidth]{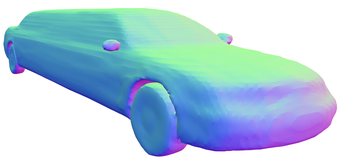} \\
		
		\midrule
		
        \textsc{Chairs}~S1
        & \makebox[\qualwidth][c]{\includegraphics[width=0.6\qualwidth]{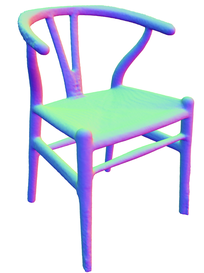}}
        & \makebox[\qualwidth][c]{\includegraphics[width=0.6\qualwidth]{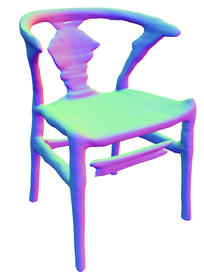}}
        & \makebox[\qualwidth][c]{\includegraphics[width=0.6\qualwidth]{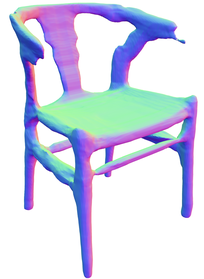}}
        & \makebox[\qualwidth][c]{\includegraphics[width=0.6\qualwidth]{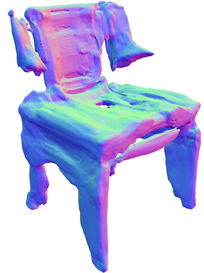}}
        & \makebox[\qualwidth][c]{\includegraphics[width=0.6\qualwidth]{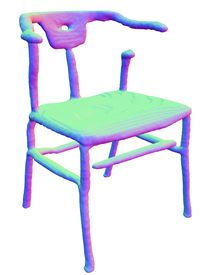}}
        & \makebox[\qualwidth][c]{\includegraphics[width=0.6\qualwidth]{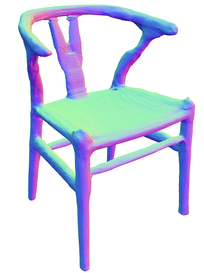}} \\
        
        \textsc{Chairs}~S2
        & \makebox[\qualwidth][c]{\includegraphics[width=0.45\qualwidth]{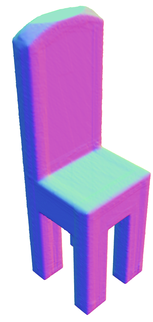}}
        & \makebox[\qualwidth][c]{\includegraphics[width=0.45\qualwidth]{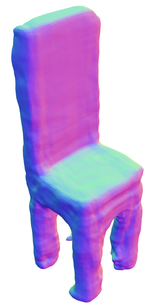}}
        & \makebox[\qualwidth][c]{\includegraphics[width=0.45\qualwidth]{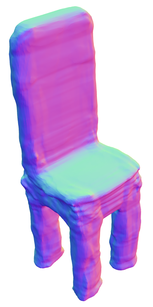}}
        & \makebox[\qualwidth][c]{\includegraphics[width=0.45\qualwidth]{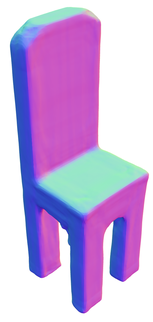}}
        & \makebox[\qualwidth][c]{\includegraphics[width=0.45\qualwidth]{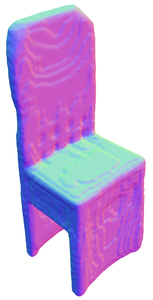}}
        & \makebox[\qualwidth][c]{\includegraphics[width=0.45\qualwidth]{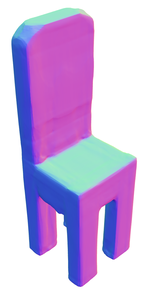}} \\
        		
		\midrule
		
		\textsc{Planes}~S1
		& \makebox[\qualwidth][c]{\includegraphics[width=1.0\qualwidth]{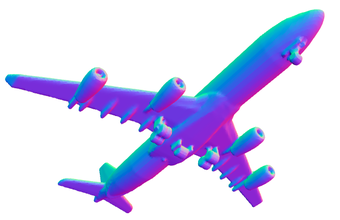}}
        & \makebox[\qualwidth][c]{\includegraphics[width=1.0\qualwidth]{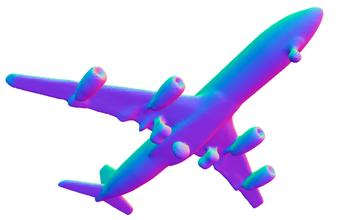}}
        & \makebox[\qualwidth][c]{\includegraphics[width=1.0\qualwidth]{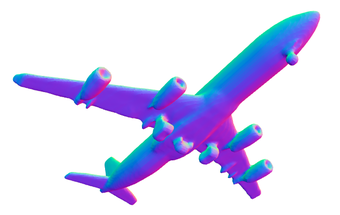}}
        & \makebox[\qualwidth][c]{\includegraphics[width=1.0\qualwidth]{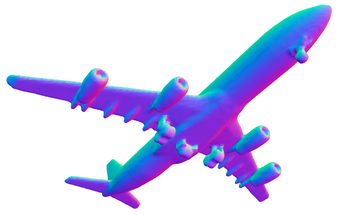}}
        & \makebox[\qualwidth][c]{\includegraphics[width=1.0\qualwidth]{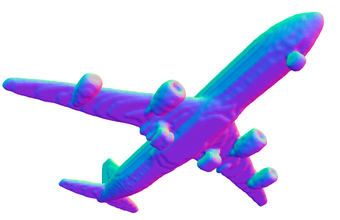}}
        & \makebox[\qualwidth][c]{\includegraphics[width=1.0\qualwidth]{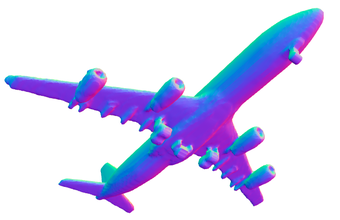}} \\
		
		\textsc{Planes}~S2
		& \makebox[\qualwidth][c]{\includegraphics[width=1.0\qualwidth]{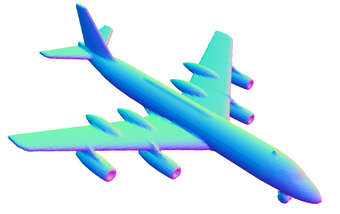}}
        & \makebox[\qualwidth][c]{\includegraphics[width=1.0\qualwidth]{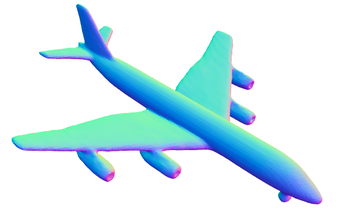}}
        & \makebox[\qualwidth][c]{\includegraphics[width=1.0\qualwidth]{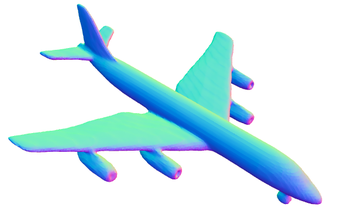}}
        & \makebox[\qualwidth][c]{\includegraphics[width=1.0\qualwidth]{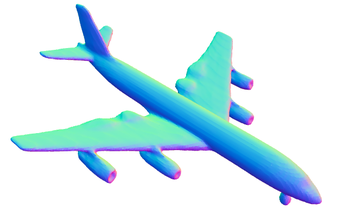}}
        & \makebox[\qualwidth][c]{\includegraphics[width=1.0\qualwidth]{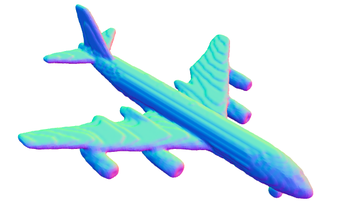}}
        & \makebox[\qualwidth][c]{\includegraphics[width=1.0\qualwidth]{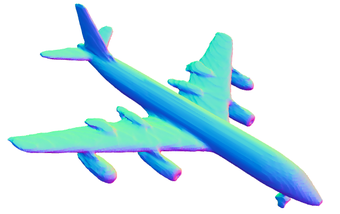}} \\
	\end{tabular}
	
	\caption{
		Qualitative comparison of 3D reconstruction on \textsc{Cars}, \textsc{Chairs}, and \textsc{Planes}.
		Rows show two shape instances per category; columns show methods.
	}
	\label{fig:qualitative_grid}
\end{figure}

\begin{figure}[t]
    \centering

    \setlength{\tabcolsep}{4pt}
    \renewcommand{\arraystretch}{1.05}

    \resizebox{0.8\textwidth}{!}{%
    \begin{tabular}{ccccccc}
        \toprule
        Shape 1 (GT)
        & Shape 1 (Reconstruction)
        & 0.25 Interpolation
        & 0.5 Interpolation
        & 0.75 Interpolation
        & Shape 2 (Reconstruction)
        & Shape 2 (GT) \\
        \midrule

        \multicolumn{7}{@{}l}{\textbf{DeepSDF}} \\
        \includegraphics[width=0.23\linewidth]{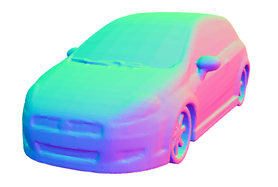} &
        \includegraphics[width=0.23\linewidth]{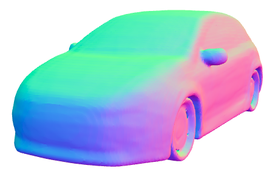} &
        \includegraphics[width=0.23\linewidth]{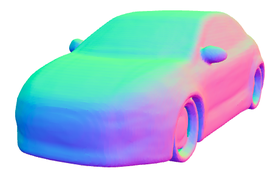} &
        \includegraphics[width=0.23\linewidth]{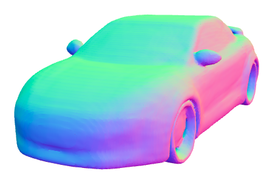} &
        \includegraphics[width=0.23\linewidth]{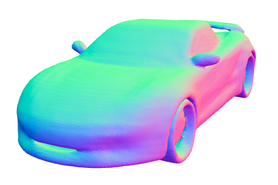} &
        \includegraphics[width=0.23\linewidth]{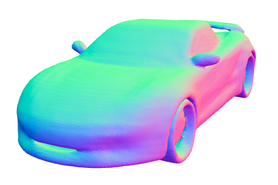} &
        \includegraphics[width=0.23\linewidth]{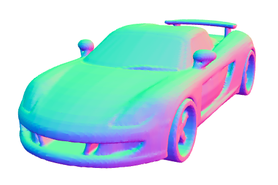} \\
        \midrule

        \multicolumn{7}{@{}l}{\textbf{Curriculum DeepSDF}} \\
        \includegraphics[width=0.23\linewidth]{figures/interpolation/gt_0.png} &
        \includegraphics[width=0.23\linewidth]{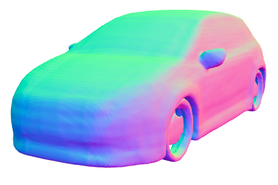} &
        \includegraphics[width=0.23\linewidth]{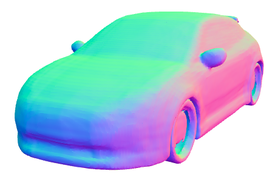} &
        \includegraphics[width=0.23\linewidth]{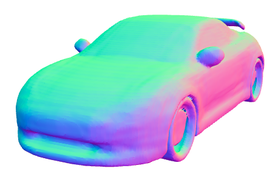} &
        \includegraphics[width=0.23\linewidth]{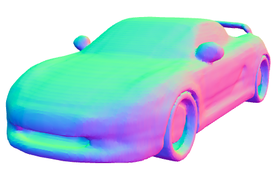} &
        \includegraphics[width=0.23\linewidth]{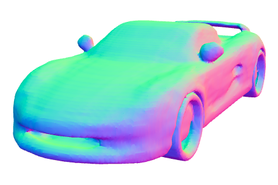} &
        \includegraphics[width=0.23\linewidth]{figures/interpolation/gt_100} \\
        \midrule

        \multicolumn{7}{@{}l}{\textbf{Curriculum Style DeepSDF}} \\
        \includegraphics[width=0.23\linewidth]{figures/interpolation/gt_0.png} &
        \includegraphics[width=0.23\linewidth]{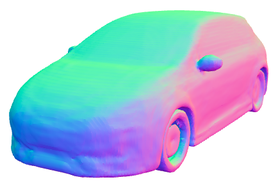} &
        \includegraphics[width=0.23\linewidth]{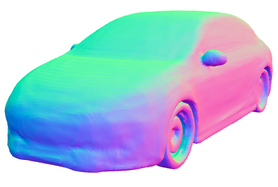} &
        \includegraphics[width=0.23\linewidth]{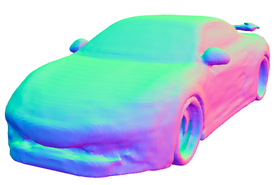} &
        \includegraphics[width=0.23\linewidth]{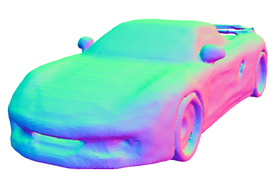} &
        \includegraphics[width=0.23\linewidth]{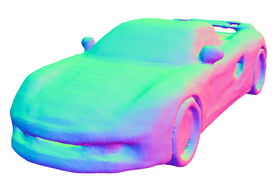} &
        \includegraphics[width=0.23\linewidth]{figures/interpolation/gt_100} \\
        \midrule

        \multicolumn{5}{@{}l}{\textbf{\ours}} \\
        \includegraphics[width=0.23\linewidth]{figures/interpolation/gt_0.png} &
        \includegraphics[width=0.23\linewidth]{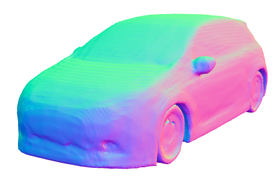} &
        \includegraphics[width=0.23\linewidth]{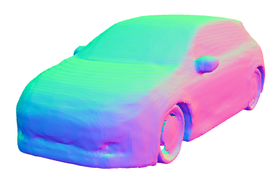} &
        \includegraphics[width=0.23\linewidth]{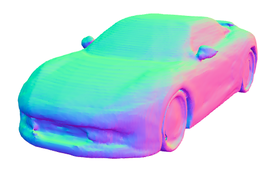} &
        \includegraphics[width=0.23\linewidth]{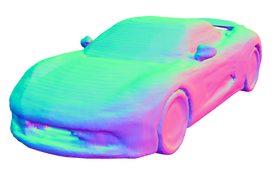} &
        \includegraphics[width=0.23\linewidth]{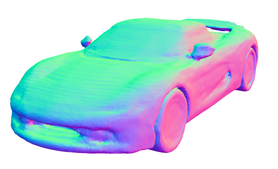} &
        \includegraphics[width=0.23\linewidth]{figures/interpolation/gt_100} \\
        \bottomrule
    \end{tabular}%
    }%

    \caption{
        Global latent interpolation between two shapes across different 3D reconstruction baselines and \ours.
    }
    \label{fig:interpolation_grid}
\end{figure}

\begin{figure}[t]
    \centering
    \setlength{\tabcolsep}{2pt}
    \renewcommand{\arraystretch}{1.0}

    \resizebox{0.8\linewidth}{!}{%
    \begin{tabular}{ccccc}
        {\scriptsize Shape 1} &
        {\scriptsize A} &
        {\scriptsize B} &
        {\scriptsize C} &
        {\scriptsize Shape 2} \\[4pt]
        \includegraphics[width=0.18\linewidth]{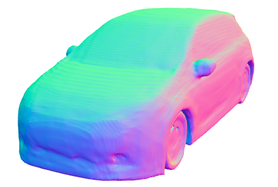} &
        \includegraphics[width=0.18\linewidth]{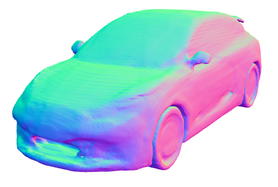} &
        \includegraphics[width=0.18\linewidth]{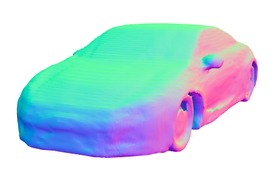} &
        \includegraphics[width=0.18\linewidth]{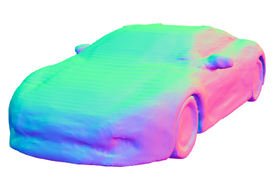} &
        \includegraphics[width=0.18\linewidth]{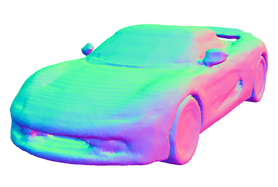} \\

        \includegraphics[width=0.18\linewidth]{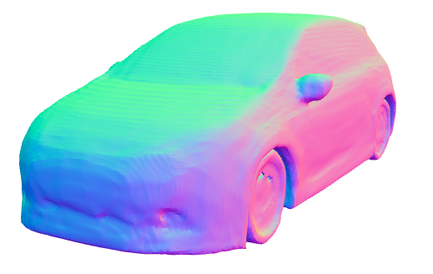} &
        \includegraphics[width=0.18\linewidth]{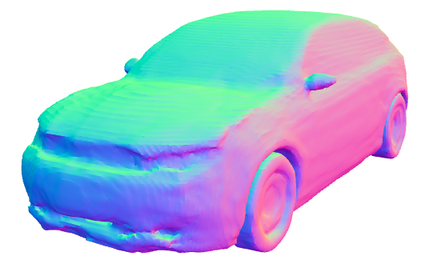} &
        \includegraphics[width=0.18\linewidth]{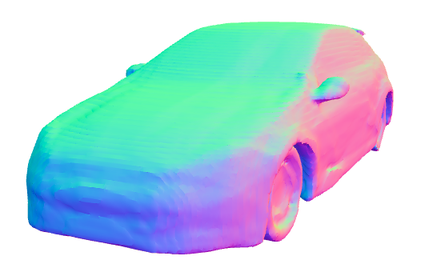} &
        \includegraphics[width=0.18\linewidth]{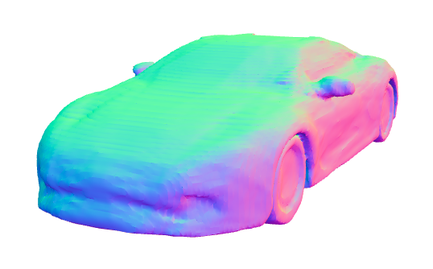} &
        \includegraphics[width=0.18\linewidth]{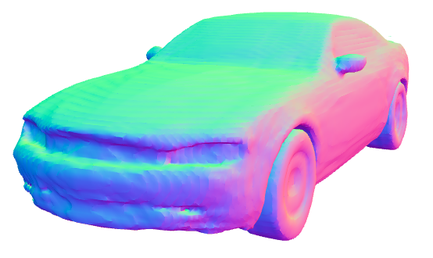} \\

    \end{tabular}%
    }

    \caption{
        StyleFields cross-instance style mixing between two car shapes (source on the left, target on the right). Each column shows a hybrid produced by selectively substituting style codes of Shape~1 with codes from Shape~2. 
    (A) Coarse styles $(w^+_0, w^+_1)$ from Shape~1 and fine styles $(w^+_2, w^+_3)$ from Shape~2. 
    (B) Mid-level styles $(w^+_1, w^+_2)$ from Shape~1 and coarsest and finest styles $(w^+_0, w^+_3)$ from Shape~2. 
    (C) Fine styles $(w^+_2, w^+_3)$ from Shape~1 and coarse styles $(w^+_0, w^+_1)$ from Shape~2.
    }
    \label{fig:style_mixing}
\end{figure}

\begin{figure}[t]
    \centering
    \setlength{\tabcolsep}{1pt}
    \renewcommand{\arraystretch}{1.0}

    \newlength{\imgw}
    \setlength{\imgw}{\dimexpr(\linewidth-6\tabcolsep)/4\relax}

    \begin{tabular}{@{}cccc@{}}
        \includegraphics[width=\imgw]{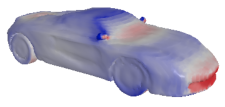} &
        \includegraphics[width=\imgw]{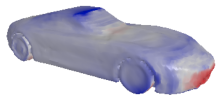} &
        \includegraphics[width=\imgw]{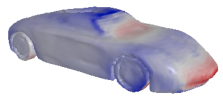} &
        \includegraphics[width=\imgw]{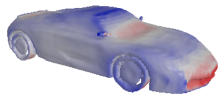} \\

        \includegraphics[width=\imgw]{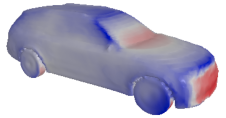} &
        \includegraphics[width=\imgw]{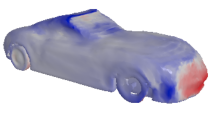} &
        \includegraphics[width=\imgw]{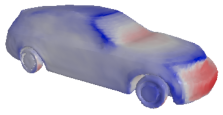} &
        \includegraphics[width=\imgw]{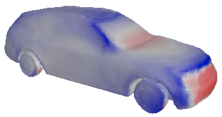} \\

        \parbox[t]{\imgw}{\centering\scriptsize (Initial\\reconstruction)} &
        \parbox[t]{\imgw}{\centering\scriptsize (Optimize\\latent $z$)} &
        \parbox[t]{\imgw}{\centering\scriptsize (Optimize\\coarse style)} &
        \parbox[t]{\imgw}{\centering\scriptsize (Optimize\\fine style)}
    \end{tabular}

    \caption{
        \textbf{Drag Optimization.}
        Optimizing the global latent $z$ induces global shape drift and often converges to similar solutions across different initial shapes.
        In contrast, optimizing per-scale style codes $w^{+}$ enables targeted edits at specific geometric scales (coarse vs.\ fine).
        Surface color represents the predicted air pressure (blue to red).
    }
    \label{fig:drag_optimization}
\end{figure}

Given a training set of shapes represented by signed distance fields (SDFs) $\{S_i : \mathbb{R}^3 \to \mathbb{R}\}$,
our goal is to learn an instance-conditioned implicit decoder
\begin{align}
    f_\theta(\mathbf{x}, \mathbf{z}_i) \approx S_i(\mathbf{x}),
    \label{eq:decoder_objective}
\end{align}
where $\mathbf{x} \in \mathbb{R}^3$ is a query point and $\mathbf{z}_i\in \mathbb{R}^{d_z}$ is a latent code associated with shape $i$, which is learned in an auto-decoding fashion~\cite{park2019deepsdf}.
Standard implicit models condition the decoder on a \emph{single} global latent vector, forcing all geometric scales from coarse global structure to fine surface detail, to be represented in one entangled space. Because the network receives no architectural or training signal encouraging depth-wise specialization, layers depth do not reliably encode increasingly higher-frequency geometry, making structured, scale-specific manipulation difficult. Even in more recent models based on the  \emph{VecSet} latent space~\cite{zhang20233dshape2vecset,chen2025dora}, these levels of details are not explicitly disentangled within the latent vectors.

\ours replaces the single global latent vector by an \emph{ordered set} of style codes $\{w_j^+\in \mathbb{R}^{d_w}\}$ that are injected at different depths of the decoder through StyleGAN-inspired modulation layers. Each code modulates a distinct set of layers, allowing different geometric scales to be adjusted independently.

To ensure that these codes specialize meaningfully, we train the model using a coarse-to-fine strategy combining two mechanisms:
(i) a curriculum over decoder depth, where deeper layers are progressively
introduced during training, and (ii) a hierarchical supervision using multiple
smoothed versions of the target SDF. Specifically, we precompute Gaussian-smoothed
targets $\{S_{\sigma_k}\}$ with increasing standard-deviations and supervise early layers with the heavily smoothed targets and deeper layers with the lightly smoothed ones. Together, these mechanisms
induce a consistent coarse–to-fine organization across the network so that early
style codes learn global shape structure and deeper codes capture high-frequency
details.

The resulting representation supports level-specific editing and cross-instance style mixing, where coarse geometry from one shape can be combined with fine detail from another—operations that are difficult or impossible with traditional single-latent INRs. \ours therefore provides high-fidelity reconstruction together with explicit, controllable organization across geometric granularity.

In the remainder of this section, we present the architecture of our decoder, our training strategy, and our approach to combining styles from different objects.

\subsection{Decoder Architecture}

Our goal is to build an implicit decoder that can represent a shape with multiple levels of geometric detail.
Rather than conditioning the entire network on a global latent representation, we divide the decoder into a small number of
depth ranges and allow each range to be modulated by its own latent \emph{style code}.
This section describes how these style codes are constructed and how they influence the decoder.

\paragraph{Base MLP structure.}
The decoder $f_\theta$ is composed of $L$ residual blocks, each a small MLP with 2 layers.
Given an input point $\mathbf{x} \in \mathbb{R}^3$ and a collection of style codes $\{w_j^+\}_j$, the decoder outputs a signed distance value,
\[
f_\theta(\mathbf{x}; \{w_j^+\}_j) : \mathbb{R}^3 \rightarrow \mathbb{R}.
\]
Residual blocks allow the network to grow deeper while remaining stable and easy to optimize.

\paragraph{From a global latent to multiple style codes.}
Each training shape has a global latent vector $z \in \mathbb{R}^{d_z}$ that summarizes instance-specific geometry.
Following style-based generative models, we do not use $z$ directly.
Instead, a mapping network $g_\phi$ transforms it into an intermediate latent,
\[
w = g_\phi(z) \in \mathbb{R}^{d_w},
\]
which is then projected into a \emph{small set} of style codes, with linear projection $A_j$, 
\[
w_j^+ = A_j w, \qquad j \in \mathcal{J}.
\]
Here, $\mathcal{J}$ is a set of depth indices (\eg, four indices), each corresponding to one “granularity level.”
This construction replaces the traditional single global code with multiple codes that will influence different parts of the decoder.

\paragraph{How style codes modulate the decoder.} 

If $j \in \mathcal{J}$, meaning that the block is modulated, we compute scale--shift parameters from the style code,
\[
(\gamma_j, \beta_j) = \mathrm{Aff}_j(w_j^+),
\]
where $\mathrm{Aff}_j$ is a learned affine transformation.
We then apply Adaptive Instance Normalization (AdaIN) to modulate the features at every layer in the $j$-th residual block. Denoting $h_j$ the feature tensor produced at such a layer, we write:

\begin{align}
\mathrm{AdaIN}(h_j;\gamma_j,\beta_j)
=
\gamma_j \cdot
\frac{h_j - \mu(h_j)}{\sigma(h_j) + \varepsilon}
+
\beta_j,
\label{eq:adain}
\end{align}
where $\mu(h_j)$ and $\sigma(h_j)$ are the per-feature mean and standard deviation. Every layer in block $j$ is modulated this way.
Blocks that are not in $\mathcal{J}$ operate normally.

Intuitively, AdaIN allows each style code to reshape the feature distribution of the decoder blocks it modulates.
To expose multiple levels of control, the indices in $\mathcal{J}$ are arranged from shallow to deep, forming a sequence of coarse-to-fine modulation groups, and all residual blocks between two consecutive indices share the same style code.
This provides the decoder with a small number of “control knobs,” each influencing a different band of geometric detail—from broad, low-frequency structure at shallow depths to progressively finer detail at deeper depths. Importantly, this division of responsibilities does \emph{not} arise from the architecture alone: it is enforced by the training strategy described next, which uses curriculum-based depth growth and multi-scale supervision to consistently steer early groups toward coarse geometry and later groups toward high-frequency surface detail.

\subsection{Coarse-to-Fine Training}

During training, for each training SDF $S$, we sample points $\mathbf{x}$ and supervise the
auxiliary heads $o_k$ and the final output head. Each auxiliary head predicts
$\hat{s}_k(\mathbf{x})$ and is matched to a smoothed target $S_{\sigma_k}$, while
the final head predicts $\hat{s}_{\mathrm{fin}}(\mathbf{x})$ and is trained on the
original SDF.
We minimize the $\ell_1$ regression loss
\begin{align}
\mathcal{L}_{\text{sdf}} =
&\mathbb{E}_{\mathbf{x}} \Big[
    \sum_{k=0}^{3}
        \lambda_k \, |\hat{s}_k(\mathbf{x}) - S_{\sigma_k}(\mathbf{x})|
    +\, \lambda_{\mathrm{fin}} \,
        |\hat{s}_{\mathrm{fin}}(\mathbf{x}) - S(\mathbf{x})| \Big] \; .
\label{eq:loss}
\end{align}
Following standard practice~\cite{park2019deepsdf}, we clip the SDF values to $[-\tau,\tau]$ to focus learning near the surface.
To ensure that the multiple modulation groups present in our architecture specialize in the desired manner when we minimize $\mathcal{L}_{\text{sdf}}$, we introduce two complementary mechanisms 
\begin{enumerate}
 \vspace{-2mm}
 \item a curriculum that gradually increases active network depth~\cite{duan2020curriculum}, 
 \item a multi-scale supervision using smoothed SDF targets.
 \vspace{-2mm}
\end{enumerate}
Together, they create a consistent progression from coarse global shape to refined geometry across decoder depth. We discuss them below.

\paragraph{Curriculum over Decoder Depth.}
The final decoder contains $L$ residual blocks, but training begins with only the
first $L^{(0)}$ block active. Deeper blocks are introduced progressively,
\[
L^{(0)} \prec L^{(1)} \prec \cdots \prec L^{(T)} = L,
\]
with newly added blocks warmed up using a smaller learning rate for stability.
This staged growth allows shallow blocks to capture stable global structure
before deeper ones begin refining geometry.

\paragraph{Multi-scale Supervision with Auxiliary Output Heads.}
To couple this progression with the training signal, we attach lightweight
prediction heads at selected depths. For each SDF $S$, we precompute smoothed
targets
\[
S_{\sigma_0},\; S_{\sigma_1},\; S_{\sigma_2},\; S_{\sigma_3},
\qquad
\mathrm{with} \;\; \sigma_0 > \sigma_1 > \sigma_2 > \sigma_3 > 0,
\]
via 3D Gaussian convolutions. Shallow heads are supervised with the most smoothed
targets, while deeper heads receive increasingly sharp versions, with the final
decoder output trained on the original SDF. Importantly, we continue to supervise output heads even after the following blocks are introduced.

This supervision differs from the tolerance scheduling introduced in Curriculum
DeepSDF~\cite{duan2020curriculum}, where a gradually decreasing tolerance reduces
early training difficulty. Instead, we provide explicit smoothed SDF fields at
multiple scales and associate them with specific depths, offering direct coarse-to-fine
training signals.

\paragraph{Joint effect.}
Depth growth defines an ordered progression of capacity, while multi-scale
supervision aligns this capacity with increasingly detailed targets. Together,
they establish a reliable coarse-to-fine organization across the decoder, so the
style codes modulating different depth ranges naturally acquire distinct and
predictable roles.

\subsection{Style Mixing and Level-Selective Editing}

The multi-style latent structure of \ours makes it possible to combine
different geometric levels from different instances. Given two shapes with
latents $z_A$ and $z_B$, we obtain their style codes
$\{w^+_{A,j}\}$ and $\{w^+_{B,j}\}$ through the mapping network.
To mix shapes, we choose a split index $c$ and construct a hybrid style set
\[
\widetilde{w}^+_j =
\begin{cases}
w^+_{A,j}, & j \le c, \\
w^+_{B,j}, & j > c.
\end{cases}
\]
Decoding with $f_\theta(\cdot;\{\widetilde{w}^+_j\})$ yields a shape that
preserves coarse structure from $A$ (small $j$) and fine detail from $B$ (large
$j$). Because different style codes specialize to different depth ranges,
this operation provides a controllable, level-specific editing mechanism that
is difficult to achieve with single-latent implicit models.

\section{Experiments}
\label{sec:experiments}
In this section, we evaluate \ours{} ability to represent shapes by comparing reconstruction quality to various baselines and through interpolation within the latent space. We include implementation details, ablation studies, and additional results in the supplementary material.

\subsection{Setup}
\paragraph{Datasets.}
We evaluate on three category-level benchmarks: \textsc{Cars}, \textsc{Chairs} and \textsc{Planes}. We split the datasets into train/validation/test sets and normalize each shape to the unit cube. We sample points with their corresponding SDF to create input and targets, and extract meshes from predicted SDFs using Marching-Cubes~\cite{lorensen1998marching}.

\paragraph{Metrics.}
We report symmetric Chamfer Distance (CD;~$\downarrow$) between predicted and reference surfaces and volumetric Intersection-over-Union (IoU; $\uparrow$) computed from SDFs thresholded at zero on a dense grid. For CD, we uniformly sample points on both surfaces and average bidirectional squared distances; for IoU, we use a $256^3$ grid.

\paragraph{Baselines.}
We compare against the base approach of learning implicit representation DeepSDF~\cite{park2019deepsdf}, its curriculum version Curriculum DeepSDF~\cite{duan2020curriculum} as a baseline on coarse-to-fine learning, a variant with AdaIn style injections~\cite{huang2017arbitrary} without auxiliary output heads, which we refer to a  Curriculum Style DeepSDF, and DECOLLAGE~\cite{chen2024decollage} that performs shape detailization given a coarse voxelized input and target style shape.
We additionally compare to part-based approaches: BAE-NET~\cite{chen2019baenet} and DAE-NET~\cite{chen2023daenet} that learn an unsupervised part decomposition, and against PQ-NET~\cite{wu20pqnet}, PASTA~\cite{li2024pasta}, and PartSDF~\cite{talabot2025partsdf} that represent known part assemblies. Finally, we also compare against 3DShape2VecSet~\cite{zhang20233dshape2vecset}, a
representation that encodes shape into large \textit{sets} of latent vectors.

\subsection{Reconstruction}
We first evaluate all methods on the reconstruction of unseen test shapes.
Table~\ref{tab:main} reports results on \textsc{Cars}, \textsc{Chairs}, and \textsc{Planes} (lower CD is better; higher IoU is better).
For baselines, we report reconstruction by optimizing a single latent code $z$ at test time; for \ours{}, we additionally report reconstruction by optimizing the per-scale style codes $\{w_j^+\}_j$, which improves over optimizing $z$ alone.
Overall, \ours{} achieves the strongest performance on \textsc{Cars} and the best IoU on \textsc{Planes}, while remaining competitive on \textsc{Chairs}. We do not report part-based baselines on \textsc{Planes} since this category does not provide meaningful semantic part annotations for such methods.

On \textsc{Chairs}, the full (multi-part) PartSDF model surpasses \ours{}, which we attribute to its explicit \emph{spatial} part parameterization: it assigns each semantic component its own latent and composes the shape from multiple part SDFs, while explicitly encouraging coherent yet localized part geometry and interactions.
This inductive bias is particularly beneficial for chairs, where thin structures and strong part-level variability (e.g., legs and armrests) dominate reconstruction error, since part-wise latents enable targeted corrections to delicate regions without perturbing the rest of the shape.

In contrast, \ours{} primarily disentangles geometry across \emph{frequency bands} (coarse-to-fine) rather than across \emph{spatial parts}, which can limit how independently such localized structures can be adjusted.
Importantly, despite not using part annotations, \ours{} still outperforms the \emph{single-part} PartSDF (PartSDF-1P) variant on \textsc{Chairs}, indicating that our multi-code, scale-aware parameterization provides a stronger instance-optimization space than a single global latent even when both methods are part-agnostic at inference time.

\begin{table*}[t]
\centering
\caption{Quantitative reconstruction on \textsc{Cars}, \textsc{Chairs}, and \textsc{Planes}. Lower CD is better; higher IoU is better. All methods share splits, sampling, and extraction.}
\label{tab:main}
\setlength{\tabcolsep}{3pt}
\resizebox{\textwidth}{!}{
\begin{tabular}{lcc|cc|cccc}
\toprule
& \multicolumn{2}{c}{\textsc{Cars}} & \multicolumn{2}{c}{\textsc{Chairs}} & \multicolumn{2}{c}{\textsc{Planes}} \\
\cmidrule(lr){2-3} \cmidrule(lr){4-5} \cmidrule(lr){6-7}
Method & CD ($\downarrow$) & IoU (\%, $\uparrow$) & CD ($\downarrow$) & IoU (\%, $\uparrow$) & CD ($\downarrow$) & IoU (\%, $\uparrow$) \\
\midrule
DeepSDF \cite{park2019deepsdf}          & \texttt{$1.6\times10^{-4}$} & 97.27 & \texttt{$8.8\times10^{-4}$} & 89.74 & \texttt{$8.0\times10^{-4}$} & 89.28 \\
Curriculum DeepSDF \cite{duan2020curriculum} & \texttt{$1.6\times10^{-4}$} & 97.19 & \texttt{$1.2\times10^{-3}$} & 89.33 & \texttt{$9.2\times10^{-4}$} & 90.57 \\
Curriculum + Style DeepSDF    & \texttt{$1.3\times10^{-4}$} & 97.87 & \texttt{$5.7\times10^{-4}$} & 90.12 & \texttt{$7.8\times10^{-4}$} & 91.26 \\
DECOLLAGE~\cite{chen2024decollage} & \texttt{$1.6\times10^{-3}$} & 78.97 & \texttt{$3.0\times10^{-3}$} & 49.96 & \texttt{$3.3\times10^{-3}$} & 39.12 \\
DAE-NET~\cite{chen2023daenet} & \texttt{$2.8\times10^{-3}$} & 81.91 & \texttt{$1.0\times10^{-2}$} & 56.37 & \texttt{$-$} & \texttt{$-$} \\
BAE-NET~\cite{chen2019baenet} & \texttt{$3.3\times10^{-3}$} & 75.17 & \texttt{$1.1\times10^{-2}$} & 40.79 & \texttt{$-$} & \texttt{$-$} \\
PQ-NET~\cite{wu20pqnet} & \texttt{$3.0\times10^{-3}$} & 72.00 & \texttt{$3.9\times10^{-3}$} & 49.09 & \texttt{$-$} & \texttt{$-$} \\
PASTA~\cite{li2024pasta} & \texttt{$1.2\times10^{-3}$} & 83.88 & \texttt{$4.1\times10^{-3}$} & 60.36 & \texttt{$-$} & \texttt{$-$} \\
PartSDF~\cite{talabot2025partsdf} & \texttt{$1.3\times10^{-4}$} & 98.02 & \texttt{$\boldsymbol{1.3\times10^{-4}}$} & \textbf{97.17} & \texttt{$-$} & \texttt{$-$} \\
PartSDF-1P (Single Part)~\cite{talabot2025partsdf} & \texttt{$1.4\times10^{-4}$} & 97.41 & \texttt{$3.7\times10^{-4}$} & 93.78 & \texttt{$-$} & \texttt{$-$} \\
3DShape2VecSet~\cite{zhang20233dshape2vecset} & \texttt{$2.7\times10^{-4}$} & 92.03 & \texttt{$3.4\times10^{-4}$} & 92.13 & \texttt{$\boldsymbol{7.8\times10^{-5}}$} & \texttt{$92.10$} \\
\ours: Reconstructing with Optimizing $Z$                & \texttt{$1.3\times10^{-4}$} & 97.36 & \textbf{\texttt{$1.8\times10^{-4}$}} & 94.39 & \texttt{$6.6\times10^{-4}$} & 92.61 \\
\ours: Reconstructing with Optimizing $W+$                & \texttt{$\boldsymbol{1.2\times10^{-4}}$} & \textbf{98.43} & \texttt{$1.6\times10^{-4}$} & 95.31 & \texttt{$6.4\times10^{-4}$} & \textbf{93.66} \\
\bottomrule
\end{tabular}
}
\end{table*}

Figure~\ref{fig:qualitative_grid} shows representative reconstructions on baselines closest to \ours{} on each \textsc{Cars}, \textsc{Chairs}, and \textsc{Planes} datasets. DeepSDF recovers smooth global shapes but often under-expresses sharp creases and vents. Curriculum DeepSDF improves mid-frequency fidelity yet blurs fine patterns. Our hybrid baseline, \emph{Curriculum Style DeepSDF}, improves reconstruction fidelity on most examples; however, it still fails on one of the \textsc{Chair} samples, where thin structures and fine details are not recovered reliably. DECOLLAGE struggles to recover finer details because of its reliance on voxelized data and its limited style latent space, which does not generalize well to unseen styles. Finally, \ours reproduces cleaner silhouettes and transfers high-frequency detail on \textsc{Cars}, \textsc{Chairs}, and \textsc{Planes}.

\subsection{Global Latent Interpolation}
We additionally perform interpolation in the latent space between two shapes.
Figure~\ref{fig:interpolation_grid} shows the interpolation between global latent code $z$, following a simple linear interpolation:

\begin{equation*}
z_{\text{interpolated}}(t) = (1-t)\, z_1 + t\, z_2,
\end{equation*}
with $t \in [0, 1]$ the interpolation factor.

As shown in the figure, interpolating a single global latent code $z$ causes \emph{coarse structure} and \emph{fine detail} to change simultaneously, indicating that these scales are entangled in $z$. Concretely, as $t$ increases, global, low-frequency attributes (e.g., overall proportions and silhouette—wheelbase/roofline/hood length) drift together with high-frequency surface cues (e.g., sharp creases, vents/mirrors and rim structure). The intermediate shapes therefore exhibit \emph{averaged} geometry: their silhouette is not preserved from one endpoint while local details are borrowed from the other; instead, both levels transition at once, producing mixtures where neither the coarse form nor the fine features are clearly attributable to Shape~1 or Shape~2.

\section{Application: Shape Editing and Drag Optimization}
\label{sec:application}

\begin{table}[t]
    \centering
    \caption{Drag values for two cars under early and late $W^+$ style optimization.}
    \label{tab:drag_two_cars}
    \setlength{\tabcolsep}{4pt}
    \renewcommand{\arraystretch}{1.0}
    \begin{tabular}{m{0.18\linewidth}ccc}
        \toprule
        Car & Initial Drag & Early $W^+$ Drag & Late $W^+$ Drag \\
        \midrule
        \includegraphics[width=\linewidth]{figures/drag/s_init_z.png} &
        0.3778 & 0.3458 & 0.3642 \\
        \includegraphics[width=\linewidth]{figures/drag/p_init_z.png} &
        0.3698 & 0.3244 & 0.3501 \\
        \bottomrule
    \end{tabular}
\end{table}

\paragraph{Style Mixing}

We demonstrate how coarse structure and fine detail can be controlled independently across shapes with \ours{}.
Given two shapes, we construct hybrids by selectively swapping the AdaIN affine parameters $(\gamma_j,\beta_j)$ associated with shallow (coarse) or deep (fine) modulation groups. Figure~\ref{fig:style_mixing} visualizes this process. 

In Fig.~\ref{fig:style_mixing}\textbf{(A)}, keeping coarse styles $(w_0^+,w_1^+)$ from Shape~1 preserves its global silhouette and proportions, while replacing fine styles $(w_2^+,w_3^+)$ transfers Shape~2’s high-frequency cues (e.g., sharper creases and front-fascia details in the normal map). Fig.~\ref{fig:style_mixing}\textbf{(C)} shows the complementary behavior: swapping only coarse styles shifts the overall form toward Shape~2 while retaining Shape~1’s fine-scale surface detail. Finally, Fig.~\ref{fig:style_mixing}\textbf{(B)} highlights intermediate control: taking the mid-level styles $(w_1^+,w_2^+)$ from Shape~1 retains meso-scale panel structure.

Overall, the consistent, targeted changes across these hybrids indicate that different style groups predominantly control different geometric scales, enabling predictable cross-instance editing rather than entangled global interpolation.

\paragraph{Drag Optimization}

We apply our framework to aerodynamics-aware shape editing on \textsc{Cars}, using a learned surrogate that predicts drag directly from a surface mesh. Since Marching Cubes~\cite{lorensen1998marching} is not differentiable, we convert our predicted SDF into a differentiable surface representation using MeshSDF~\cite{remelli2020meshsdf}, which allows gradients from the drag objective to flow back to the latent/style parameters through the surface extraction stage. Concretely, at each optimization step we decode an SDF, extract a differentiable mesh with MeshSDF, and minimize the surrogate-predicted drag with gradient-based optimization, while keeping the edit close to the original reconstruction via a latent/style proximity regularizer to avoid unrealistic drift.

\begin{figure*}[t]
    \centering
    \setlength{\tabcolsep}{1pt}
    \renewcommand{\arraystretch}{1.0}

    \setlength{\imgw}{\dimexpr(\textwidth-6\tabcolsep)/4\relax}

    \begin{tabular}{@{}cccc@{}}
        \includegraphics[width=\imgw]{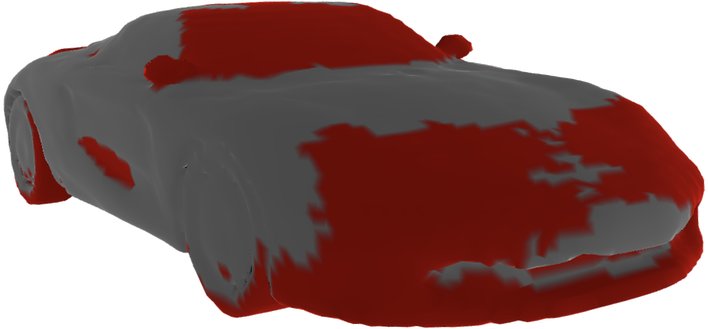} &
        \includegraphics[width=\imgw]{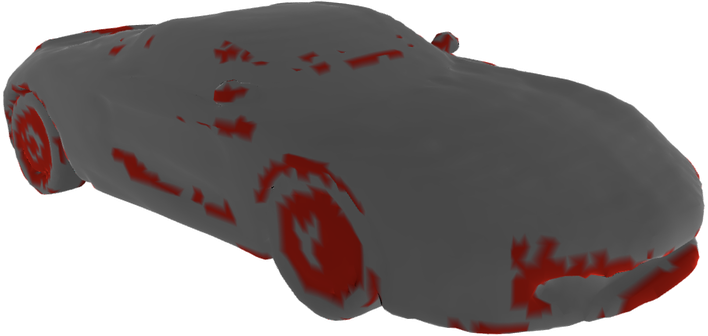} &
        \includegraphics[width=\imgw]{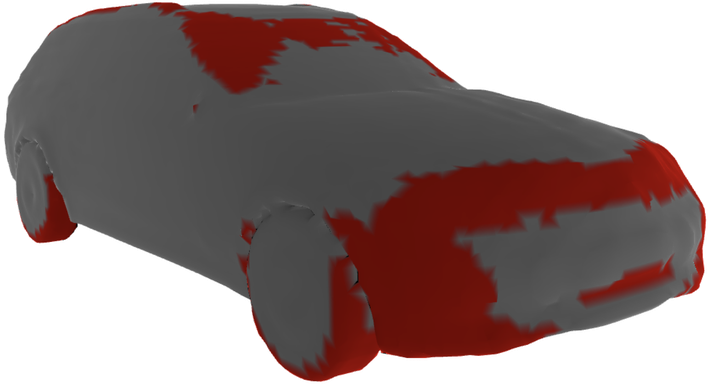} &
        \includegraphics[width=\imgw]{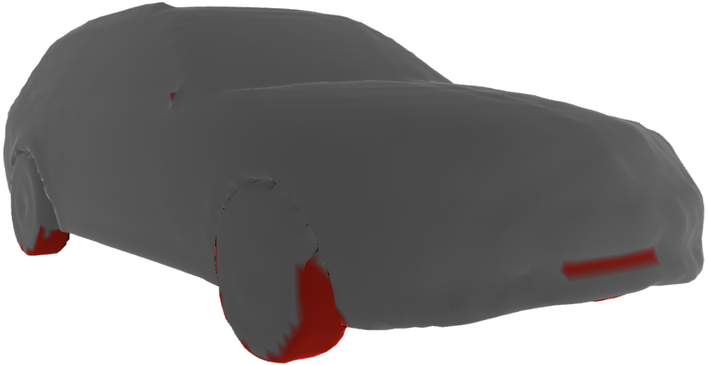} \\

        \parbox[t]{\imgw}{\centering\scriptsize Shape 1:\\Optimizing $w_0^+$ \& $w_1^+$} &
        \parbox[t]{\imgw}{\centering\scriptsize Shape 1:\\Optimizing $w_2^+$ \& $w_3^+$} &
        \parbox[t]{\imgw}{\centering\scriptsize Shape 2:\\Optimizing $w_0^+$ \& $w_1^+$} &
        \parbox[t]{\imgw}{\centering\scriptsize Shape 2:\\Optimizing $w_2^+$ \& $w_3^+$}
    \end{tabular}

    \caption{
        Drag-aware edits for four shapes. In each visualization, red regions highlight surface areas modified by the optimization to reduce drag.
    }
    \label{fig:drag_optimized_regions}
\end{figure*}

We explore three optimization regimes: (i) adjusting the global latent code $z$, (ii) modifying only the \emph{early} $w^+$ style codes associated with coarse geometry, and (iii) modifying only the \emph{late} $w^+$ style codes associated with fine-scale detail. Figure~\ref{fig:drag_optimization} compares the original reconstructions with their optimized counterparts. Optimizing the global latent produces large, holistic shape changes and tends to move different shapes toward a similar low-drag configuration, reflecting the entanglement of high-vs-low frequency structure in $z$. In contrast, restricting the optimization to subsets of $w^+$ enables targeted edits: early-layer optimization primarily adjusts low-frequency structure (global proportions and silhouette), whereas late-layer optimization refines high-frequency surface detail without substantially altering the overall shape class.

Table~\ref{tab:drag_two_cars} reports the drag values for two representative cars under early and late $w^+$ optimization: in both cases, optimizing early $w^+$ yields the largest drag reduction, whereas late $w^+$ optimization provides more modest improvements. Figure~\ref{fig:drag_optimized_regions} explains this behavior by showing where the optimizer modifies the surface (red): early-$w^+$ (coarse) optimization produces broad, contiguous edits over large aerodynamic structures (e.g., front fascia/hood and rear volume), reflecting low-frequency reshaping that most strongly affects pressure drag, while late-$w^+$ (fine) optimization concentrates changes into smaller, localized patches around sharp edges and curvature transitions, primarily refining high-frequency detail and thus having a smaller impact on drag while preserving the overall silhouette.

All additional ablations and extended analyses are provided in the supplementary material, including detailed studies of interpolation (denser interpolations are visualized in supplementary) and cross-instance style mixing on \textsc{Chairs} and \textsc{Planes}, further qualitative results and implementation details, as well as an expanded description and additional results for our drag-based optimization procedure.

\section{Conclusion and Future Work}
\label{sec:conclusion}
We presented a reconstruction-first framework that disentangles coarse structure from fine geometric detail in implicit SDF decoders, enabling controllable style mixing and faithful, instance-conditioned reconstruction. Beyond reconstruction, we demonstrated a practical application to aerodynamics-aware editing via surrogate drag optimization.
For future work, we will explore stronger supervision and representations that provide clearer, tunable control over geometric frequency.

\bibliographystyle{splncs04}
\bibliography{main}

\clearpage

\clearpage

\begin{center}
    {\LARGE\bfseries Supplementary Material}
\end{center}

\vspace{1em}

\section{Low-vs-high Frequency Features Disentanglement Analysis}

\subsection{Qualitative Evidence of Coarse-to-Fine Disentanglement}

To further examine whether the proposed representation induces a coarse-to-fine organization of shape information, we compare t-SNE visualizations obtained from three latent representations: the \emph{early} \(w^+\) style codes (\(w_0^+, w_1^+\)), the \emph{global} latent \(z\), and the \emph{late} \(w^+\) style codes (\(w_2^+, w_3^+\)). We select an ``anchor'' shape together with its \(12\) nearest neighbors and \(13\) farthest samples from test set according to Euclidean distance in the corresponding latent space, and visualize them jointly using t-SNE. In all plots, the anchor is highlighted with a blue box; the \(12\) nearest samples are shown in blue, while the \(13\) farthest samples are shown in red.

\begin{figure*}[t]
    \centering

    \begin{subfigure}{\textwidth}
        \centering
        \includegraphics[width=\textwidth]{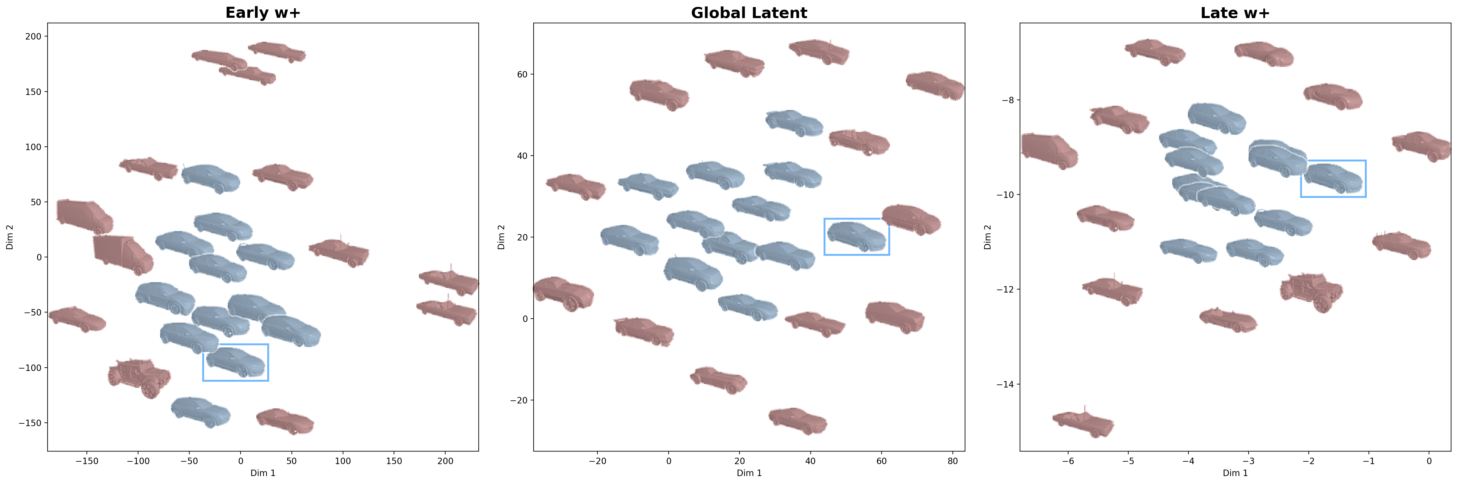}
        \caption{Anchor 1}
    \end{subfigure}

    \begin{subfigure}{\textwidth}
        \centering
        \includegraphics[width=\textwidth]{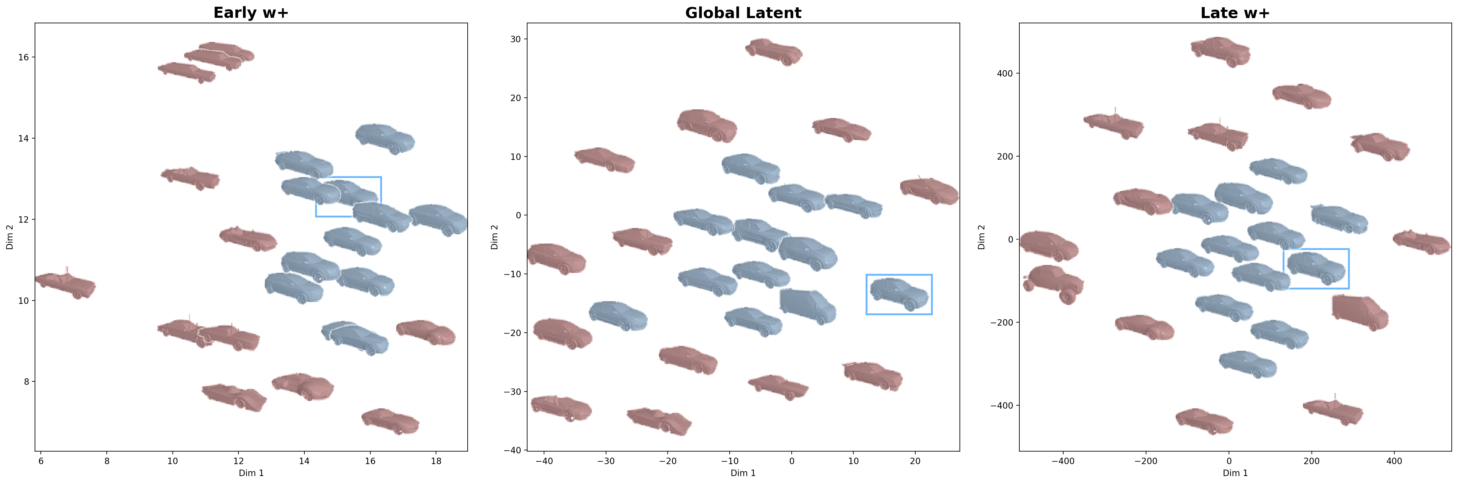}
        \caption{Anchor 2}
    \end{subfigure}

    \begin{subfigure}{\textwidth}
        \centering
        \includegraphics[width=\textwidth]{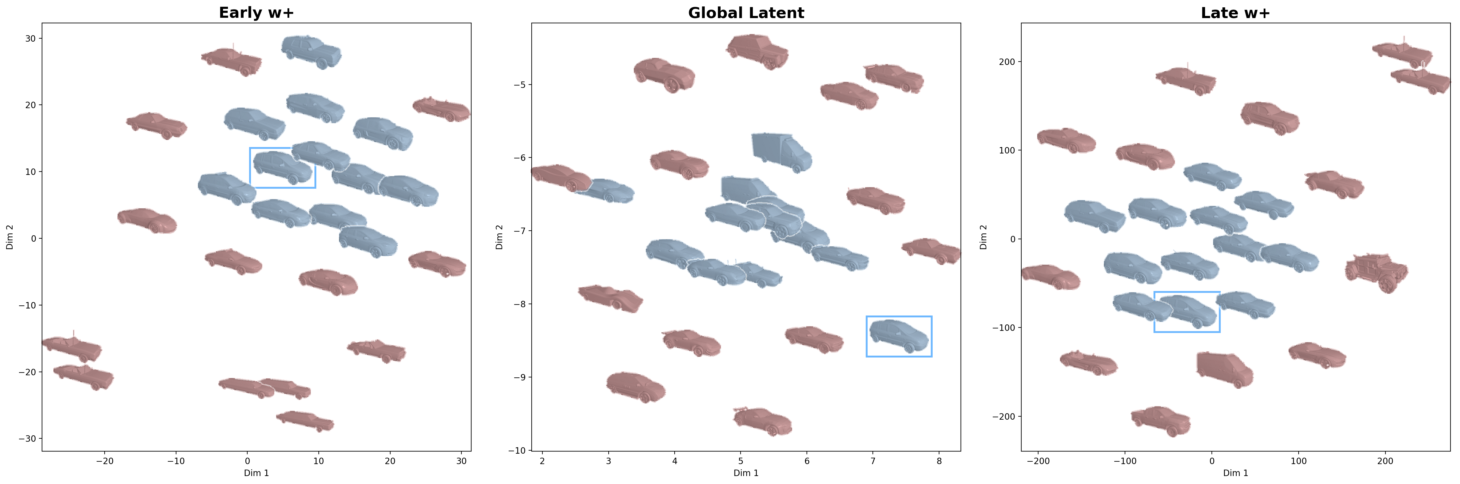}
        \caption{Anchor 3}
    \end{subfigure}

    \caption{t-SNE comparisons for anchor shapes using three latent representations: early \(w^+\), global latent \(z\), and late \(w^+\). In each row, the anchor shape is highlighted by the red box.}
    \label{fig:supp_tsne_disentanglement_set1}
\end{figure*}

\begin{figure*}[t]
    \centering

    \begin{subfigure}{\textwidth}
        \centering
        \includegraphics[width=\textwidth]{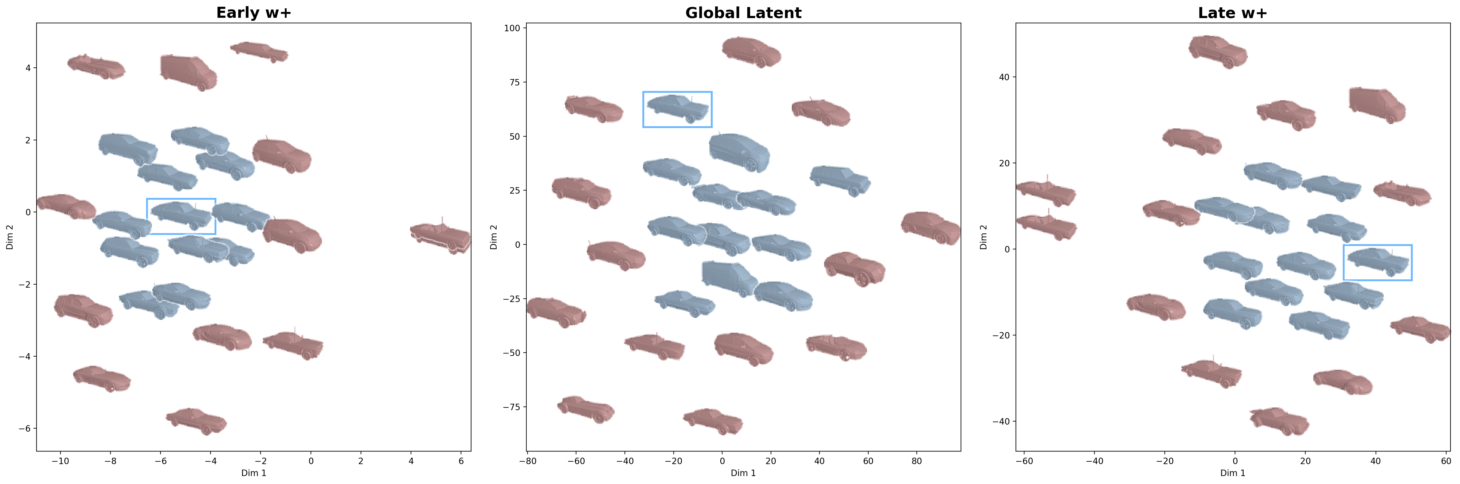}
        \caption{Anchor 4}
    \end{subfigure}

    \begin{subfigure}{\textwidth}
        \centering
        \includegraphics[width=\textwidth]{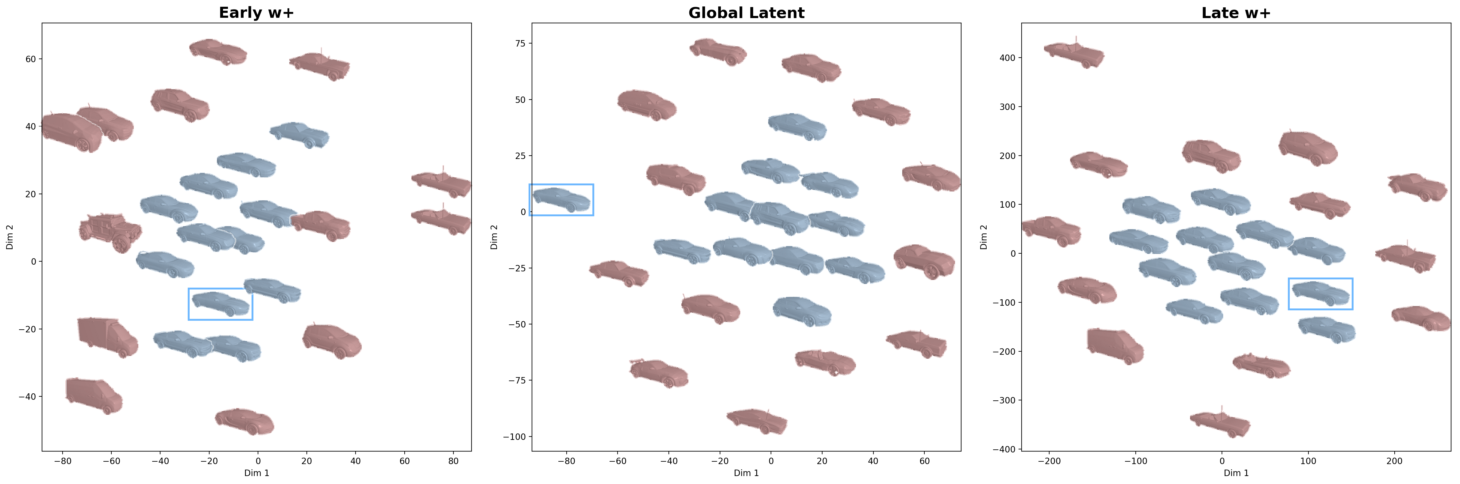}
        \caption{Anchor 5}
    \end{subfigure}

    \begin{subfigure}{\textwidth}
        \centering
        \includegraphics[width=\textwidth]{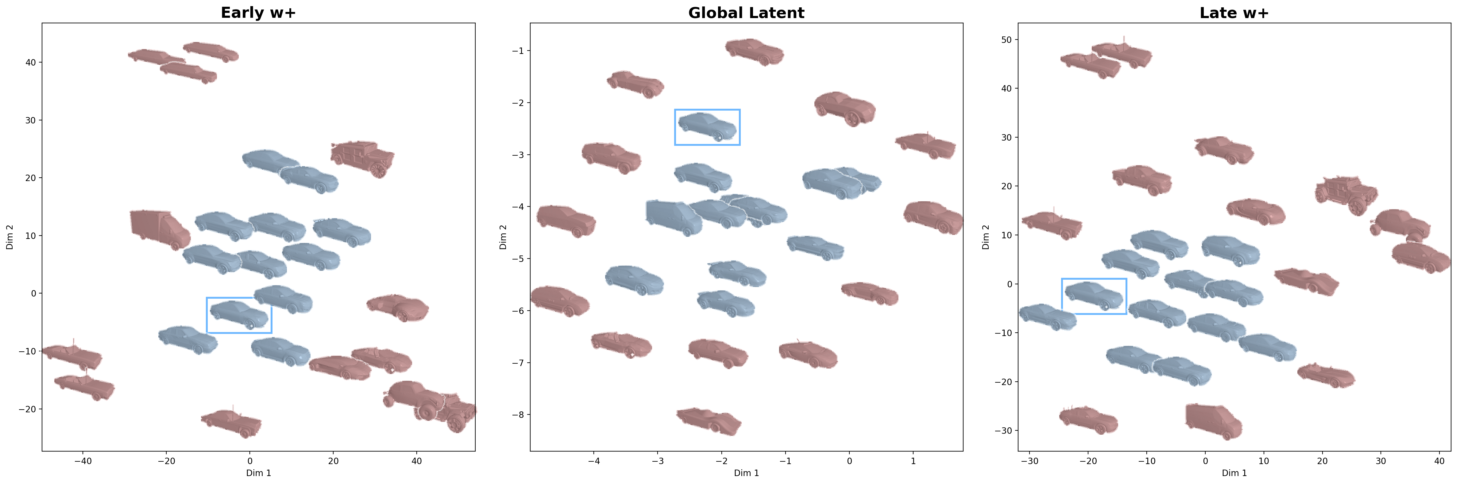}
        \caption{Anchor 6}
    \end{subfigure}

    \caption{Additional anchor-wise comparisons showing the same qualitative behavior across latent spaces.}
    \label{fig:supp_tsne_disentanglement_set2}
\end{figure*}

\begin{figure*}[t]
    \centering

    \begin{subfigure}{\textwidth}
        \centering
        \includegraphics[width=\textwidth]{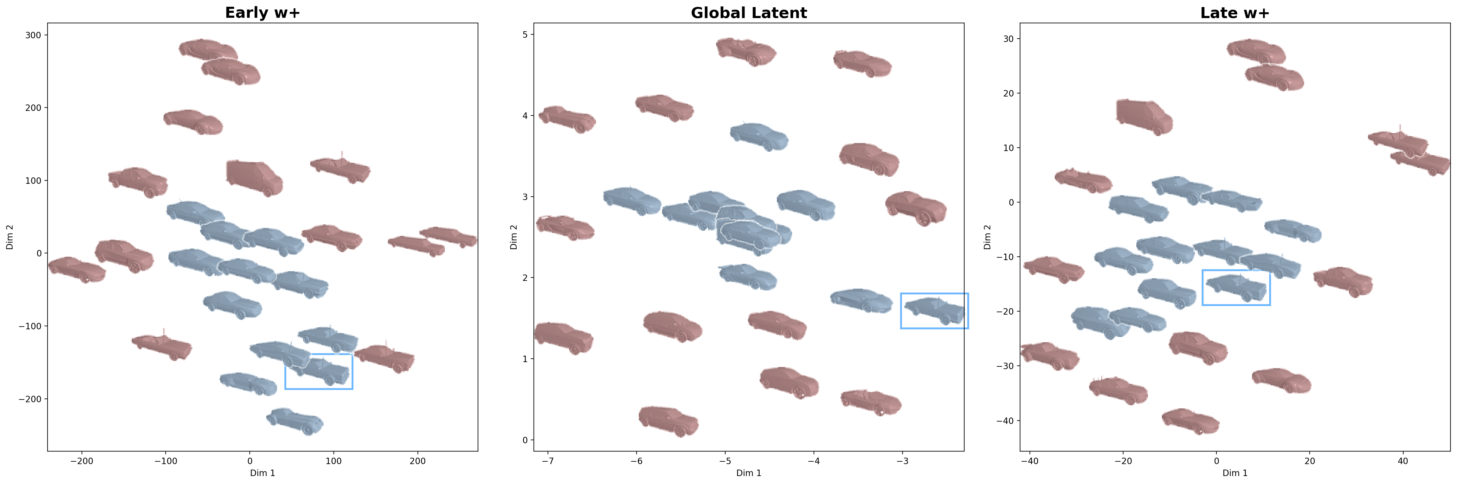}
        \caption{Anchor 7}
    \end{subfigure}

    \begin{subfigure}{\textwidth}
        \centering
        \includegraphics[width=\textwidth]{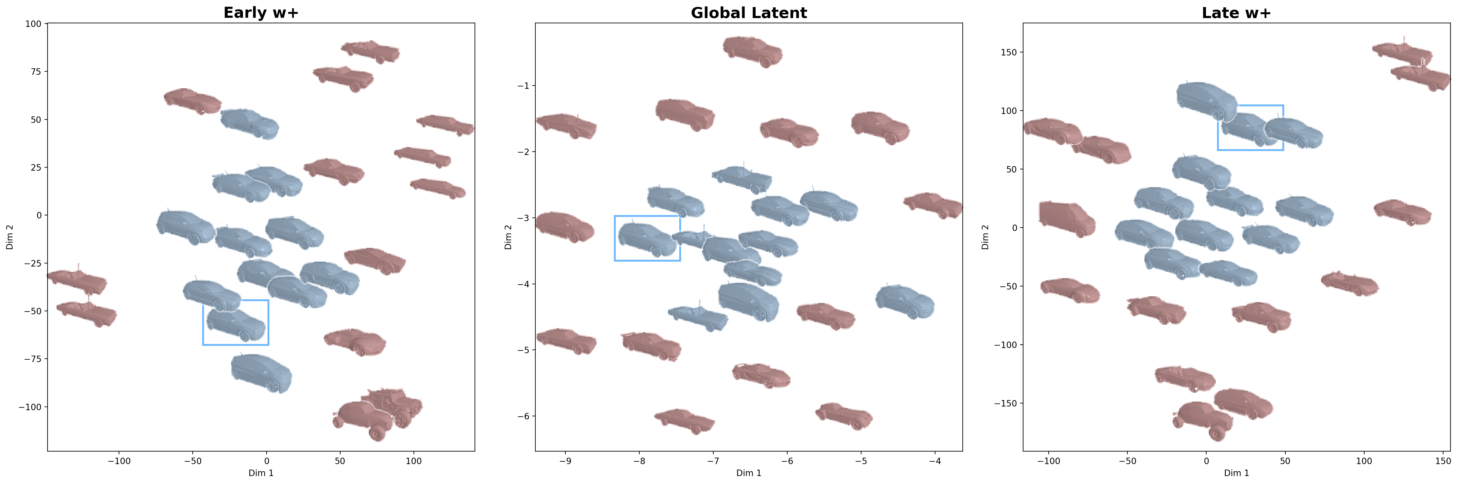}
        \caption{Anchor 8}
    \end{subfigure}

    \begin{subfigure}{\textwidth}
        \centering
        \includegraphics[width=\textwidth]{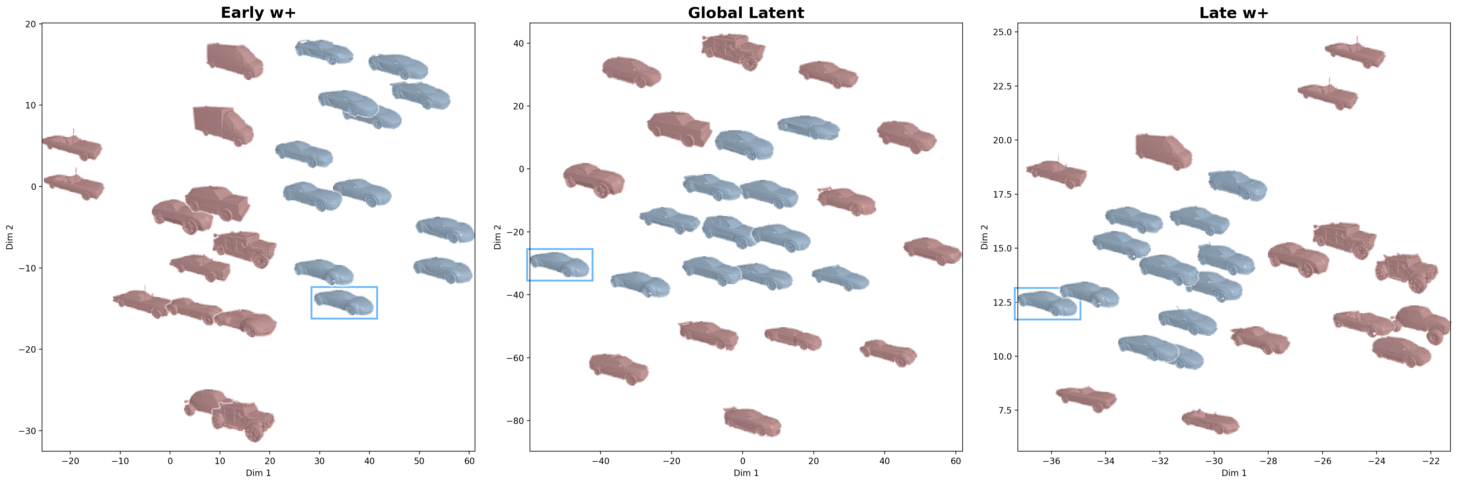}
        \caption{Anchor 9}
    \end{subfigure}

    \begin{subfigure}{\textwidth}
        \centering
        \includegraphics[width=\textwidth]{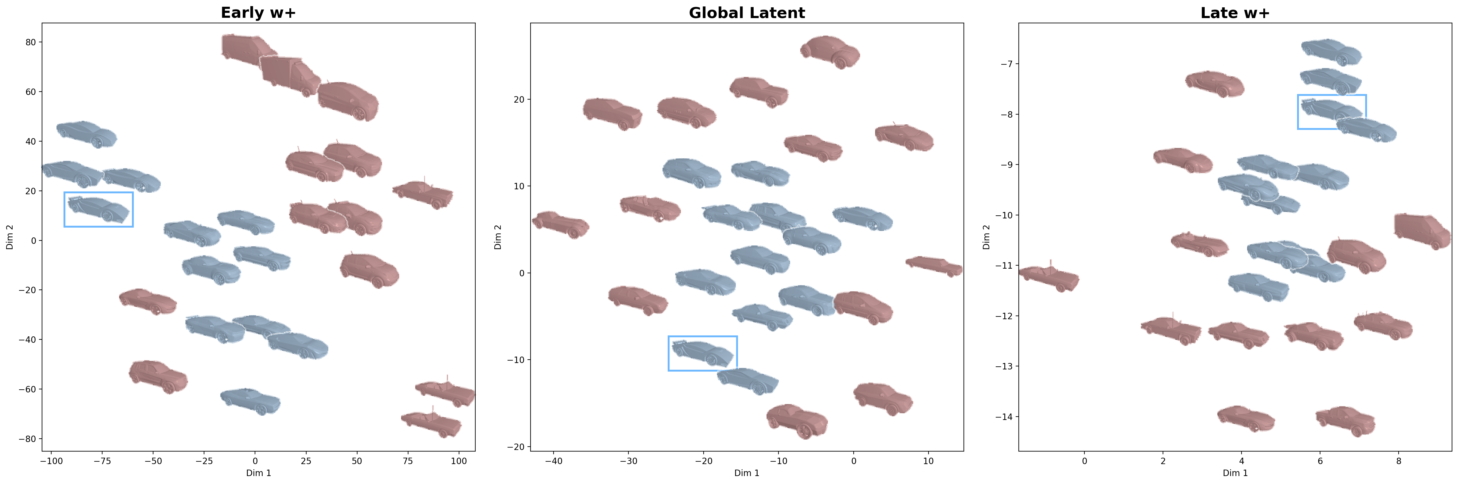}
        \caption{Anchor 10}
    \end{subfigure}

    \caption{Further examples illustrating the consistency of the latent-space organization across different anchor shapes.}
    \label{fig:supp_tsne_disentanglement_set3}
\end{figure*}

\cref{fig:supp_tsne_disentanglement_set1,fig:supp_tsne_disentanglement_set2,fig:supp_tsne_disentanglement_set3}, show the results for ten different anchors. Across these examples, the \emph{early} \(w^+\) representation exhibits the clearest organization by coarse geometry. In the left panel of each triplet, the blue neighbors typically have low variance in overall silhouette, body proportions, roofline, and cabin placement. This behavior is particularly evident for sedan-like anchors, where the nearest samples remain within the same broad coarse family, while visibly different shapes are pushed into the red set. A recurring example is the presence of highly elongated limousine-like cars among the farthest samples: in many anchors, such vehicles are consistently colored red in the early-\(w^+\) panel, indicating that strong deviations in global structure are heavily penalized in this space. More generally, vans, open-body/off-road vehicles, and unusually stretched cars are usually separated from the anchor when coarse shape differs substantially. These trends indicate that distances in the early-\(w^+\) space are strongly aligned with low-frequency geometric similarity.

By contrast, the \emph{global} latent \(z\) is noticeably more entangled. In the middle panel, the blue neighborhood around the anchor often exhibits substantially higher shape variance than in the early-\(w^+\) panel: cars with different overall lengths, cabin proportions, and body layouts can appear simultaneously within the nearest set. Thus, even when the anchor belongs to a relatively coherent sedan-like family, its nearest neighbors in \(z\) may span a broader and less semantically consistent range of coarse structures. Qualitatively, this suggests that the single global latent mixes multiple factors of variation, rather than separating coarse and fine geometric attributes into distinct levels of control.

The \emph{late} \(w^+\) codes exhibit a different behavior from the early ones. In the right panel of each triplet, the neighborhoods are generally less consistently organized by broad body class and large-scale silhouette. Although some anchors still retain nearby samples with similar overall shape, this trend is weaker and less stable across examples than in early \(w^+\). In particular, late-\(w^+\) proximity appears less dominated by coarse structure, which is consistent with the intended role of deeper style modulations in controlling more localized and higher-frequency geometric variation. Thus, while late \(w^+\) does not become completely independent of global shape, it no longer provides as clean a grouping by coarse form as early \(w^+\). An instructive example is again provided by limousine-like cars. While they frequently appear among the red farthest samples in the early-\(w^+\) panel, they do not appear among the red samples in the late-\(w^+\) panel across these examples. This suggests that, despite their coarse-shape mismatch, limousine-like cars remain relatively competitive in late-\(w^+\) distance because their high-frequency geometric characteristics are closer to the anchor than those of many other far samples. In other words, late \(w^+\) is not organized primarily by coarse silhouette; it is comparatively more sensitive to finer-scale geometric attributes.

At the same time, the late-\(w^+\) neighborhoods are not as strikingly cross-class as one might expect under an idealized notion of complete fine-detail disentanglement, where cars with substantially different global structure but similar local detail would appear adjacent to one another more frequently. A likely reason is the bias of the underlying dataset itself: the dataset contains only a limited set of naturally occurring combinations of coarse structure and local detail, and therefore does not densely populate the space of globally different yet locally similar vehicles. For example, one generally does not observe van-like bodies with sports-car-like local styling, or other equally extreme cross-family combinations, in sufficient quantity for such pairings to dominate nearest-neighbor retrieval in late \(w^+\). Consequently, some residual alignment with coarse shape is expected even if the latent is substantially more detail-biased than the early codes. For this reason, nearest-neighbor visualization alone is not the strongest test of the role of late \(w^+\). In the next subsection, we therefore complement this analysis with cross-instance style-mixing results, which provide a more direct intervention-based evaluation of whether deeper style codes preserve local geometric character while allowing coarse structure to vary.

\subsection{Additional Cross-instance Style Mixing Analysis for Late \texorpdfstring{$w^+$}{w+}}
\label{sec:supp_late_wplus_style_mixing}

\begin{figure*}[t]
    \centering

    \begin{subfigure}{\textwidth}
        \centering
        \includegraphics[width=\textwidth]{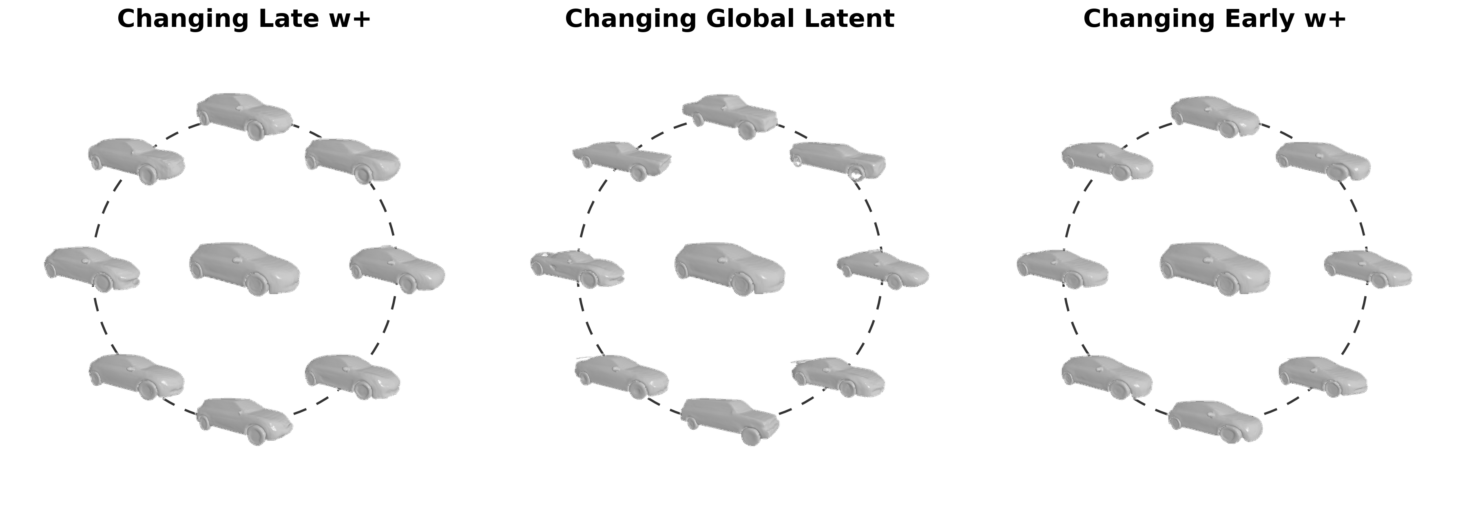}
        \caption{Anchor 1}
    \end{subfigure}

    \begin{subfigure}{\textwidth}
        \centering
        \includegraphics[width=\textwidth]{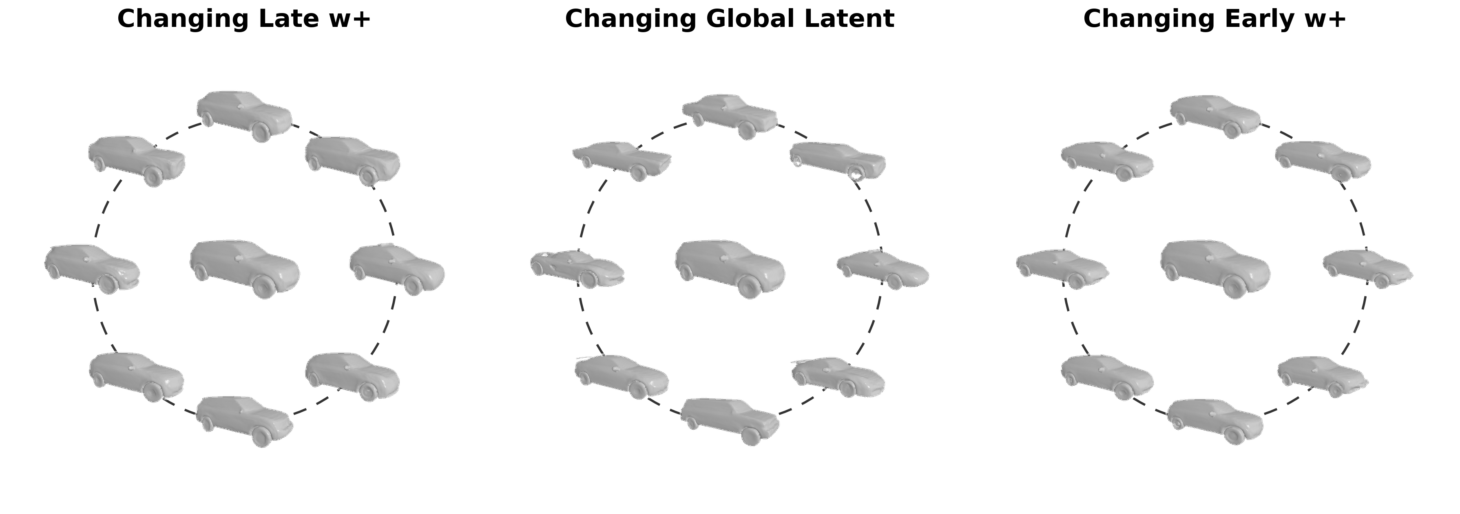}
        \caption{Anchor 2}
    \end{subfigure}

    \begin{subfigure}{\textwidth}
        \centering
        \includegraphics[width=\textwidth]{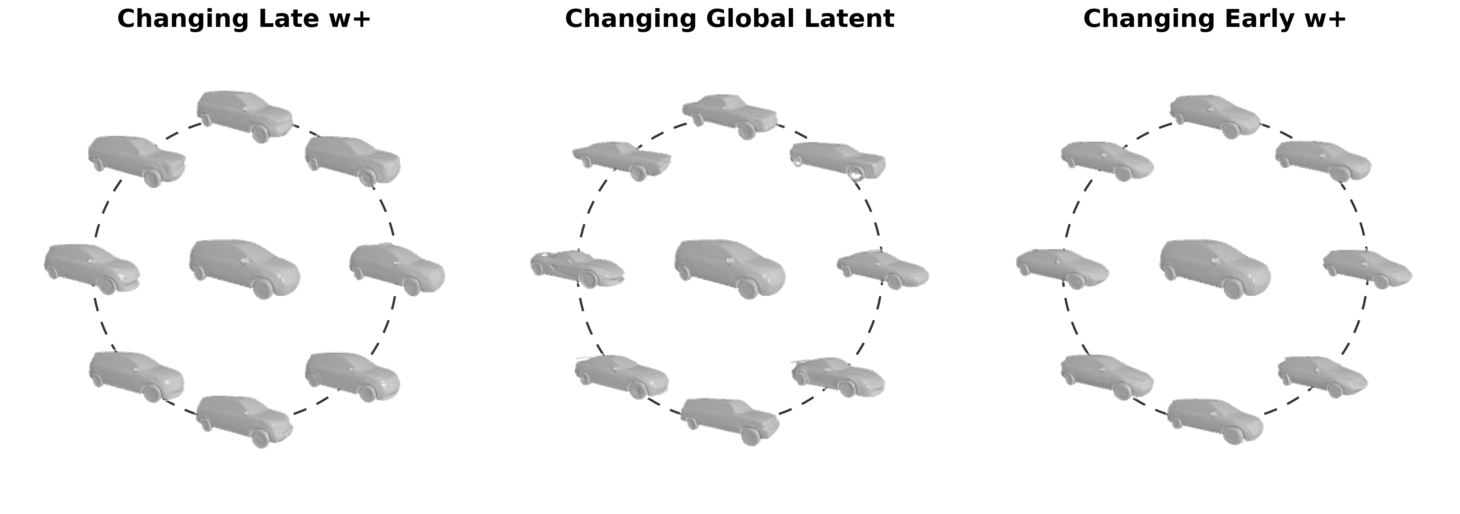}
        \caption{Anchor 3}
    \end{subfigure}

    \caption{Additional cross-instance style-mixing examples for analyzing the role of late \(W^+\). Across examples, substituting deeper style codes while keeping the early style codes fixed tends to preserve the anchor’s coarse silhouette and global proportions more strongly than edits involving the single global latent, while still altering finer geometric characteristics.}
    \label{fig:supp_late_wplus_mix_1}
\end{figure*}

\begin{figure*}[t]
    \centering

    \begin{subfigure}{\textwidth}
        \centering
        \includegraphics[width=\textwidth]{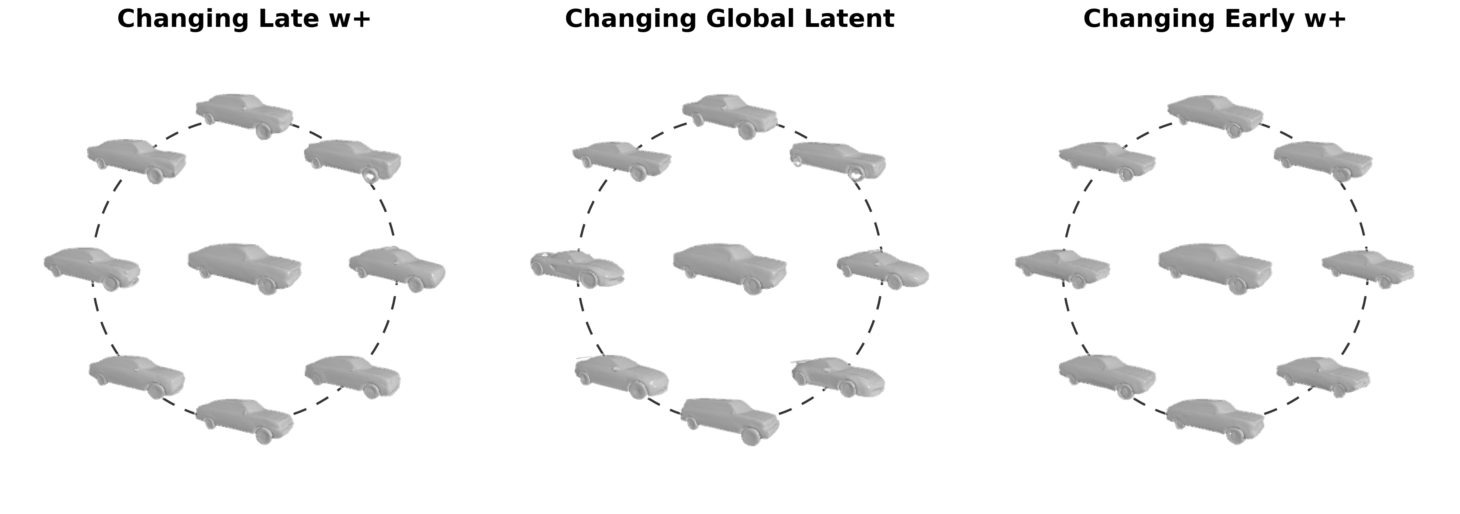}
        \caption{Anchor 4}
    \end{subfigure}

    \begin{subfigure}{\textwidth}
        \centering
        \includegraphics[width=\textwidth]{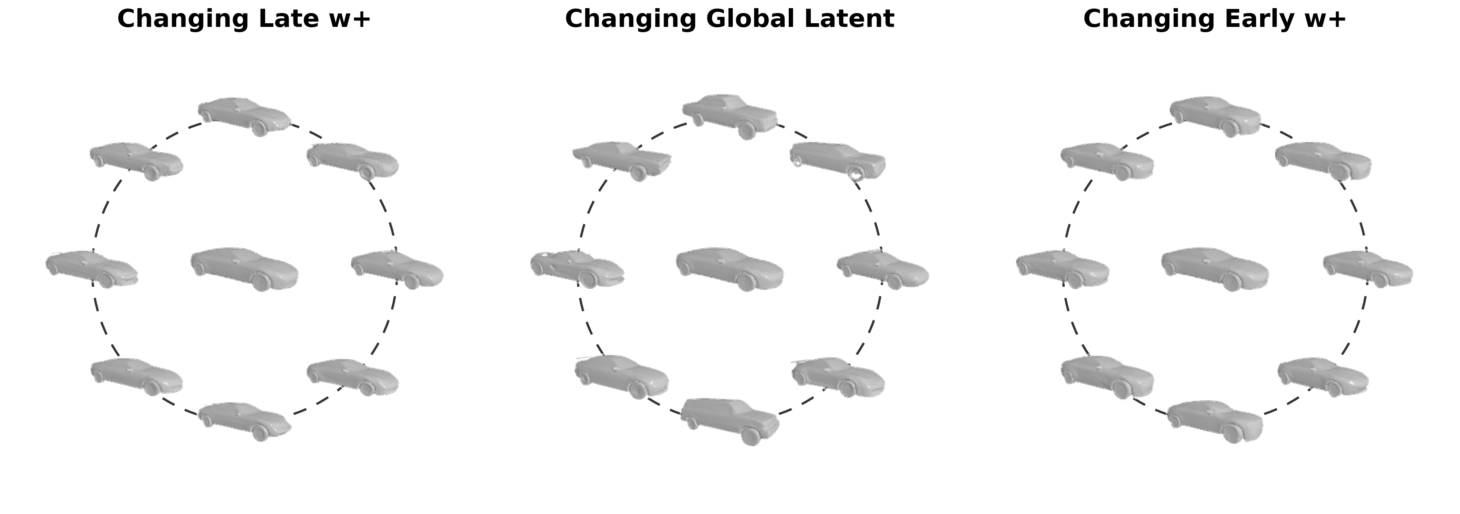}
        \caption{Anchor 5}
    \end{subfigure}

    \begin{subfigure}{\textwidth}
        \centering
        \includegraphics[width=\textwidth]{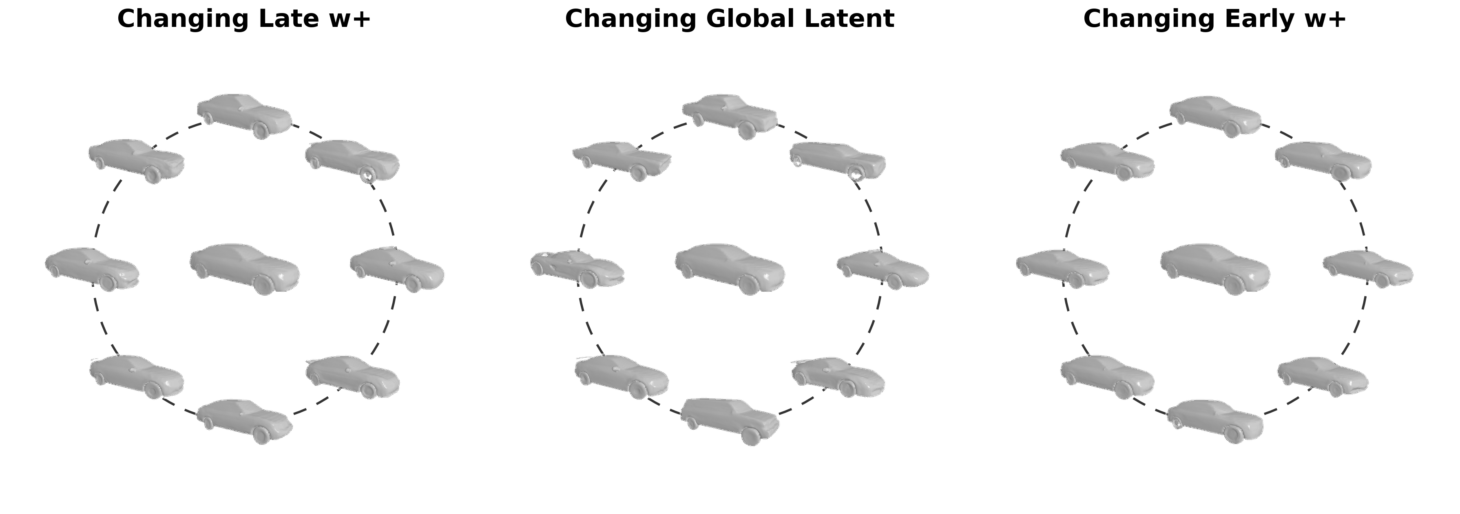}
        \caption{Anchor 6}
    \end{subfigure}

    \caption{Five more cross-instance style-mixing examples continuing the same trend.}
    \label{fig:supp_late_wplus_mix_2}
\end{figure*}

\begin{figure*}[t]
    \centering

    \begin{subfigure}{\textwidth}
        \centering
        \includegraphics[width=\textwidth]{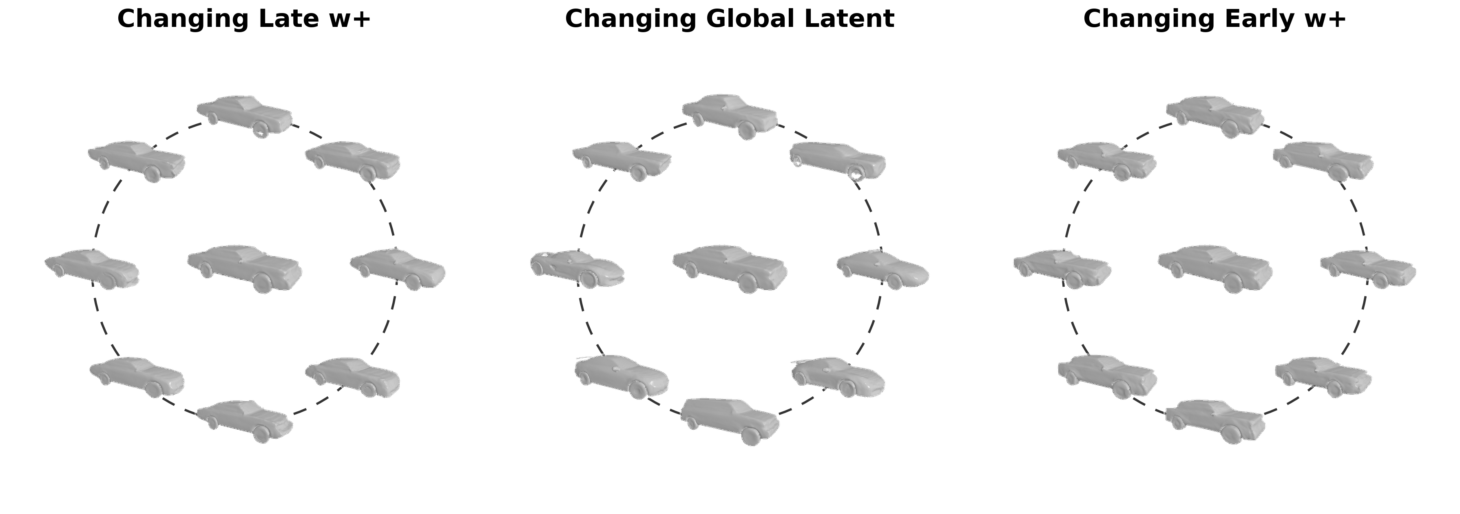}
        \caption{Anchor 7}
    \end{subfigure}

    \begin{subfigure}{\textwidth}
        \centering
        \includegraphics[width=\textwidth]{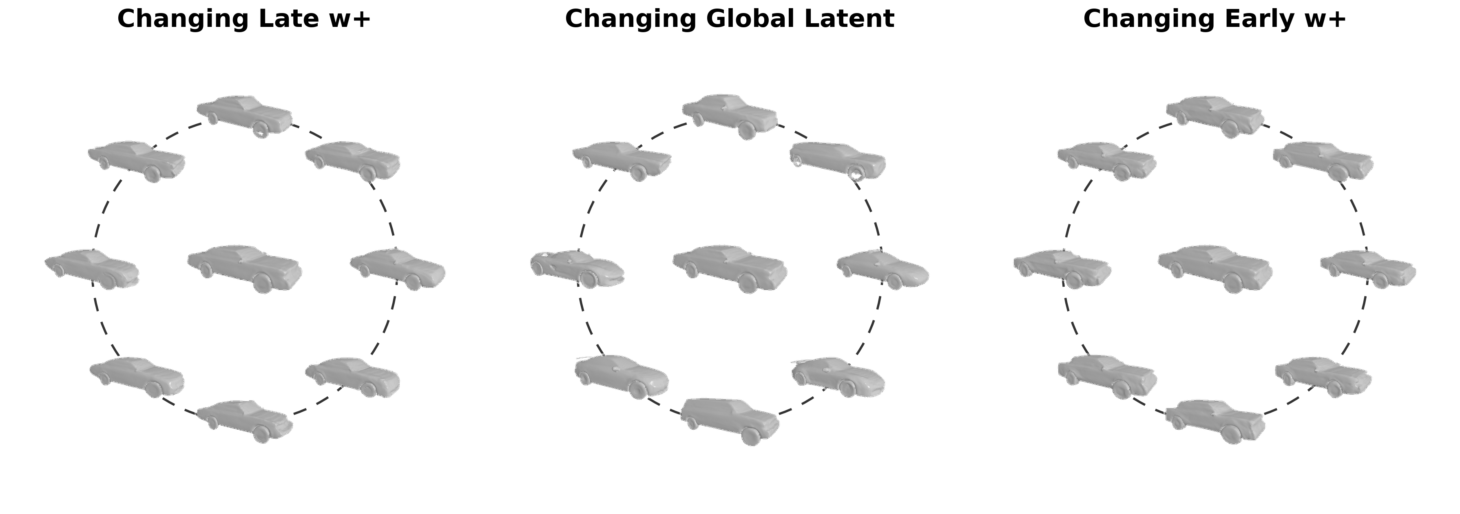}
        \caption{Anchor 8}
    \end{subfigure}

    \begin{subfigure}{\textwidth}
        \centering
        \includegraphics[width=\textwidth]{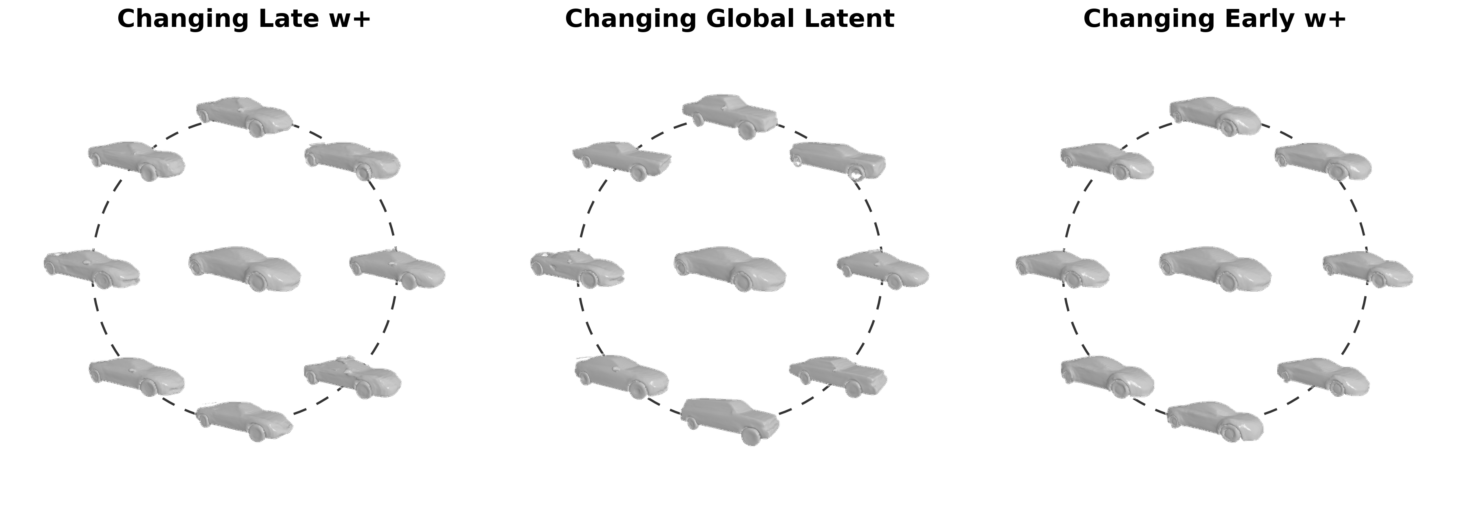}
        \caption{Anchor 9}
    \end{subfigure}

    \caption{Additional examples showing that deeper-code substitution yields more localized, detail-oriented changes, whereas changing the early style codes produces larger variation in coarse structure.}
    \label{fig:supp_late_wplus_mix_3}
\end{figure*}

\begin{figure*}[t]
    \centering

    \begin{subfigure}{\textwidth}
        \centering
        \includegraphics[width=\textwidth]{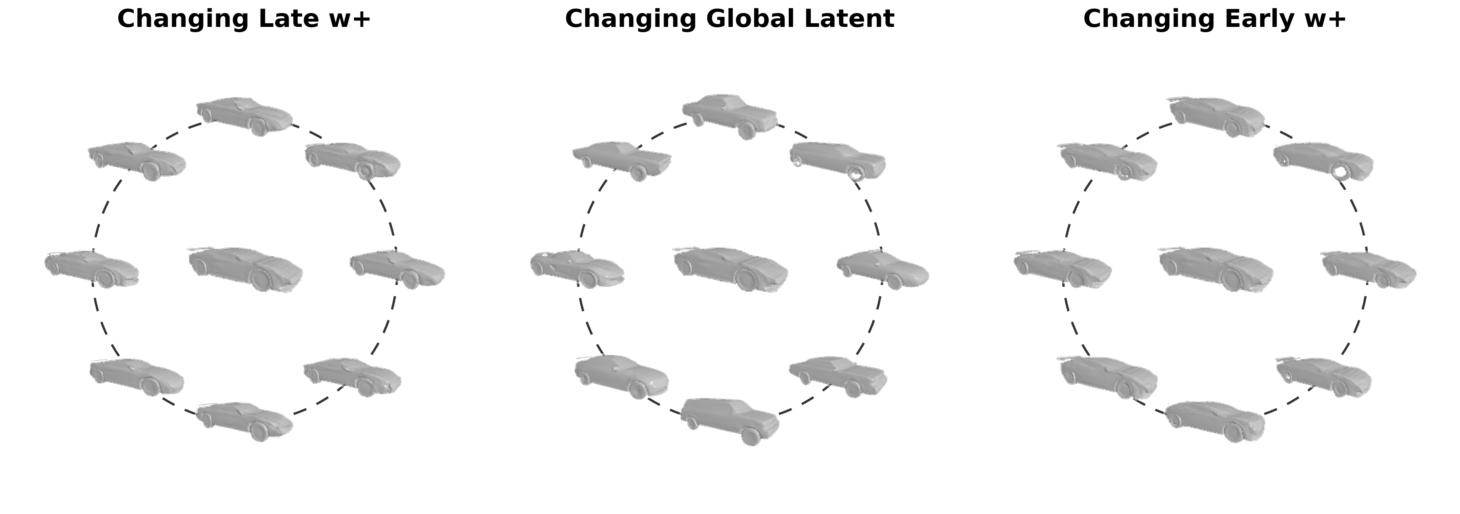}
        \caption{Anchor 10}
    \end{subfigure}

    \caption{Final style-mixing example supporting the interpretation that late \(W^+\) predominantly influences finer geometric detail rather than serving as a second entangled global code.}
    \label{fig:supp_late_wplus_mix_4}
\end{figure*}

To further clarify the role of the deeper style codes, we provide ten additional cross-instance style-mixing examples in \cref{fig:supp_late_wplus_mix_1,fig:supp_late_wplus_mix_2,fig:supp_late_wplus_mix_3,fig:supp_late_wplus_mix_4}. These visualizations are intended to complement the latent-space analyses by directly testing the intervention that is most relevant to the main paper: exchanging subsets of style codes across shapes. This is a more direct probe of the representation than latent-space visualization alone, since the central claim of \ours is not merely that the latent spaces cluster differently, but that different depth ranges support different kinds of geometric edits.

Across these examples, the late-\(w^+\) style codes exhibit a clear and consistent role. When the early style codes are held fixed and only the deeper style codes are replaced, the resulting shape usually preserves the anchor’s overall coarse structure, including its broad silhouette, body proportions, and global shape class, while changing more localized geometric characteristics. Conversely, when the late style codes are held fixed and the early style codes are changed, the generated shapes undergo much larger variation in global form. This asymmetry indicates that the deeper style codes are not acting as a second global latent. Rather, they are biased toward finer-scale geometric variation.

At the same time, these results should not be interpreted as implying that late \(w^+\) is completely free of coarse information. A more accurate conclusion, and the one supported throughout the paper, is that \ours induces a coarse-to-fine organization in which early style groups are primarily aligned with global structure, while deeper style groups predominantly control high-frequency detail. The additional style-mixing examples reinforce this interpretation by showing that substituting deeper codes yields targeted and predictable cross-instance edits that are substantially more localized than those produced by manipulating the single global latent \(z\). This is consistent with the optimization results in the main paper, where late-\(w^+\) updates refine high-frequency detail without substantially changing the overall shape class, whereas optimizing the single global latent leads to larger and more holistic changes.

Taken together, these examples provide complementary evidence for the role of late \(w^+\). Although the deeper codes are not perfectly isolated from all coarse information, they show a clear and practically useful specialization toward local geometric detail. This is precisely the property that enables the controllable style-mixing behavior emphasized in the main paper: preserving global structure while selectively modifying finer geometric characteristics across instances.

\section{Denser Global-Latent Interpolations}

\begin{figure*}[p]
    \centering
    {\small\textbf{DeepSDF}}\\[-0.15em]
\makebox[\textwidth][c]{%
\includegraphics[width=0.155\textwidth]{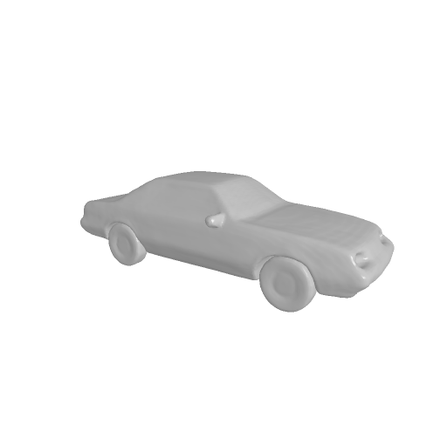}%
\includegraphics[width=0.155\textwidth]{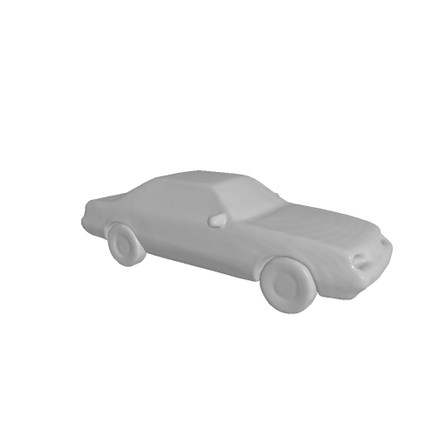}%
\includegraphics[width=0.155\textwidth]{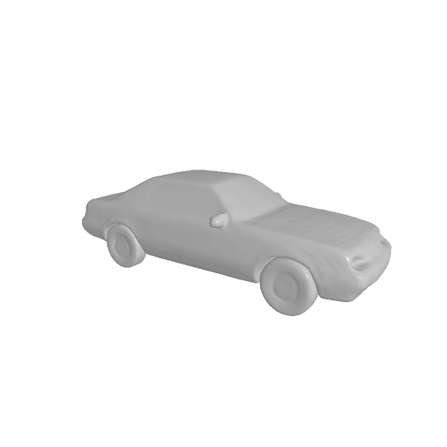}%
\includegraphics[width=0.155\textwidth]{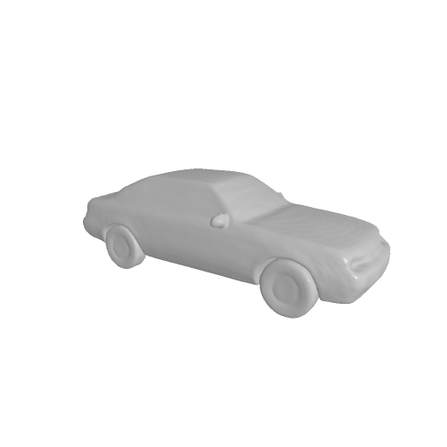}%
\includegraphics[width=0.155\textwidth]{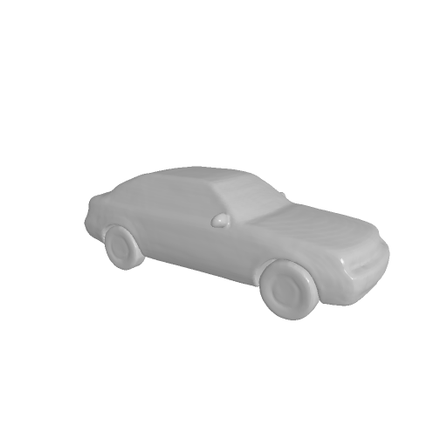}%
\includegraphics[width=0.155\textwidth]{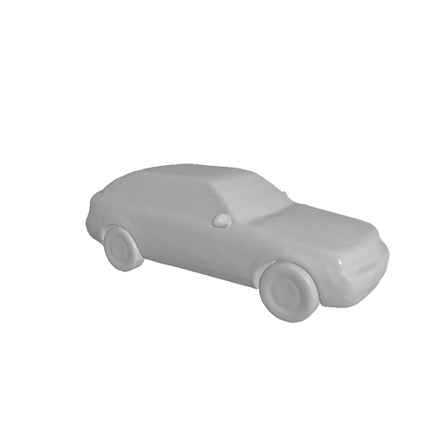}%
}\\[-0.15em]
\makebox[\textwidth][c]{%
\includegraphics[width=0.155\textwidth]{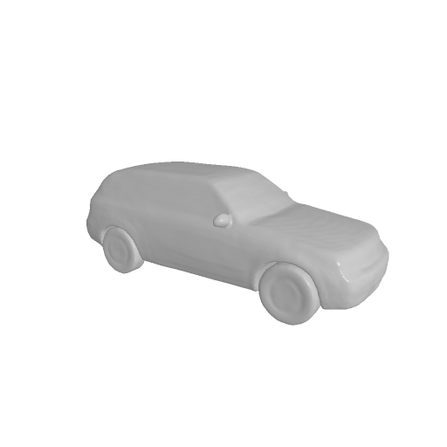}%
\includegraphics[width=0.155\textwidth]{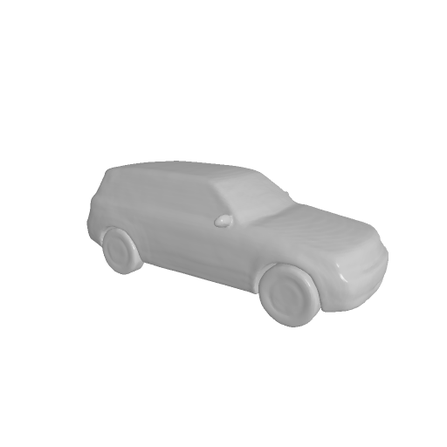}%
\includegraphics[width=0.155\textwidth]{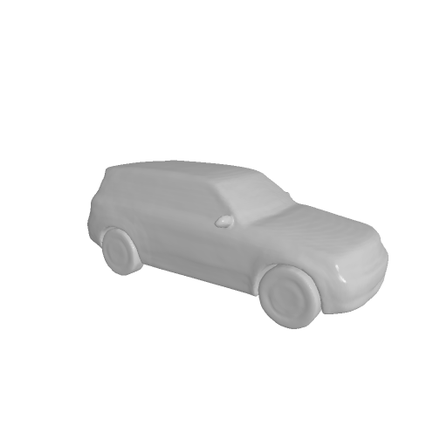}%
\includegraphics[width=0.155\textwidth]{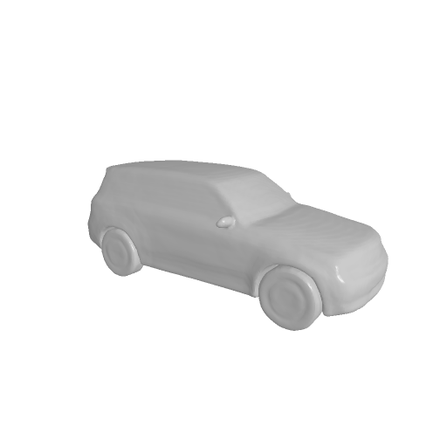}%
\includegraphics[width=0.155\textwidth]{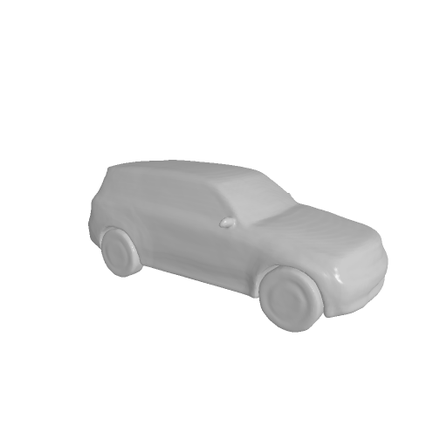}%
}\\[0.55em]
    {\small\textbf{Curriculum DeepSDF}}\\[-0.15em]
\makebox[\textwidth][c]{%
\includegraphics[width=0.155\textwidth]{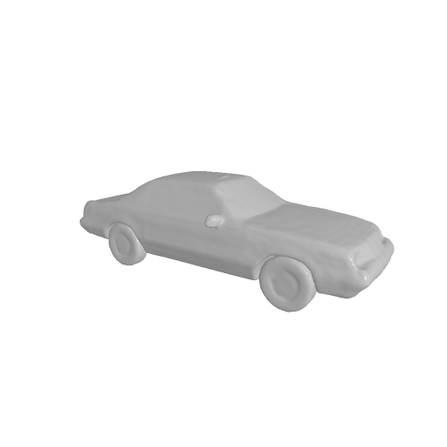}%
\includegraphics[width=0.155\textwidth]{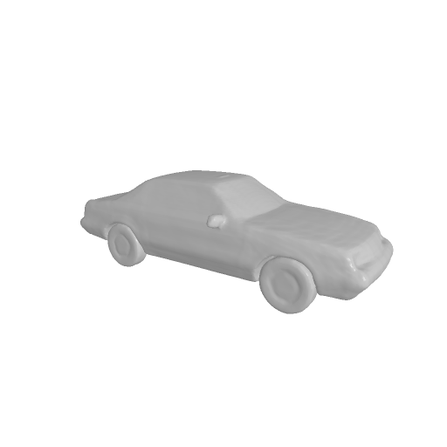}%
\includegraphics[width=0.155\textwidth]{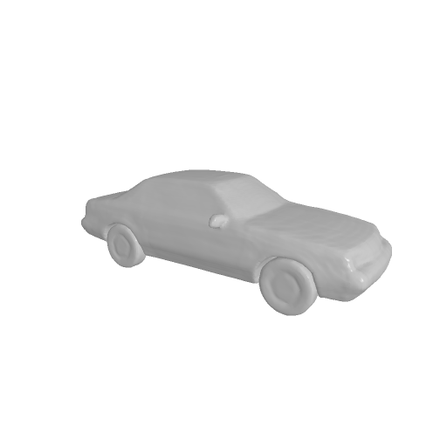}%
\includegraphics[width=0.155\textwidth]{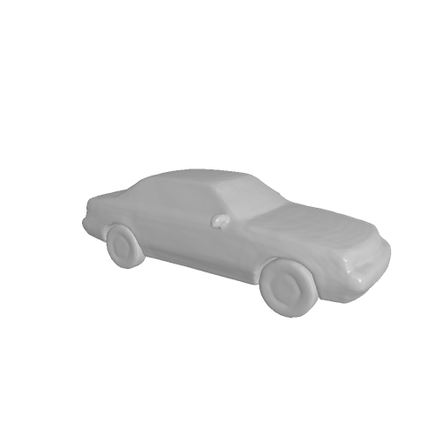}%
\includegraphics[width=0.155\textwidth]{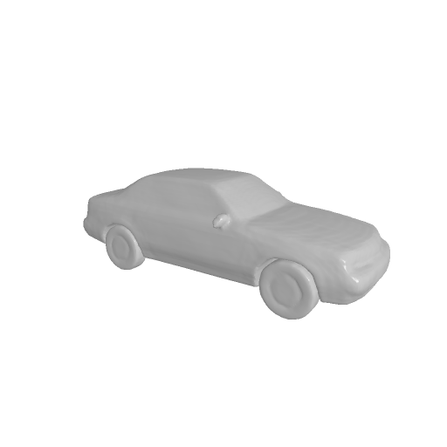}%
\includegraphics[width=0.155\textwidth]{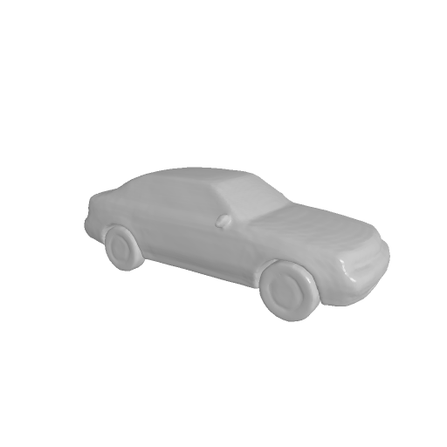}%
}\\[-0.15em]
\makebox[\textwidth][c]{%
\includegraphics[width=0.155\textwidth]{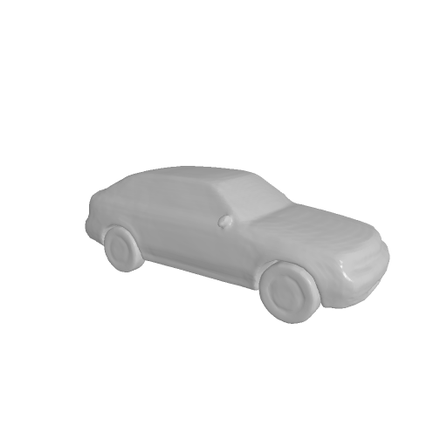}%
\includegraphics[width=0.155\textwidth]{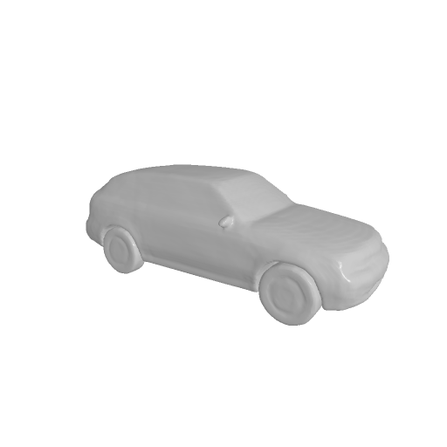}%
\includegraphics[width=0.155\textwidth]{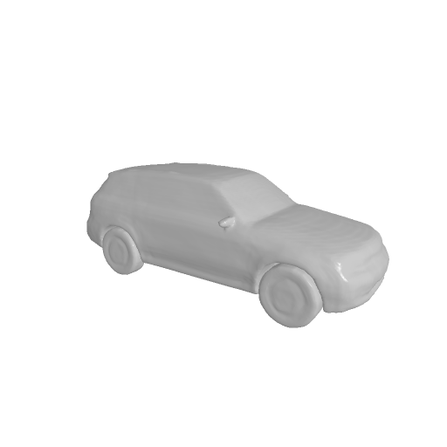}%
\includegraphics[width=0.155\textwidth]{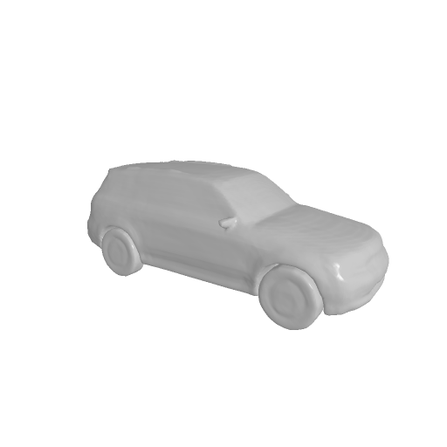}%
\includegraphics[width=0.155\textwidth]{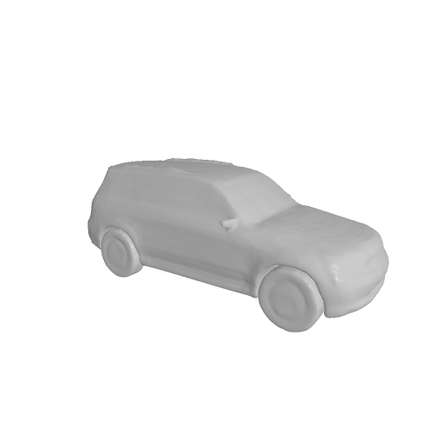}%
}\\[0.55em]
    {\small\textbf{Curriculum Style DeepSDF}}\\[-0.15em]
\makebox[\textwidth][c]{%
\includegraphics[width=0.155\textwidth]{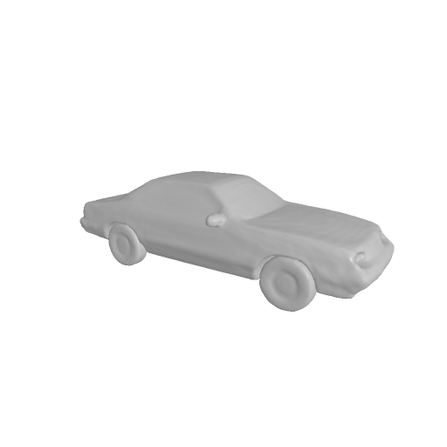}%
\includegraphics[width=0.155\textwidth]{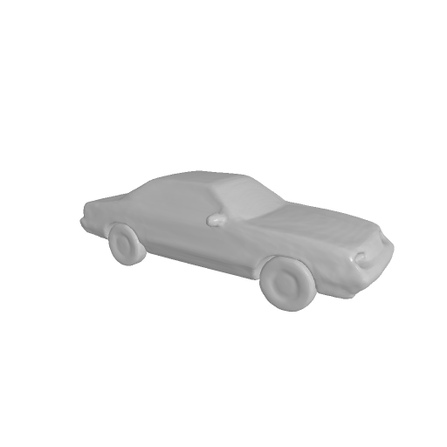}%
\includegraphics[width=0.155\textwidth]{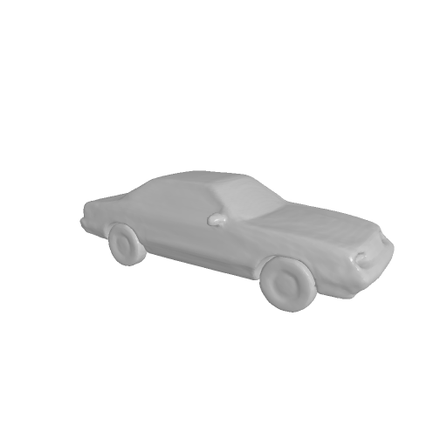}%
\includegraphics[width=0.155\textwidth]{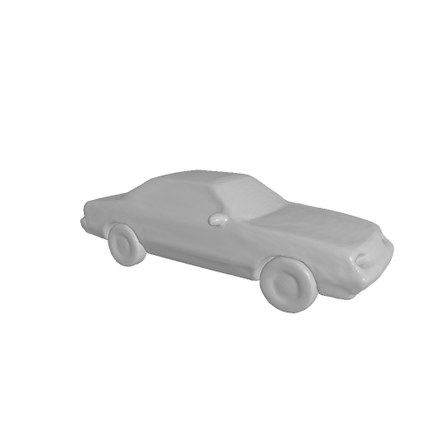}%
\includegraphics[width=0.155\textwidth]{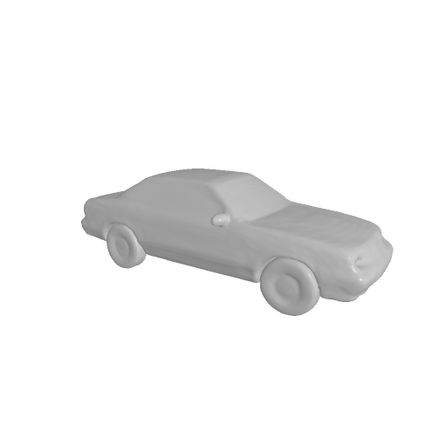}%
\includegraphics[width=0.155\textwidth]{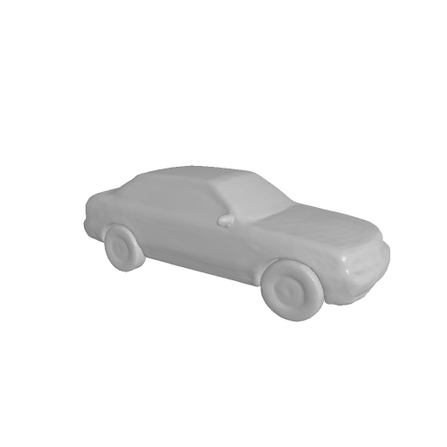}%
}\\[-0.15em]
\makebox[\textwidth][c]{%
\includegraphics[width=0.155\textwidth]{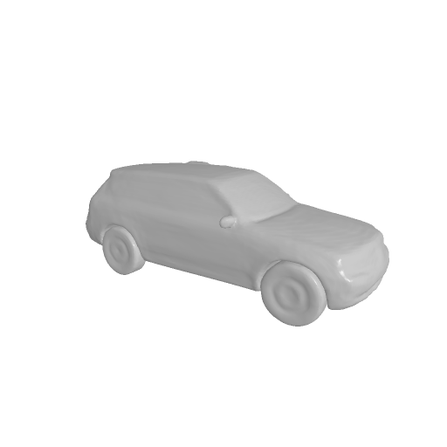}%
\includegraphics[width=0.155\textwidth]{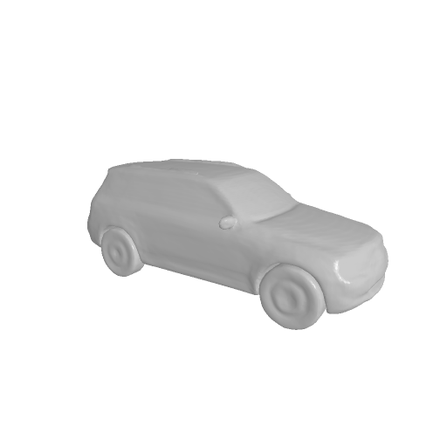}%
\includegraphics[width=0.155\textwidth]{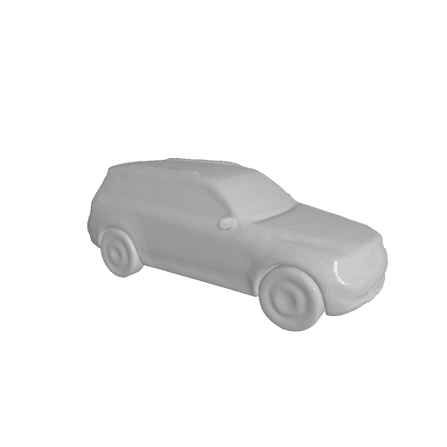}%
\includegraphics[width=0.155\textwidth]{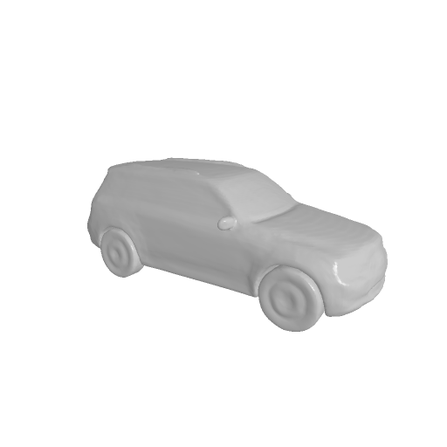}%
\includegraphics[width=0.155\textwidth]{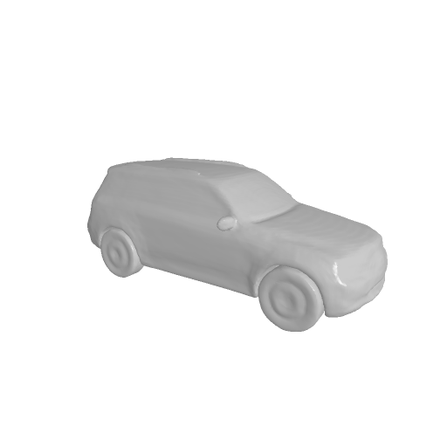}%
}\\[0.55em]
    {\small\textbf{StyleFields}}\\[-0.15em]
\makebox[\textwidth][c]{%
\includegraphics[width=0.155\textwidth]{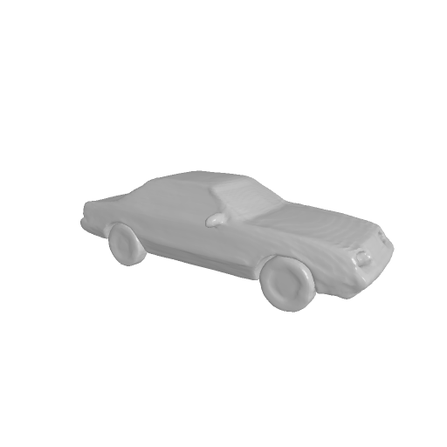}%
\includegraphics[width=0.155\textwidth]{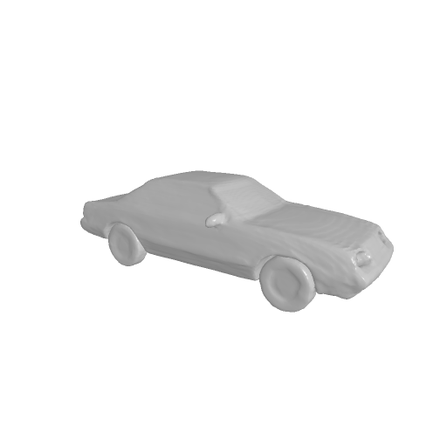}%
\includegraphics[width=0.155\textwidth]{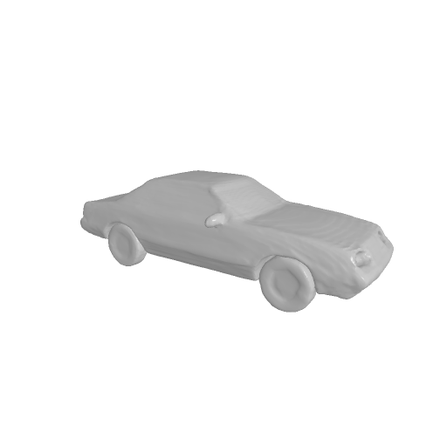}%
\includegraphics[width=0.155\textwidth]{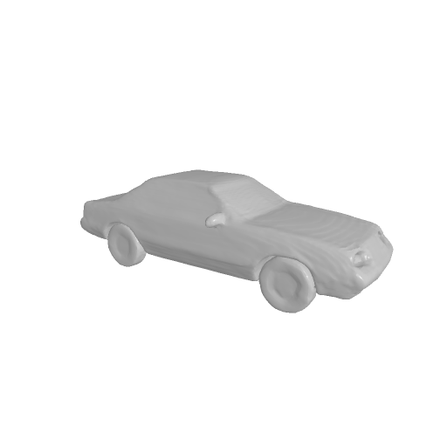}%
\includegraphics[width=0.155\textwidth]{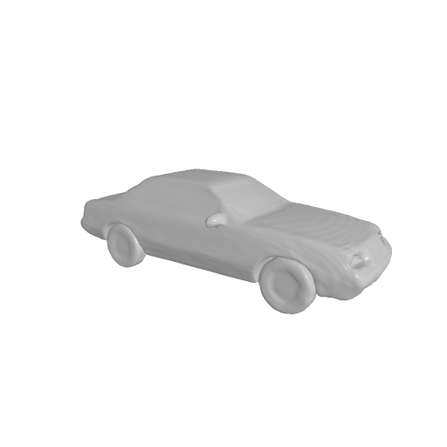}%
\includegraphics[width=0.155\textwidth]{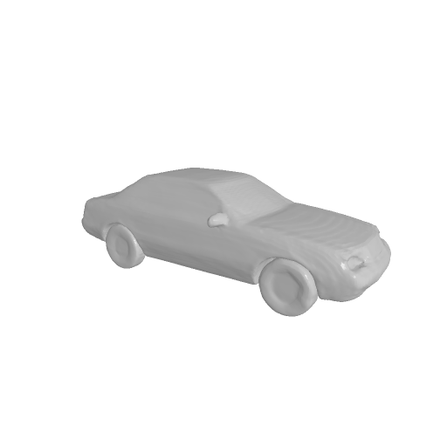}%
}\\[-0.15em]
\makebox[\textwidth][c]{%
\includegraphics[width=0.155\textwidth]{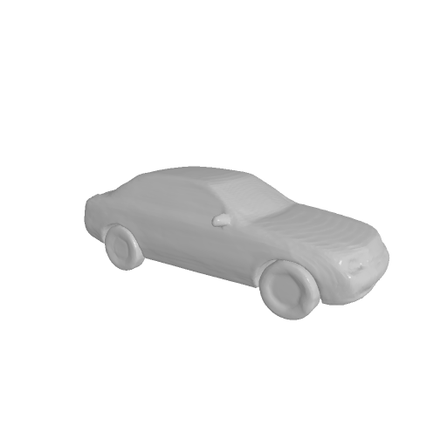}%
\includegraphics[width=0.155\textwidth]{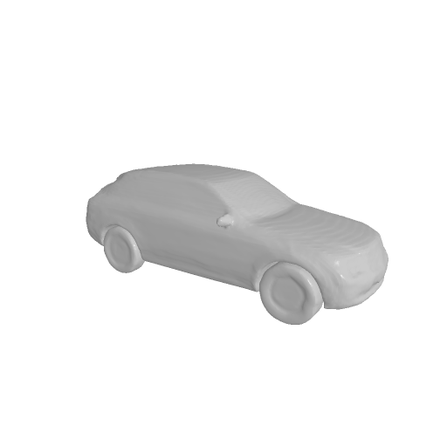}%
\includegraphics[width=0.155\textwidth]{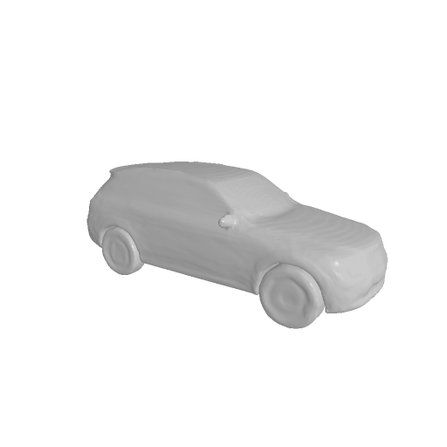}%
\includegraphics[width=0.155\textwidth]{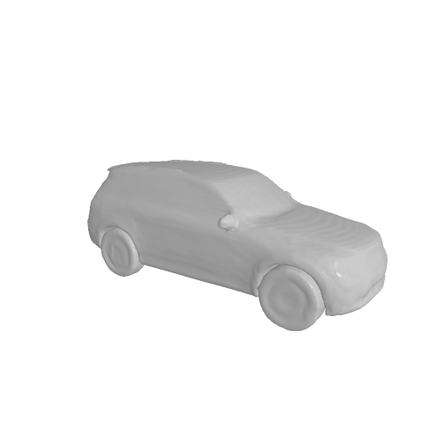}%
\includegraphics[width=0.155\textwidth]{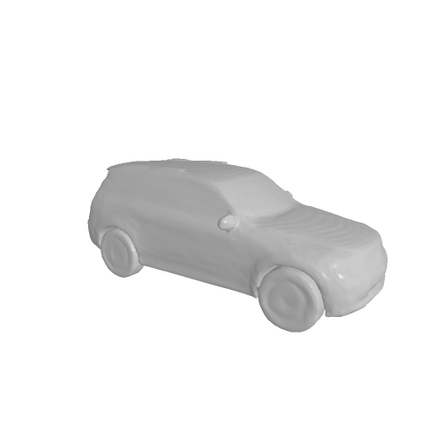}%
}\\[0.55em]
    \caption{Dense global-latent interpolation for cars (example 1). Each method is shown with interpolation coefficients \(t \in \{0,0.1,\dots,1.0\}\). The denser sampling makes clear that interpolation through a single global latent code changes coarse structure and fine detail simultaneously, producing averaged intermediate geometries rather than disentangled coarse-to-fine transitions.}
    \label{fig:supp_dense_interp_cars1}
\end{figure*}

\begin{figure*}[p]
    \centering
    {\small\textbf{DeepSDF}}\\[-0.15em]
\makebox[\textwidth][c]{%
\includegraphics[width=0.155\textwidth]{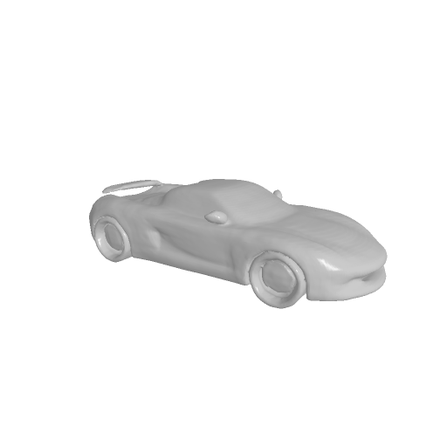}%
\includegraphics[width=0.155\textwidth]{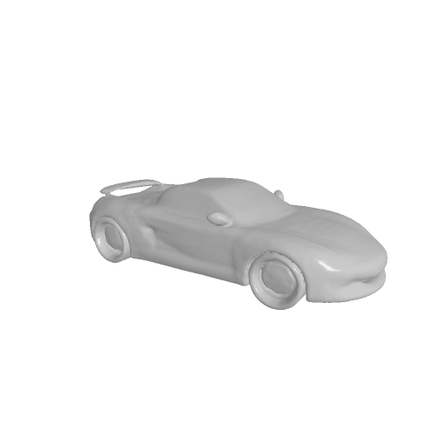}%
\includegraphics[width=0.155\textwidth]{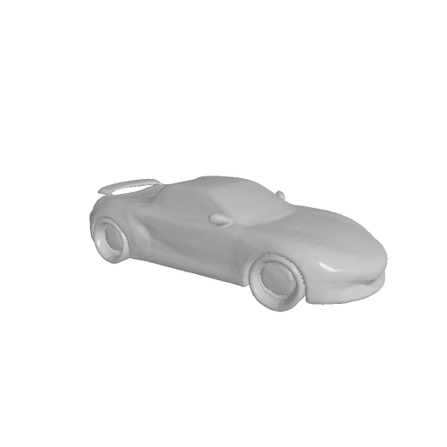}%
\includegraphics[width=0.155\textwidth]{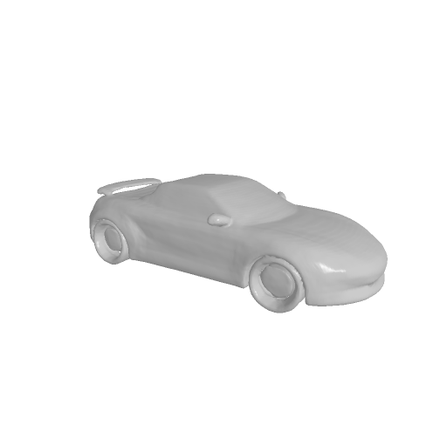}%
\includegraphics[width=0.155\textwidth]{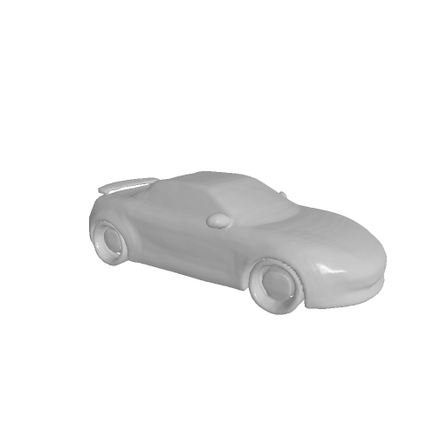}%
\includegraphics[width=0.155\textwidth]{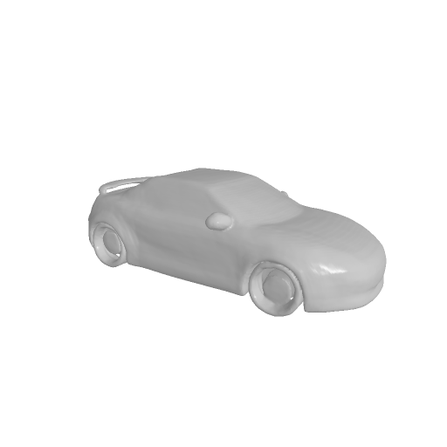}%
}\\[-0.15em]
\makebox[\textwidth][c]{%
\includegraphics[width=0.155\textwidth]{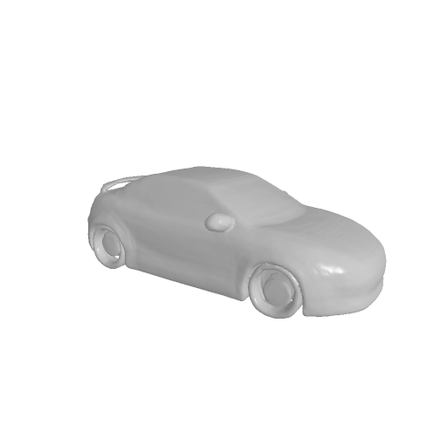}%
\includegraphics[width=0.155\textwidth]{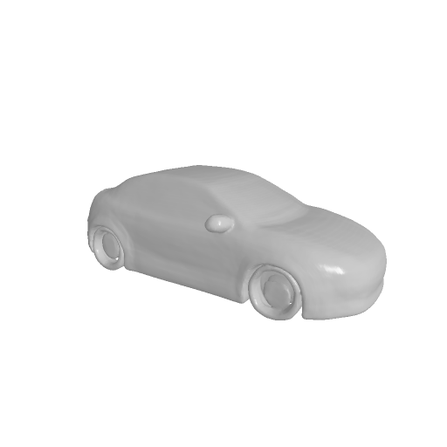}%
\includegraphics[width=0.155\textwidth]{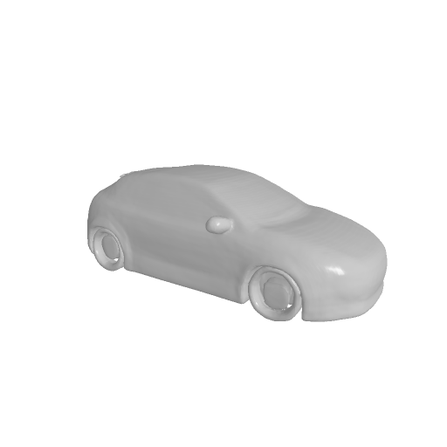}%
\includegraphics[width=0.155\textwidth]{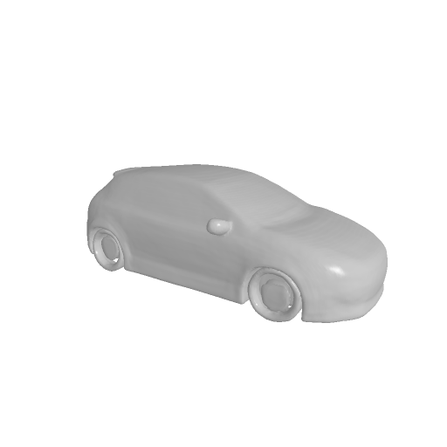}%
\includegraphics[width=0.155\textwidth]{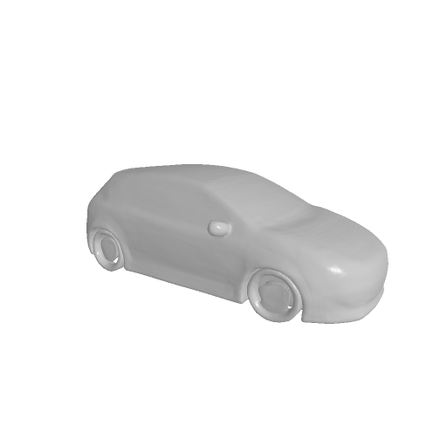}%
}\\[0.55em]
    {\small\textbf{Curriculum DeepSDF}}\\[-0.15em]
\makebox[\textwidth][c]{%
\includegraphics[width=0.155\textwidth]{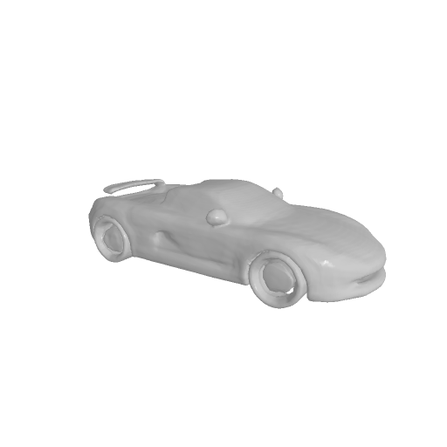}%
\includegraphics[width=0.155\textwidth]{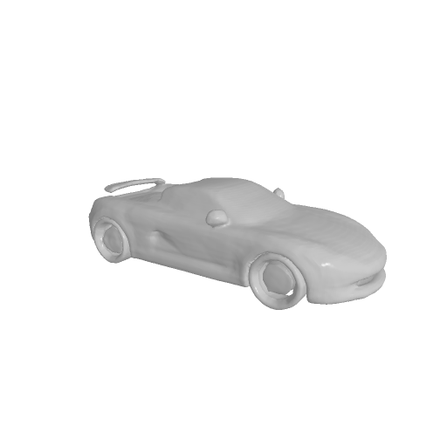}%
\includegraphics[width=0.155\textwidth]{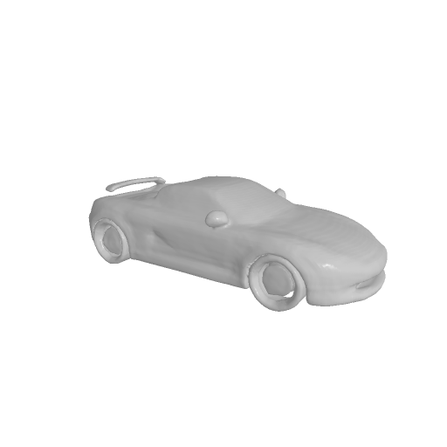}%
\includegraphics[width=0.155\textwidth]{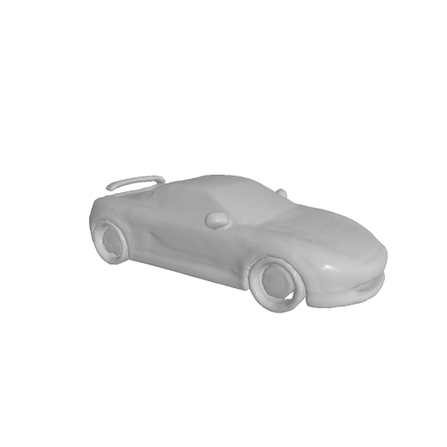}%
\includegraphics[width=0.155\textwidth]{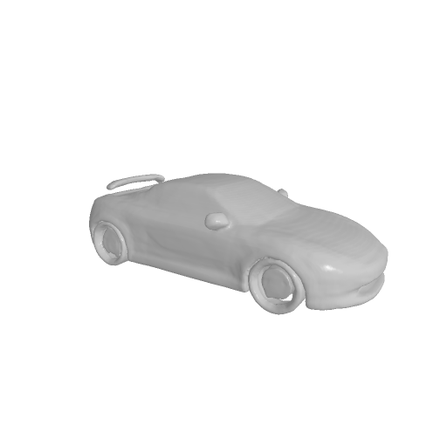}%
\includegraphics[width=0.155\textwidth]{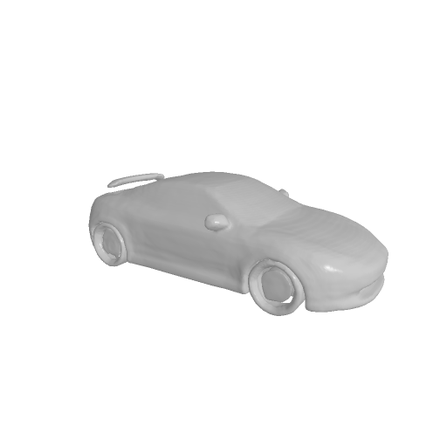}%
}\\[-0.15em]
\makebox[\textwidth][c]{%
\includegraphics[width=0.155\textwidth]{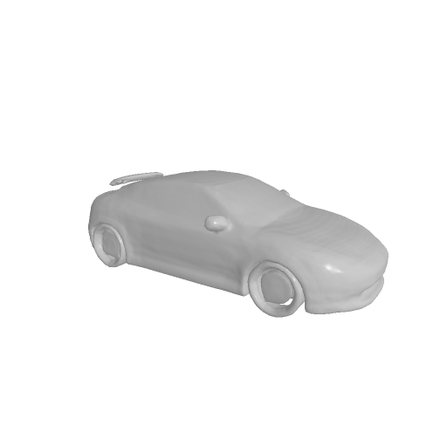}%
\includegraphics[width=0.155\textwidth]{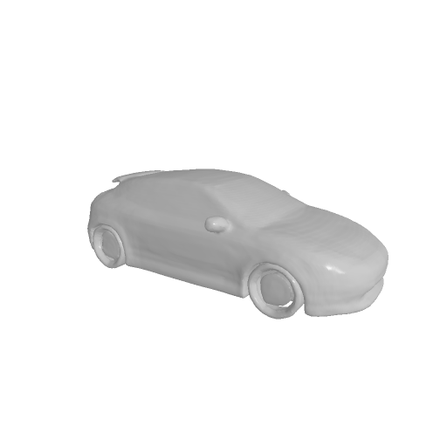}%
\includegraphics[width=0.155\textwidth]{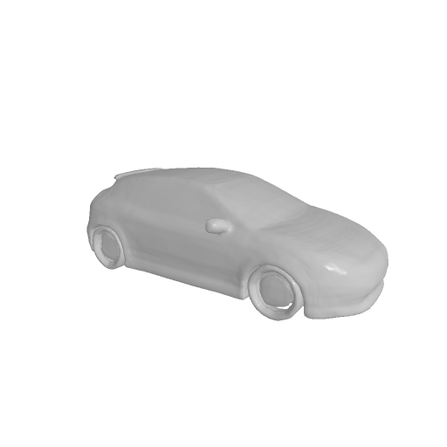}%
\includegraphics[width=0.155\textwidth]{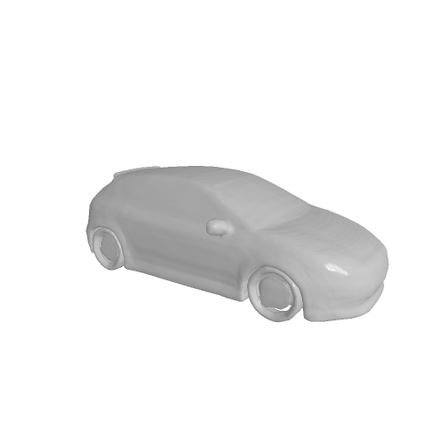}%
\includegraphics[width=0.155\textwidth]{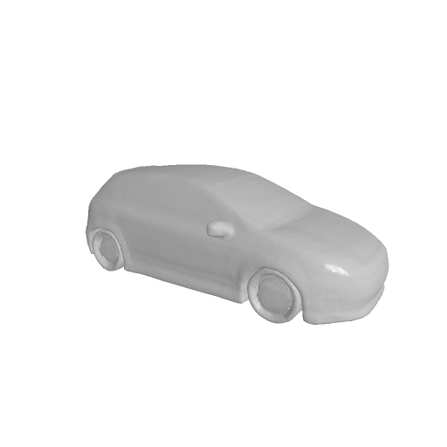}%
}\\[0.55em]
    {\small\textbf{Curriculum Style DeepSDF}}\\[-0.15em]
\makebox[\textwidth][c]{%
\includegraphics[width=0.155\textwidth]{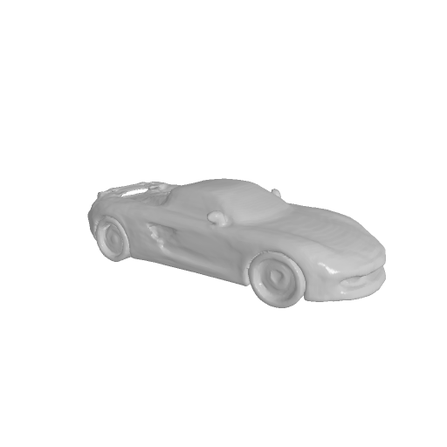}%
\includegraphics[width=0.155\textwidth]{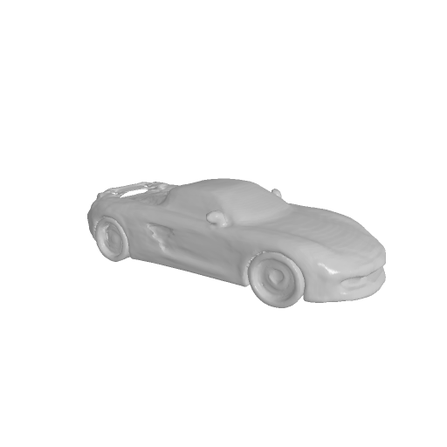}%
\includegraphics[width=0.155\textwidth]{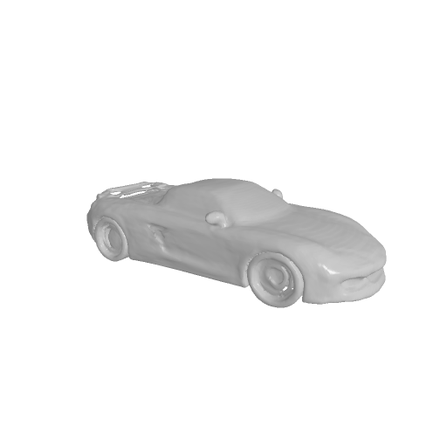}%
\includegraphics[width=0.155\textwidth]{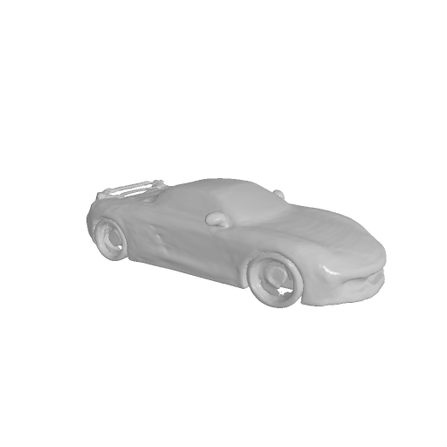}%
\includegraphics[width=0.155\textwidth]{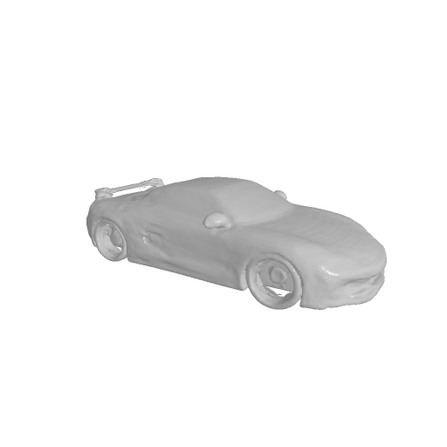}%
\includegraphics[width=0.155\textwidth]{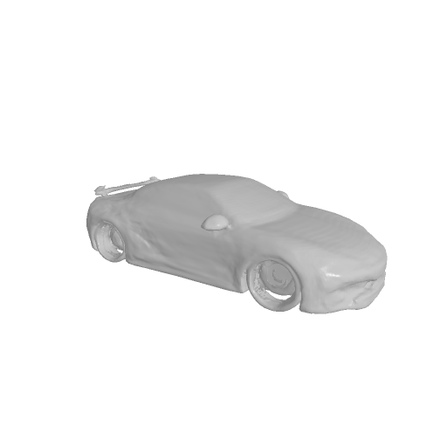}%
}\\[-0.15em]
\makebox[\textwidth][c]{%
\includegraphics[width=0.155\textwidth]{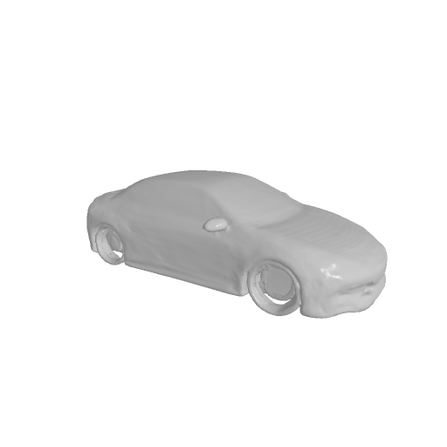}%
\includegraphics[width=0.155\textwidth]{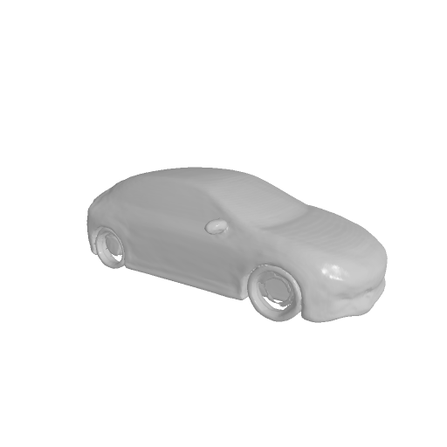}%
\includegraphics[width=0.155\textwidth]{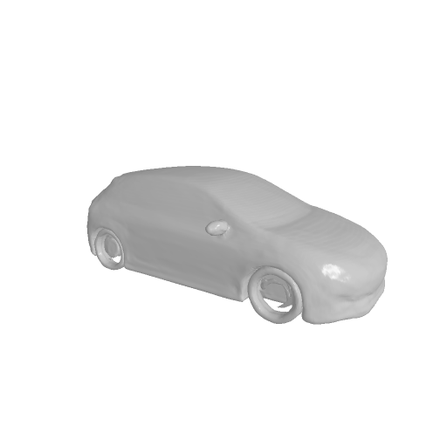}%
\includegraphics[width=0.155\textwidth]{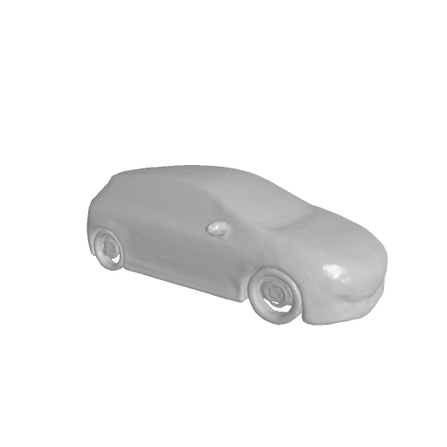}%
\includegraphics[width=0.155\textwidth]{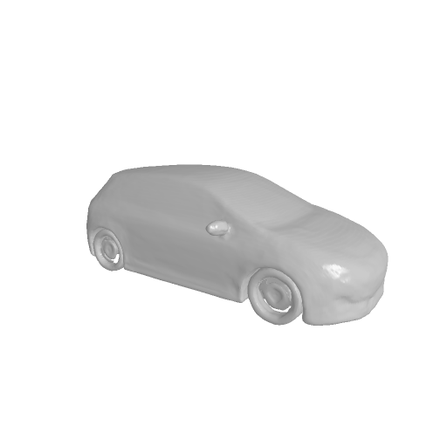}%
}\\[0.55em]
    {\small\textbf{StyleFields}}\\[-0.15em]
\makebox[\textwidth][c]{%
\includegraphics[width=0.155\textwidth]{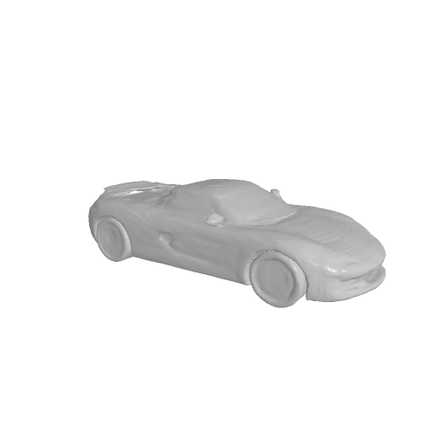}%
\includegraphics[width=0.155\textwidth]{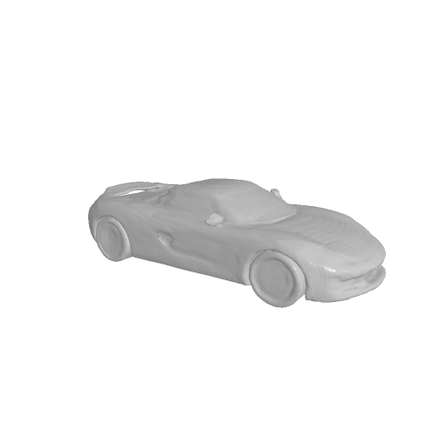}%
\includegraphics[width=0.155\textwidth]{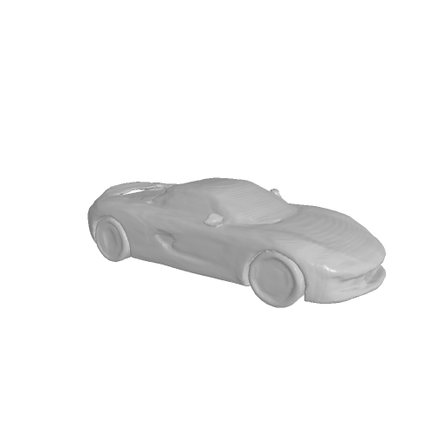}%
\includegraphics[width=0.155\textwidth]{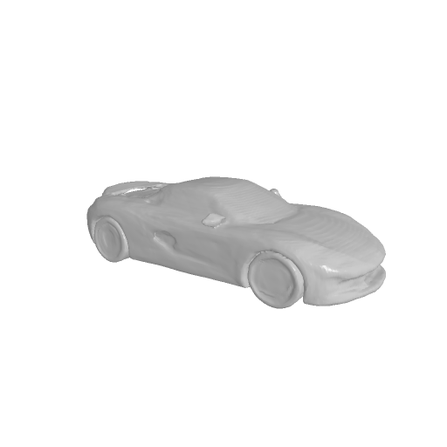}%
\includegraphics[width=0.155\textwidth]{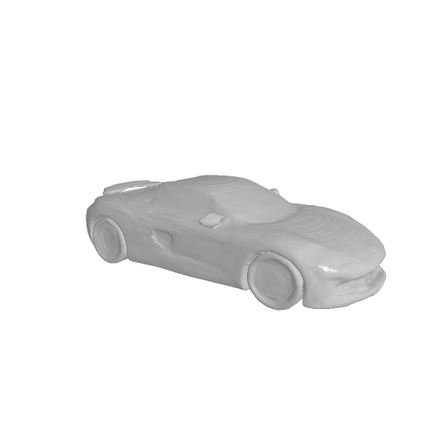}%
\includegraphics[width=0.155\textwidth]{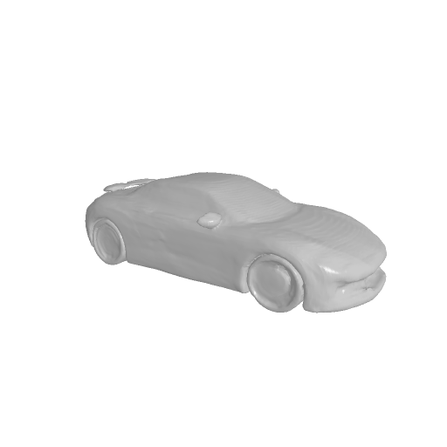}%
}\\[-0.15em]
\makebox[\textwidth][c]{%
\includegraphics[width=0.155\textwidth]{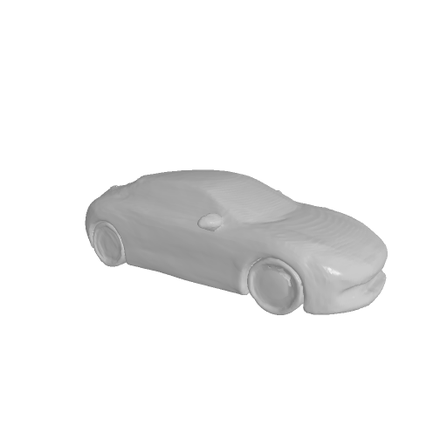}%
\includegraphics[width=0.155\textwidth]{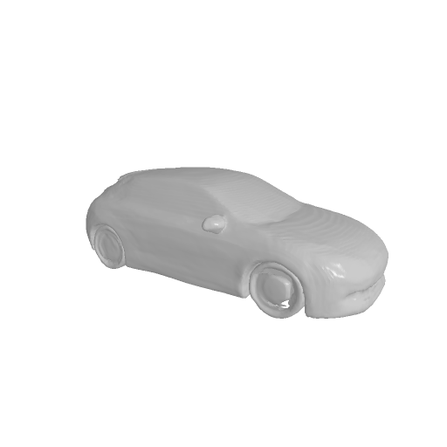}%
\includegraphics[width=0.155\textwidth]{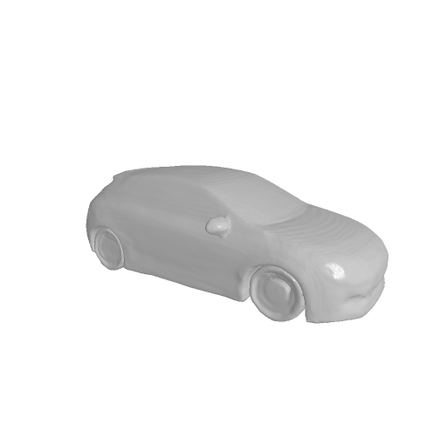}%
\includegraphics[width=0.155\textwidth]{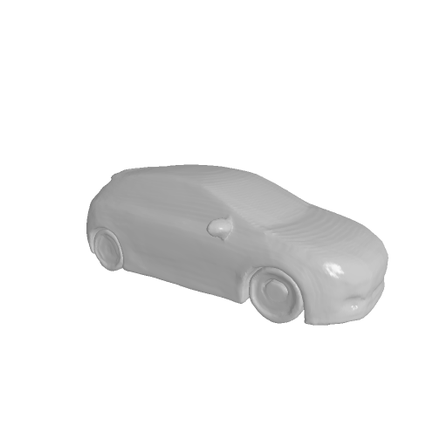}%
\includegraphics[width=0.155\textwidth]{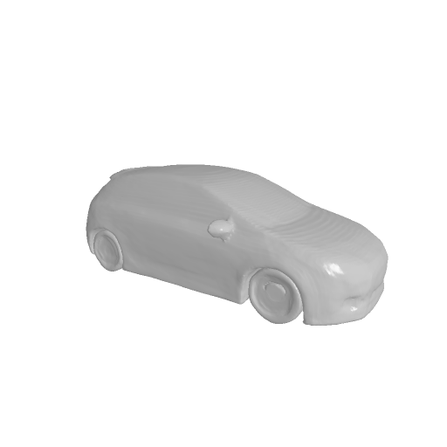}%
}\\[0.55em]
    \caption{Dense global-latent interpolation for cars (example 2). As in the main paper, none of the methods disentangle low-frequency and high-frequency geometry when interpolation is performed only in the global latent \(z\); both levels of structure evolve jointly along the path.}
    \label{fig:supp_dense_interp_cars2}
\end{figure*}

\begin{figure*}[p]
    \centering
    {\small\textbf{DeepSDF}}\\[-0.15em]
\makebox[\textwidth][c]{%
\includegraphics[width=0.155\textwidth]{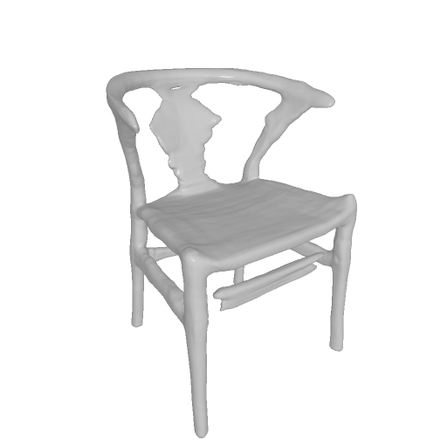}%
\includegraphics[width=0.155\textwidth]{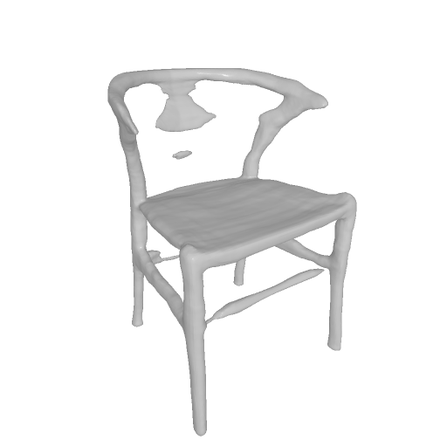}%
\includegraphics[width=0.155\textwidth]{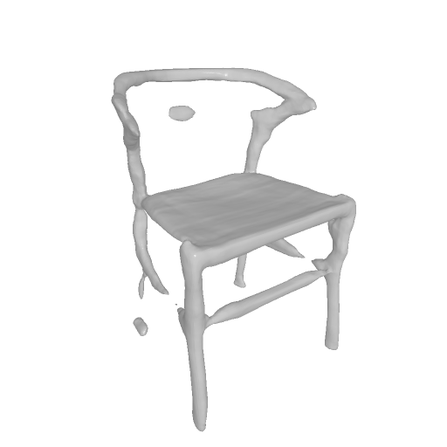}%
\includegraphics[width=0.155\textwidth]{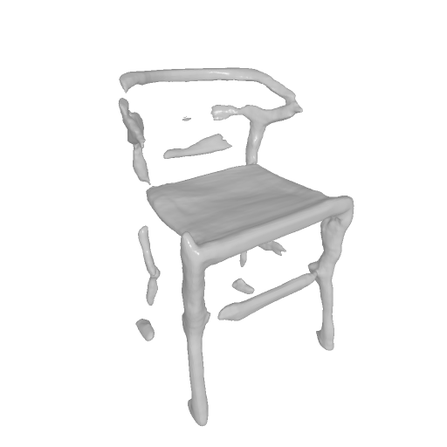}%
\includegraphics[width=0.155\textwidth]{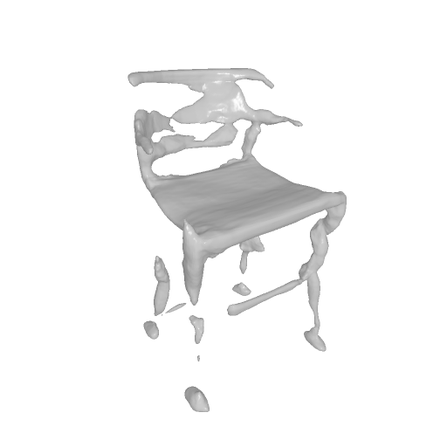}%
\includegraphics[width=0.155\textwidth]{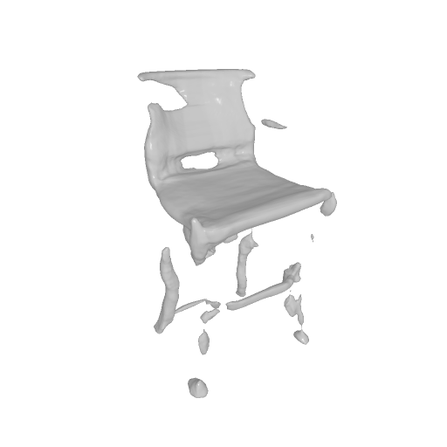}%
}\\[-0.15em]
\makebox[\textwidth][c]{%
\includegraphics[width=0.155\textwidth]{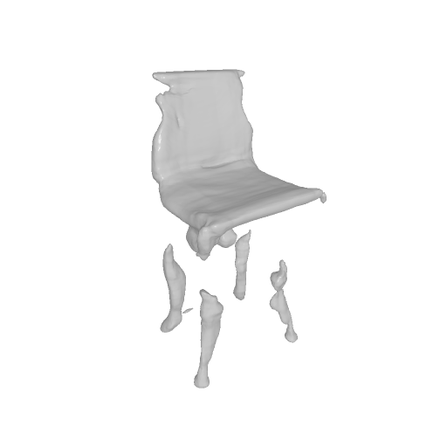}%
\includegraphics[width=0.155\textwidth]{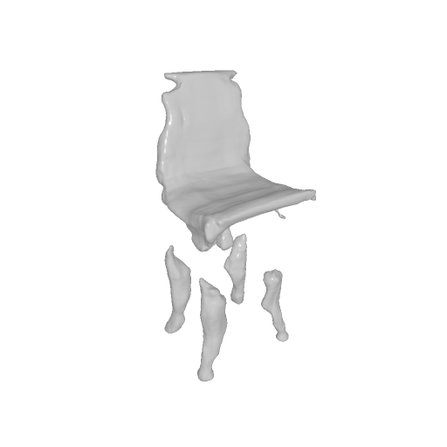}%
\includegraphics[width=0.155\textwidth]{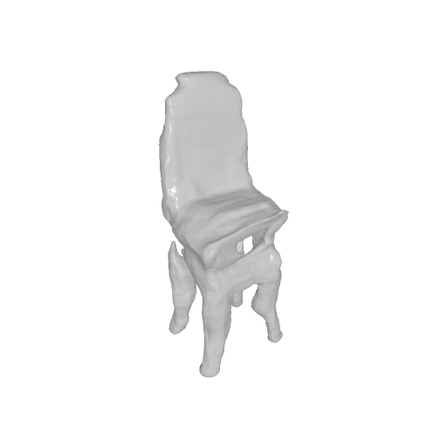}%
\includegraphics[width=0.155\textwidth]{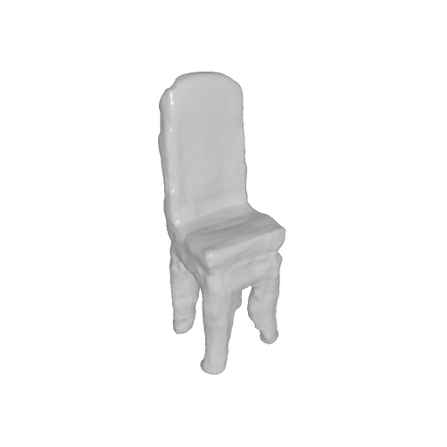}%
\includegraphics[width=0.155\textwidth]{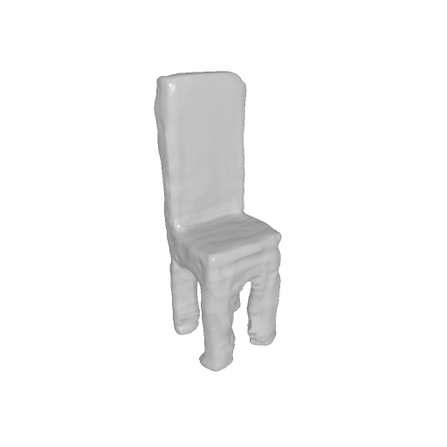}%
}\\[0.55em]
    {\small\textbf{Curriculum DeepSDF}}\\[-0.15em]
\makebox[\textwidth][c]{%
\includegraphics[width=0.155\textwidth]{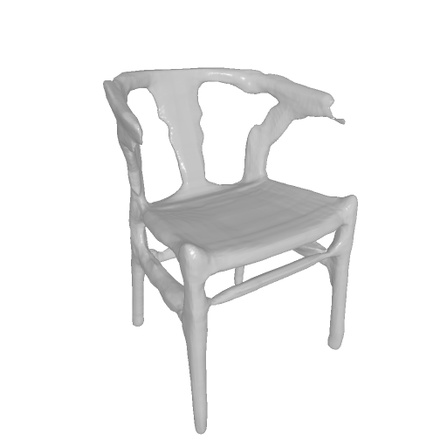}%
\includegraphics[width=0.155\textwidth]{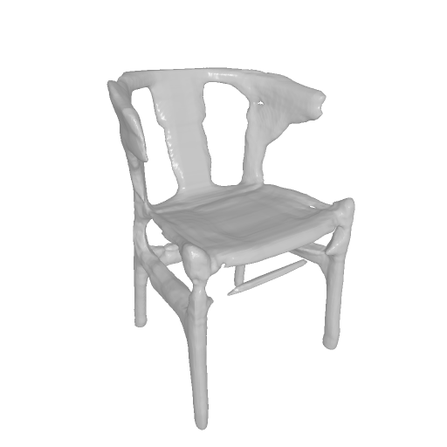}%
\includegraphics[width=0.155\textwidth]{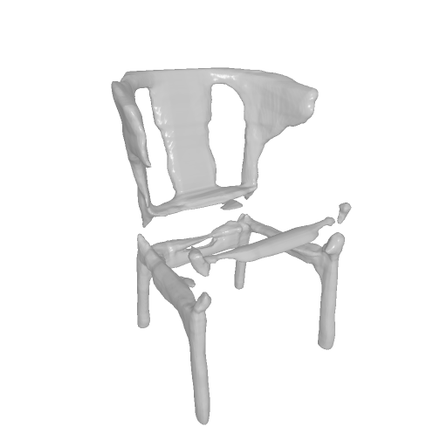}%
\includegraphics[width=0.155\textwidth]{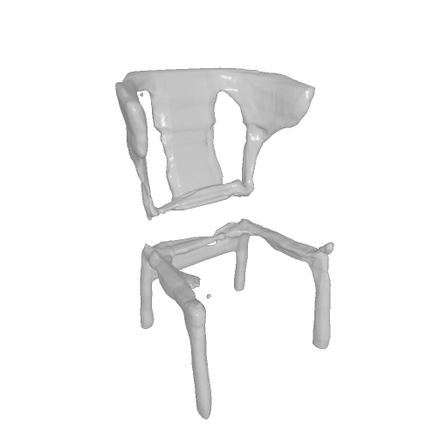}%
\includegraphics[width=0.155\textwidth]{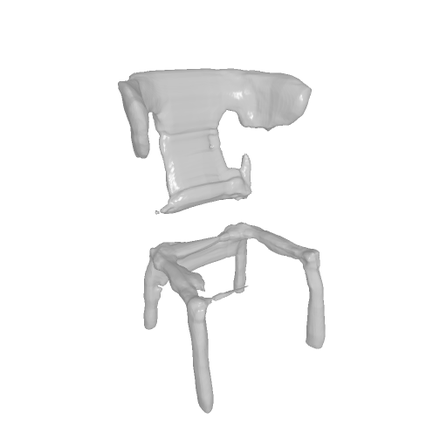}%
\includegraphics[width=0.155\textwidth]{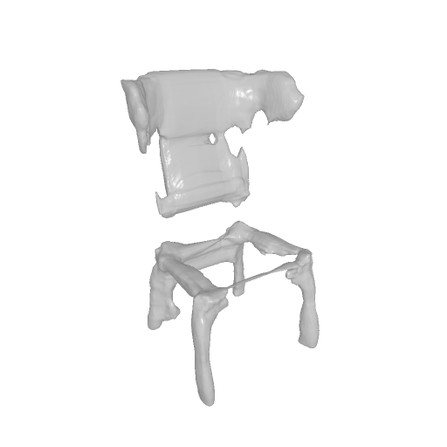}%
}\\[-0.15em]
\makebox[\textwidth][c]{%
\includegraphics[width=0.155\textwidth]{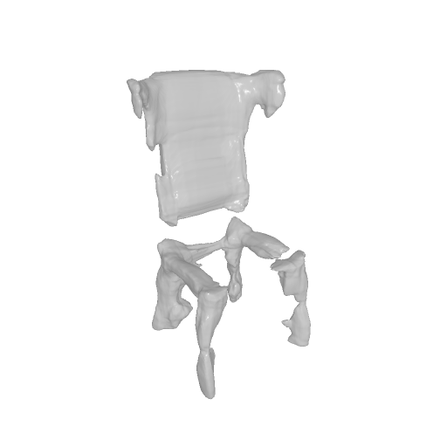}%
\includegraphics[width=0.155\textwidth]{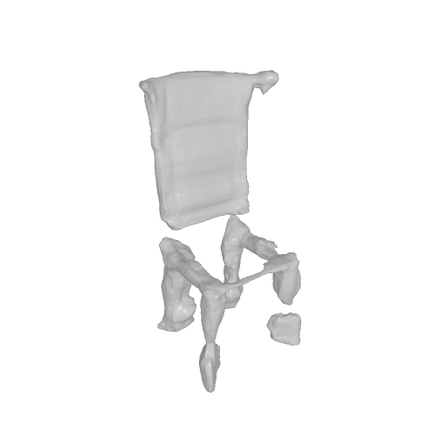}%
\includegraphics[width=0.155\textwidth]{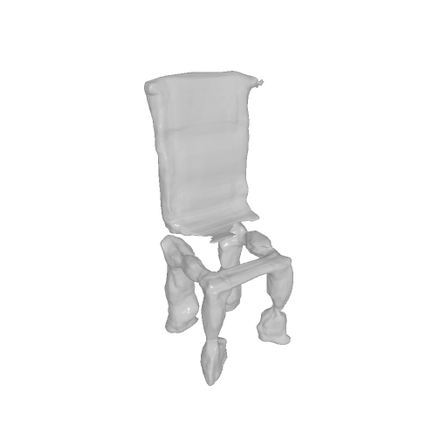}%
\includegraphics[width=0.155\textwidth]{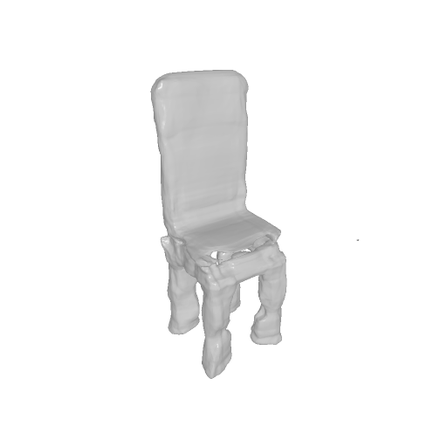}%
\includegraphics[width=0.155\textwidth]{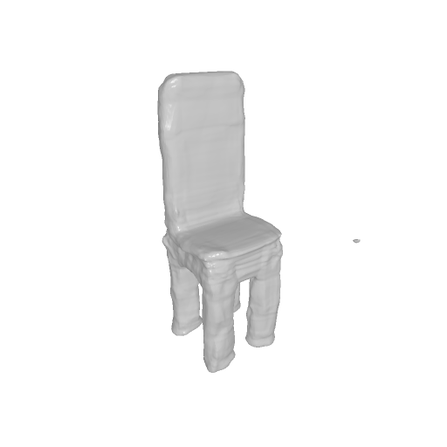}%
}\\[0.55em]
    {\small\textbf{Curriculum Style DeepSDF}}\\[-0.15em]
\makebox[\textwidth][c]{%
\includegraphics[width=0.155\textwidth]{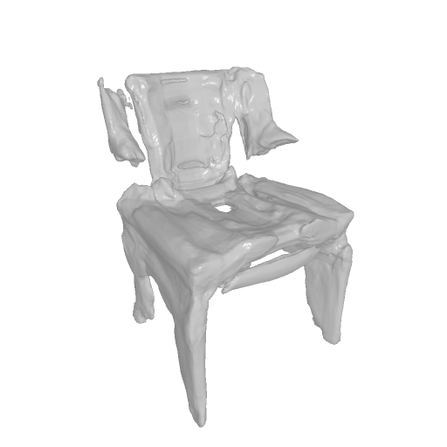}%
\includegraphics[width=0.155\textwidth]{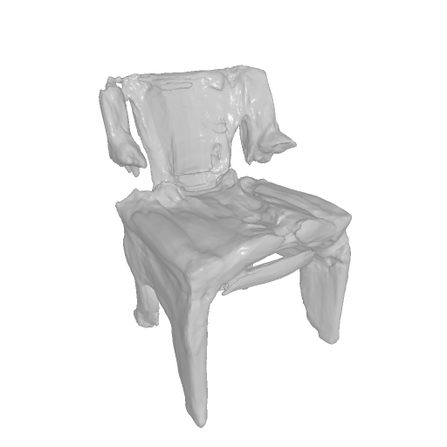}%
\includegraphics[width=0.155\textwidth]{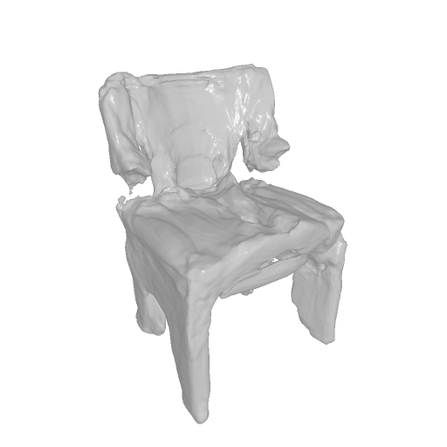}%
\includegraphics[width=0.155\textwidth]{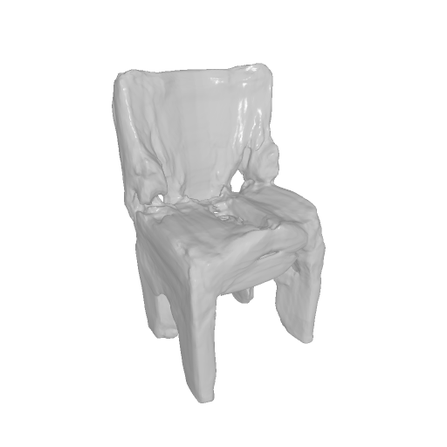}%
\includegraphics[width=0.155\textwidth]{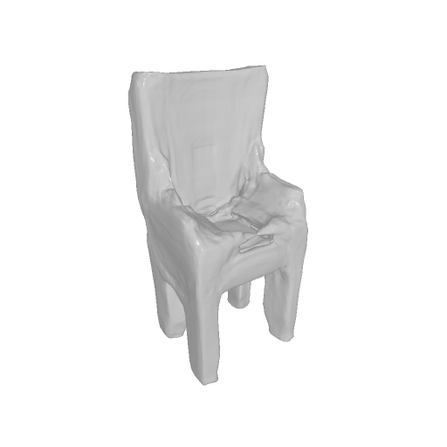}%
\includegraphics[width=0.155\textwidth]{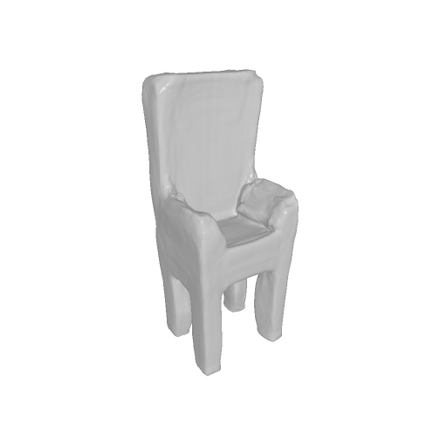}%
}\\[-0.15em]
\makebox[\textwidth][c]{%
\includegraphics[width=0.155\textwidth]{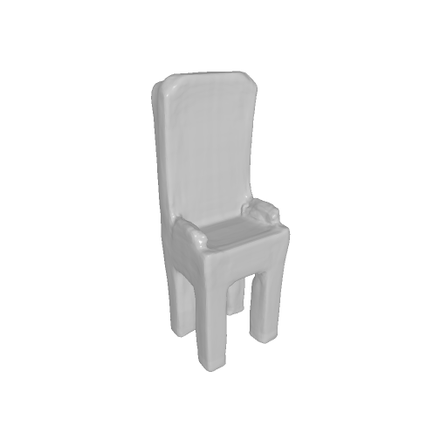}%
\includegraphics[width=0.155\textwidth]{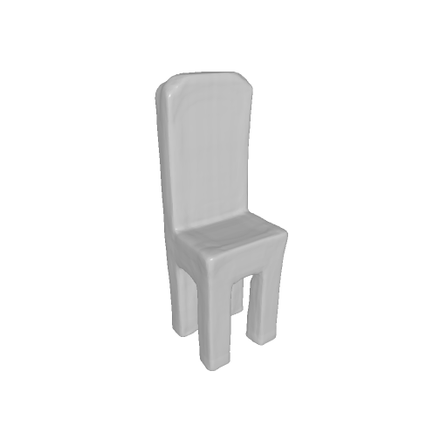}%
\includegraphics[width=0.155\textwidth]{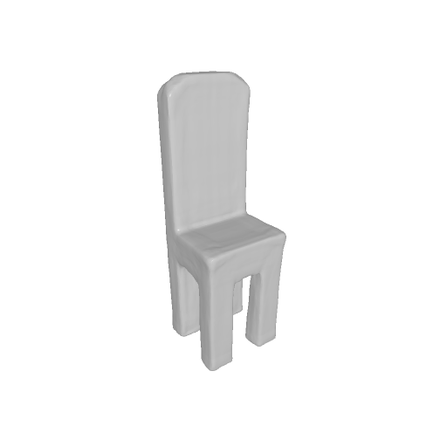}%
\includegraphics[width=0.155\textwidth]{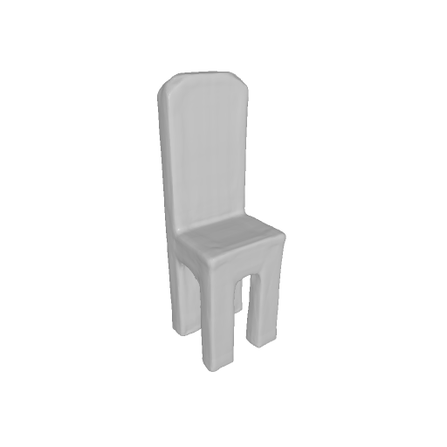}%
\includegraphics[width=0.155\textwidth]{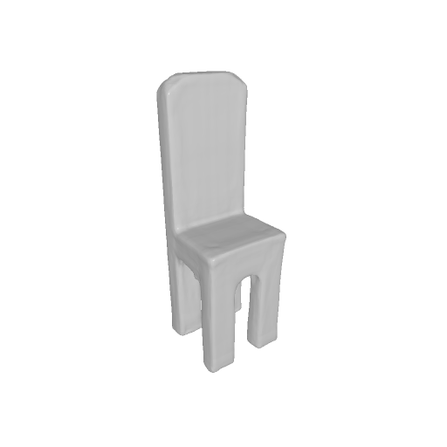}%
}\\[0.55em]
    {\small\textbf{StyleFields}}\\[-0.15em]
\makebox[\textwidth][c]{%
\includegraphics[width=0.155\textwidth]{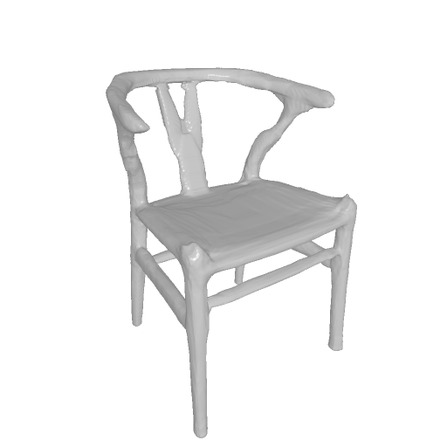}%
\includegraphics[width=0.155\textwidth]{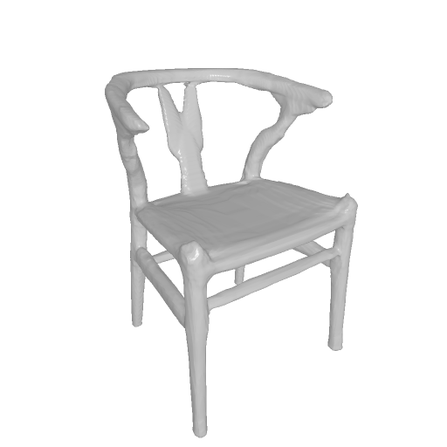}%
\includegraphics[width=0.155\textwidth]{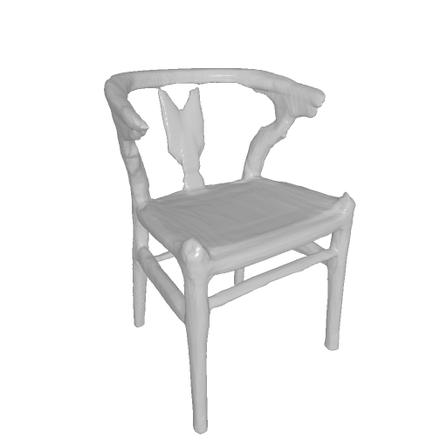}%
\includegraphics[width=0.155\textwidth]{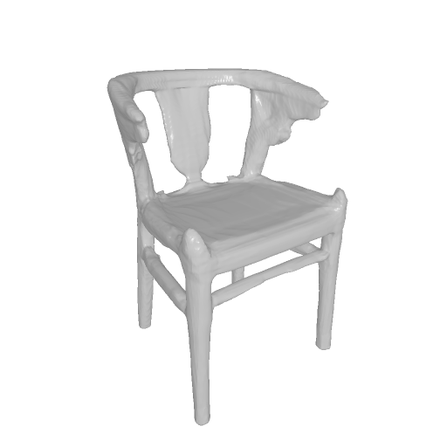}%
\includegraphics[width=0.155\textwidth]{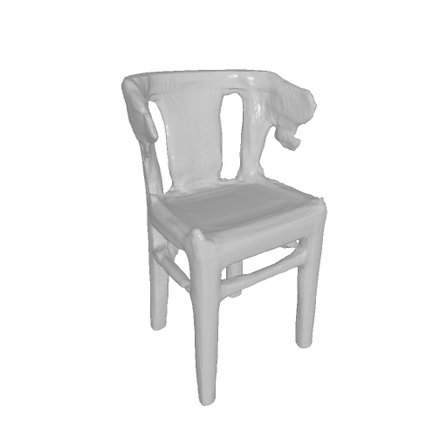}%
\includegraphics[width=0.155\textwidth]{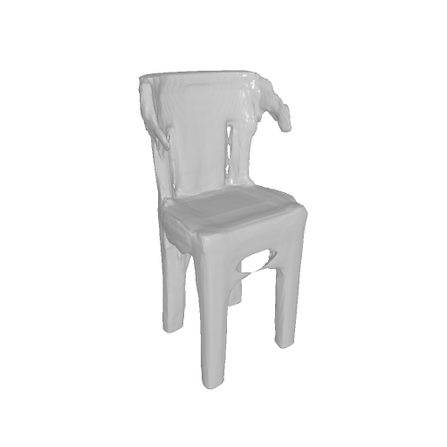}%
}\\[-0.15em]
\makebox[\textwidth][c]{%
\includegraphics[width=0.155\textwidth]{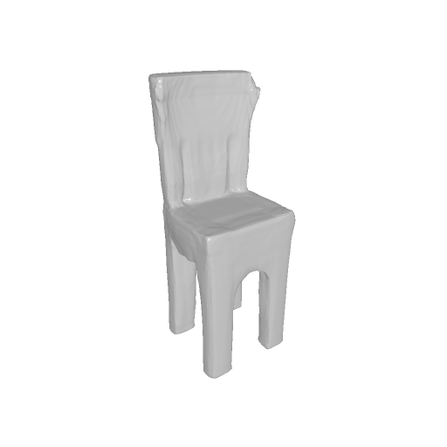}%
\includegraphics[width=0.155\textwidth]{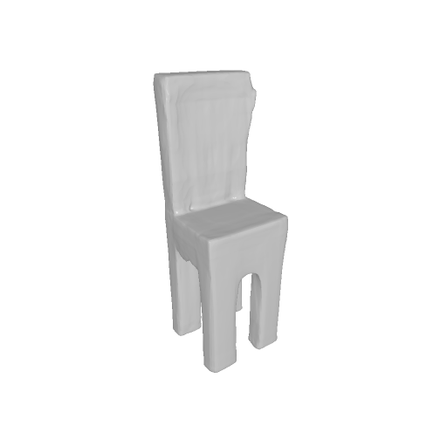}%
\includegraphics[width=0.155\textwidth]{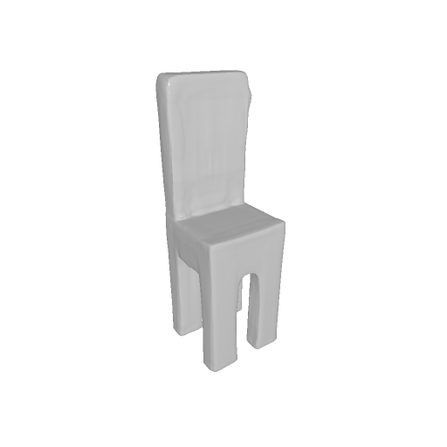}%
\includegraphics[width=0.155\textwidth]{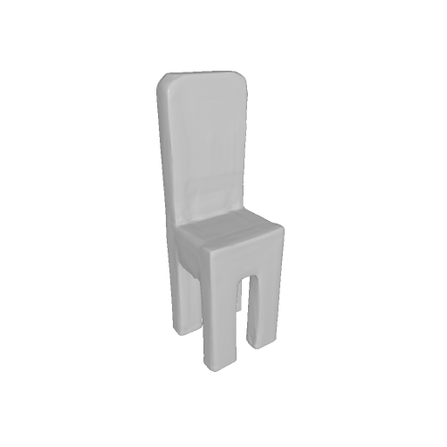}%
\includegraphics[width=0.155\textwidth]{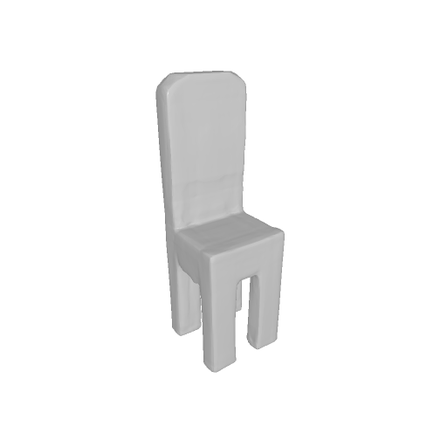}%
}\\[0.55em]
    \caption{Dense global-latent interpolation for chairs. While interpolation in the single global latent \(z\) remains entangled for all methods, StyleFields produces substantially cleaner and more structurally plausible intermediate chairs. In particular, seat, backrest, and leg geometry remain more coherent throughout the trajectory, whereas the baselines exhibit stronger degradation and instability in the intermediate shapes.}
    \label{fig:supp_dense_interp_chairs}
\end{figure*}

\begin{figure*}[p]
    \centering
    {\small\textbf{DeepSDF}}\\[-0.15em]
\makebox[\textwidth][c]{%
\includegraphics[width=0.155\textwidth]{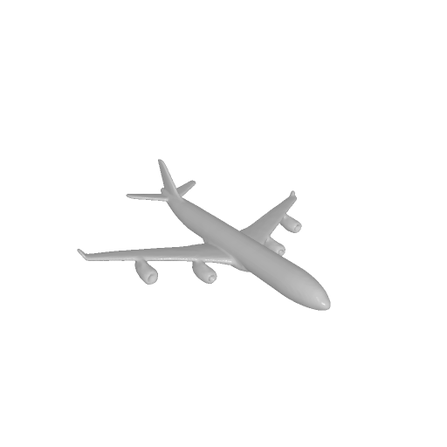}%
\includegraphics[width=0.155\textwidth]{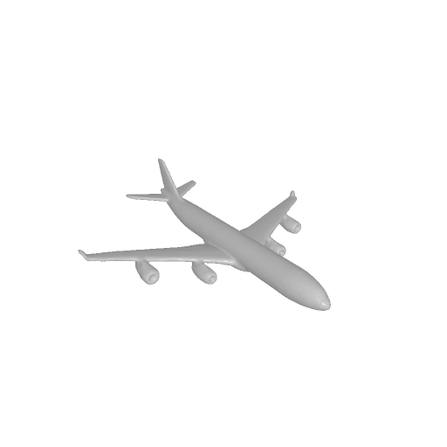}%
\includegraphics[width=0.155\textwidth]{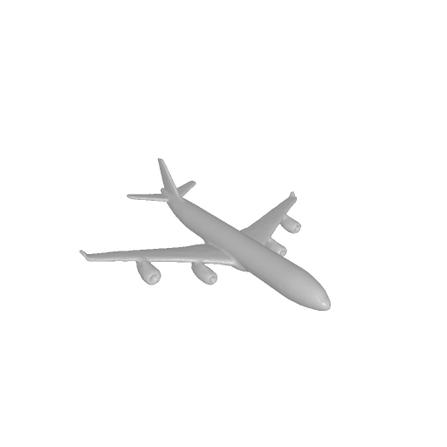}%
\includegraphics[width=0.155\textwidth]{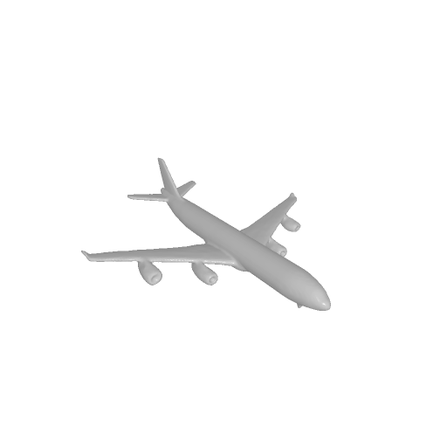}%
\includegraphics[width=0.155\textwidth]{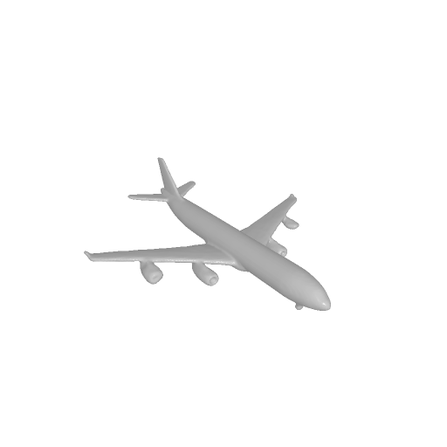}%
\includegraphics[width=0.155\textwidth]{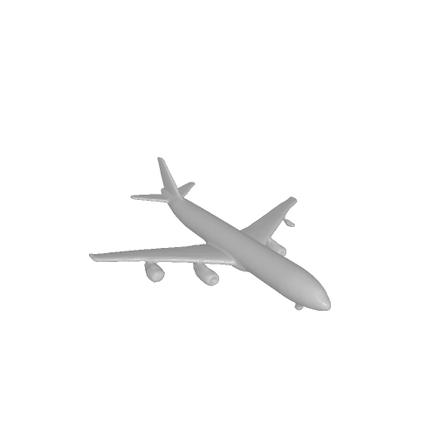}%
}\\[-0.15em]
\makebox[\textwidth][c]{%
\includegraphics[width=0.155\textwidth]{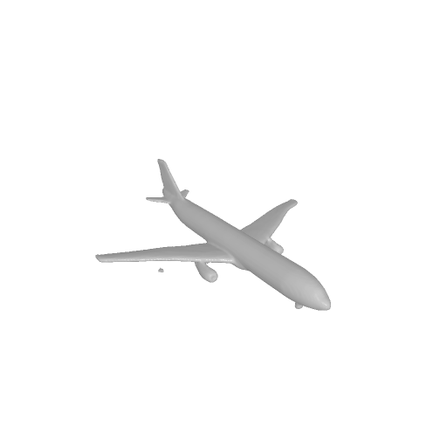}%
\includegraphics[width=0.155\textwidth]{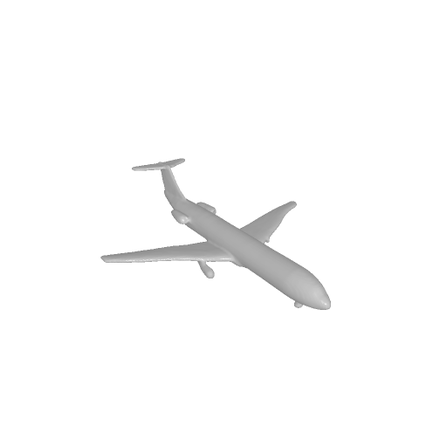}%
\includegraphics[width=0.155\textwidth]{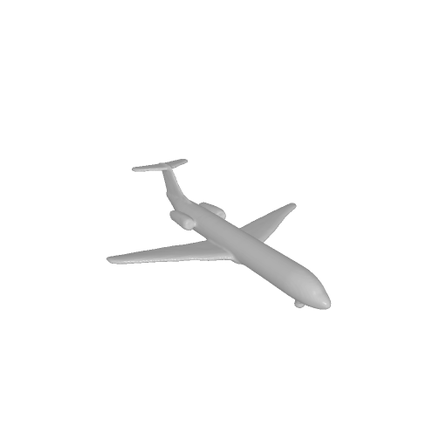}%
\includegraphics[width=0.155\textwidth]{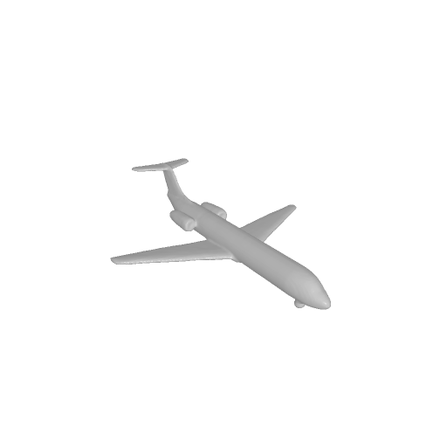}%
\includegraphics[width=0.155\textwidth]{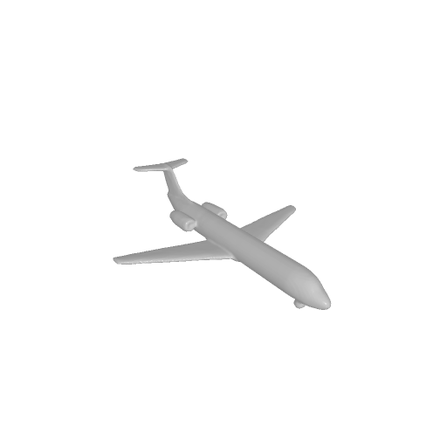}%
}\\[0.55em]
    {\small\textbf{Curriculum DeepSDF}}\\[-0.15em]
\makebox[\textwidth][c]{%
\includegraphics[width=0.155\textwidth]{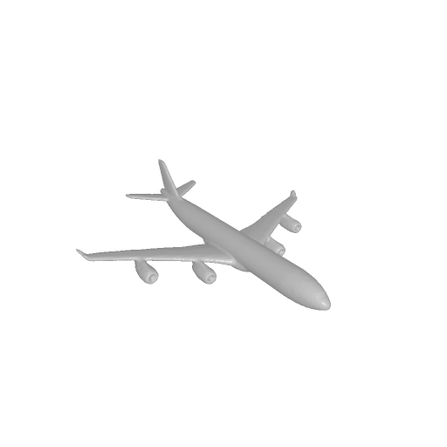}%
\includegraphics[width=0.155\textwidth]{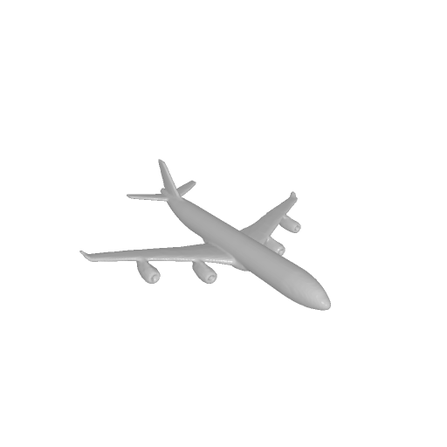}%
\includegraphics[width=0.155\textwidth]{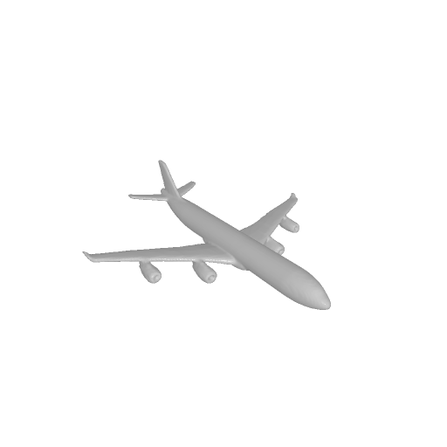}%
\includegraphics[width=0.155\textwidth]{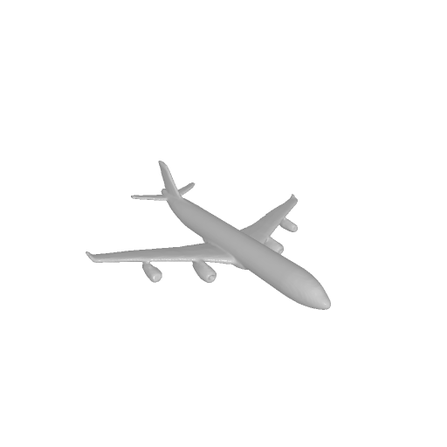}%
\includegraphics[width=0.155\textwidth]{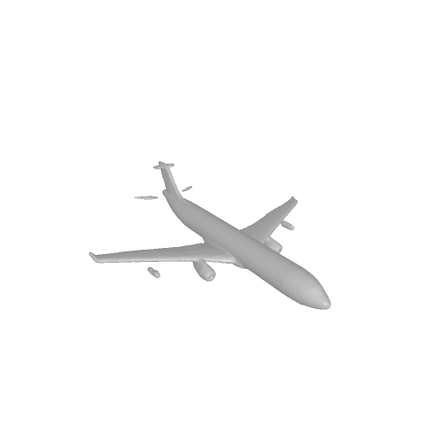}%
\includegraphics[width=0.155\textwidth]{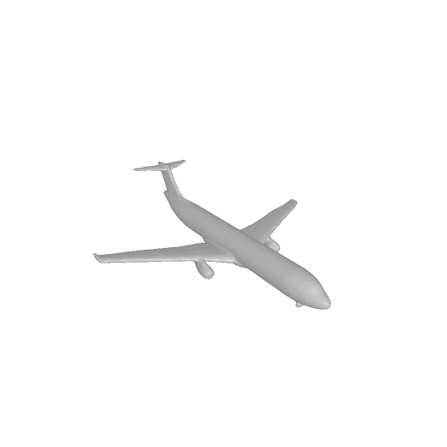}%
}\\[-0.15em]
\makebox[\textwidth][c]{%
\includegraphics[width=0.155\textwidth]{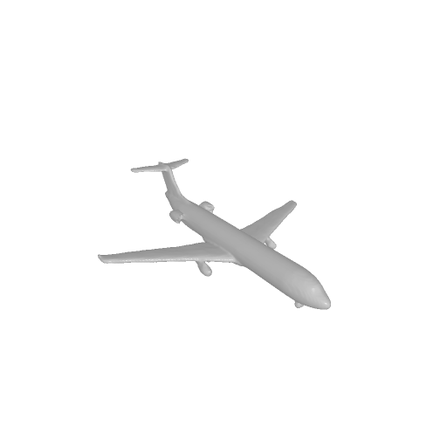}%
\includegraphics[width=0.155\textwidth]{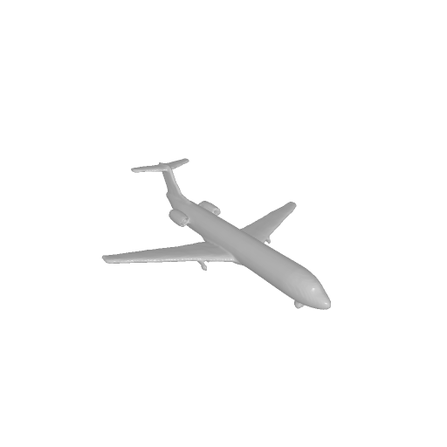}%
\includegraphics[width=0.155\textwidth]{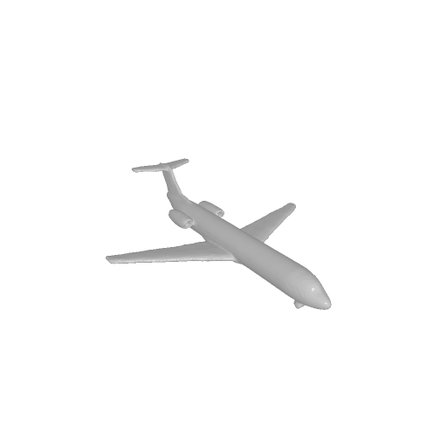}%
\includegraphics[width=0.155\textwidth]{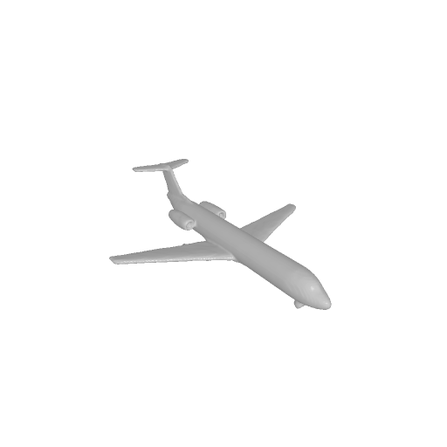}%
\includegraphics[width=0.155\textwidth]{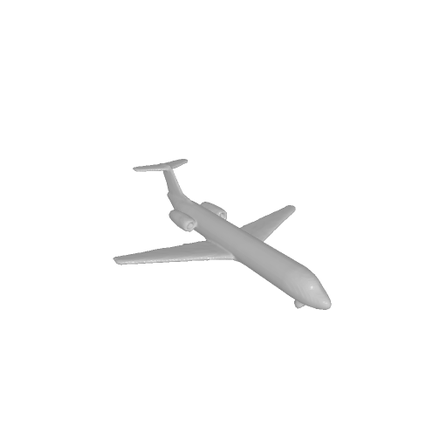}%
}\\[0.55em]
    {\small\textbf{Curriculum Style DeepSDF}}\\[-0.15em]
\makebox[\textwidth][c]{%
\includegraphics[width=0.155\textwidth]{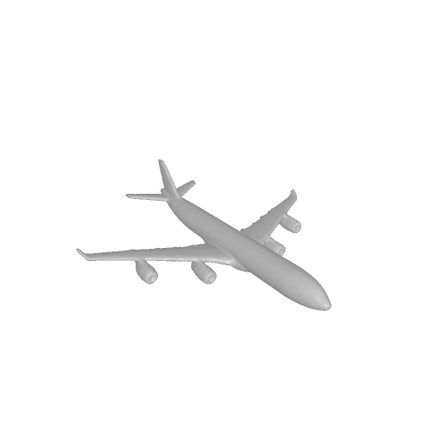}%
\includegraphics[width=0.155\textwidth]{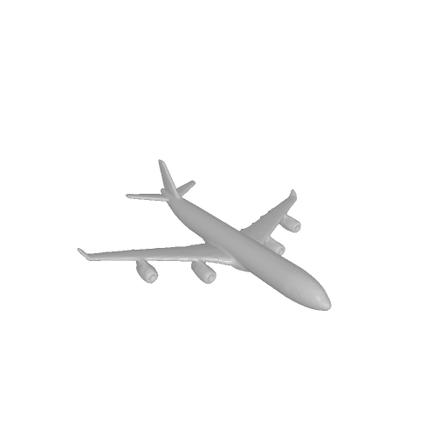}%
\includegraphics[width=0.155\textwidth]{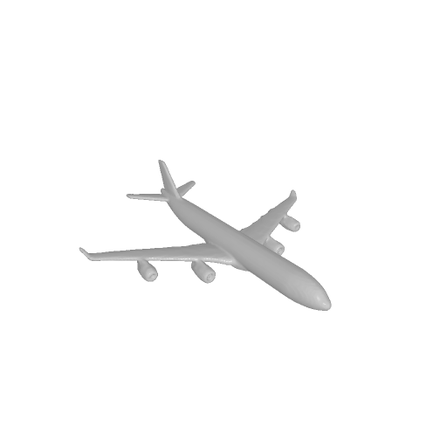}%
\includegraphics[width=0.155\textwidth]{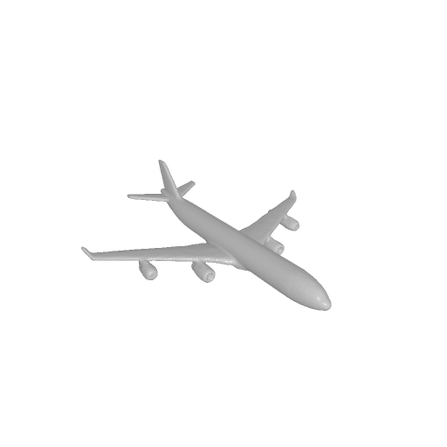}%
\includegraphics[width=0.155\textwidth]{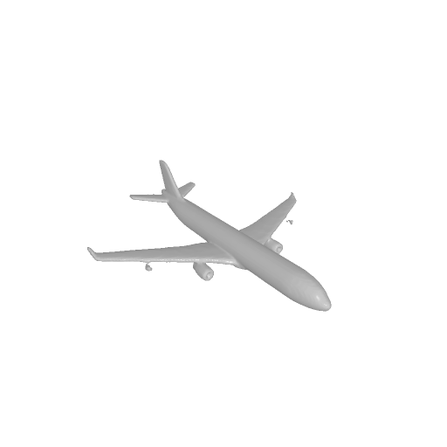}%
\includegraphics[width=0.155\textwidth]{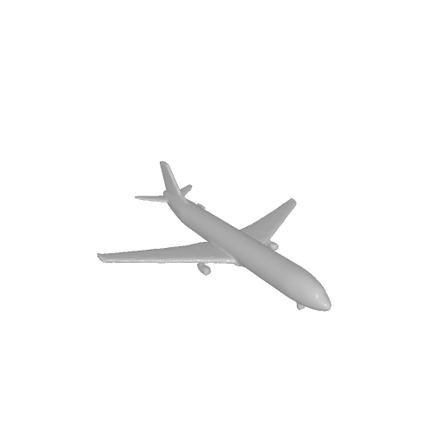}%
}\\[-0.15em]
\makebox[\textwidth][c]{%
\includegraphics[width=0.155\textwidth]{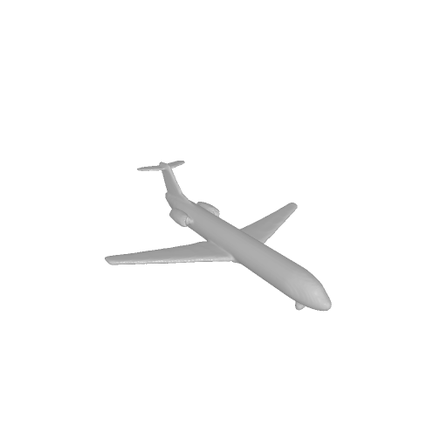}%
\includegraphics[width=0.155\textwidth]{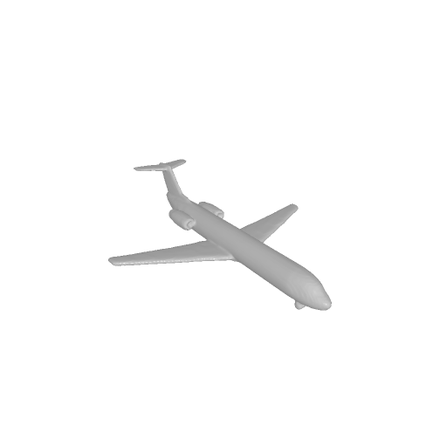}%
\includegraphics[width=0.155\textwidth]{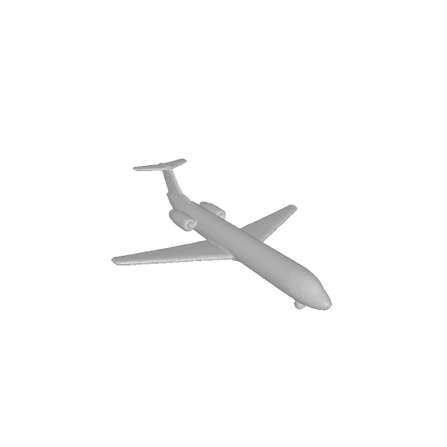}%
\includegraphics[width=0.155\textwidth]{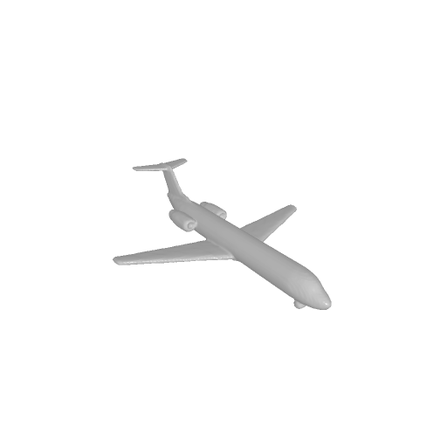}%
\includegraphics[width=0.155\textwidth]{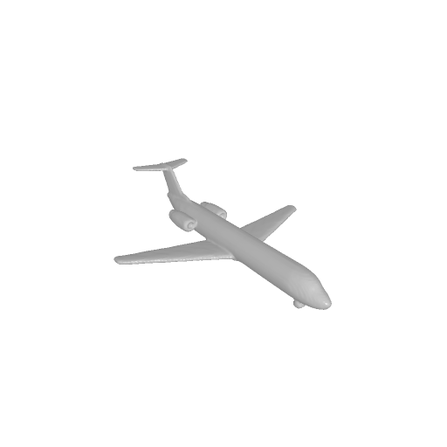}%
}\\[0.55em]
    {\small\textbf{StyleFields}}\\[-0.15em]
\makebox[\textwidth][c]{%
\includegraphics[width=0.155\textwidth]{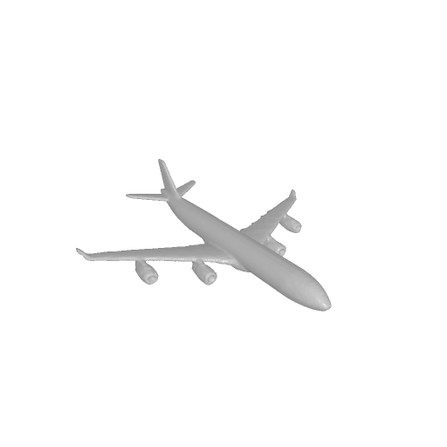}%
\includegraphics[width=0.155\textwidth]{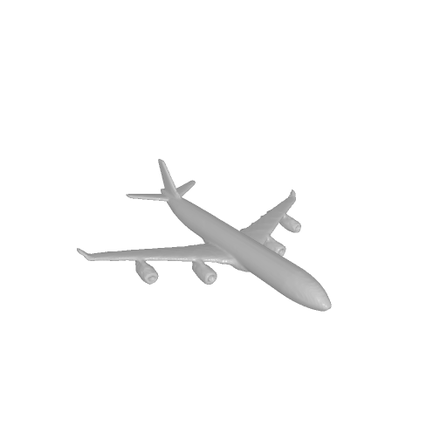}%
\includegraphics[width=0.155\textwidth]{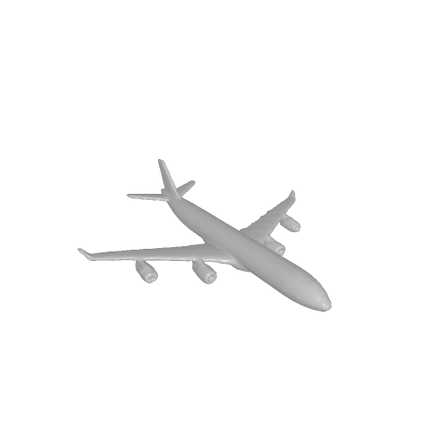}%
\includegraphics[width=0.155\textwidth]{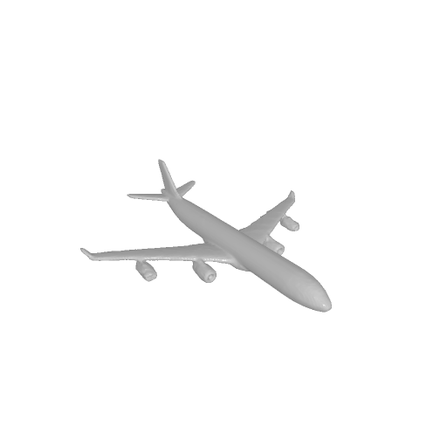}%
\includegraphics[width=0.155\textwidth]{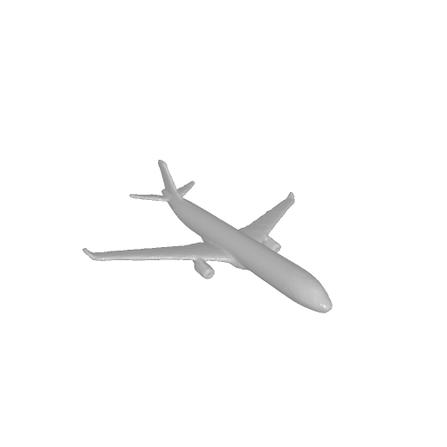}%
\includegraphics[width=0.155\textwidth]{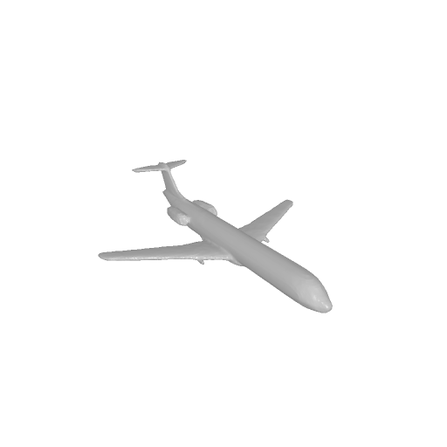}%
}\\[-0.15em]
\makebox[\textwidth][c]{%
\includegraphics[width=0.155\textwidth]{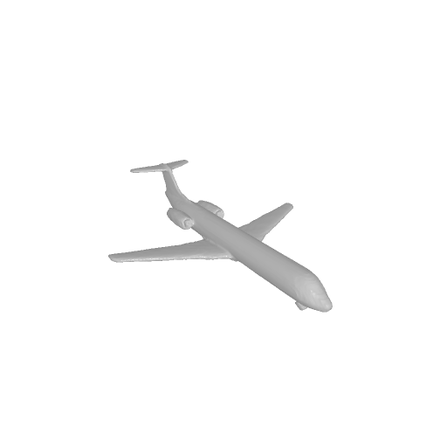}%
\includegraphics[width=0.155\textwidth]{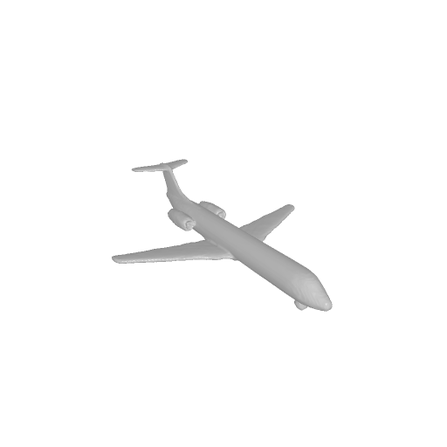}%
\includegraphics[width=0.155\textwidth]{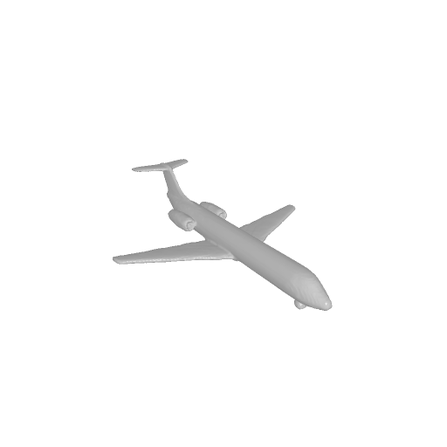}%
\includegraphics[width=0.155\textwidth]{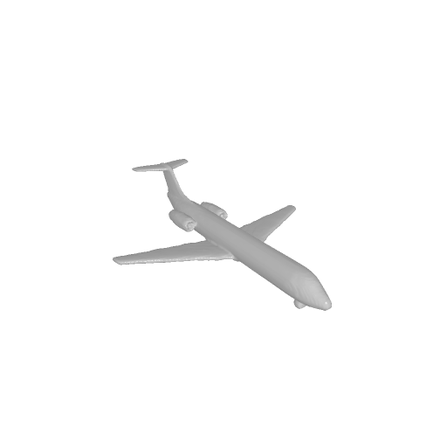}%
\includegraphics[width=0.155\textwidth]{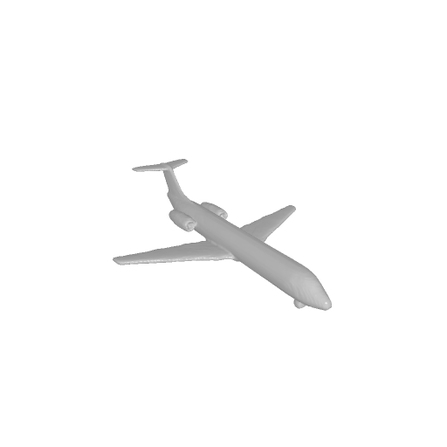}%
}\\[0.55em]
    \caption{Dense global-latent interpolation for planes. The trajectories are visually smoother than in the chair case, but the same entanglement remains: coarse airplane structure and finer geometric details change together along the interpolation path, rather than being controlled independently by scale.}
    \label{fig:supp_dense_interp_planes}
\end{figure*}

We complement the global-latent interpolation results of the main paper with denser trajectories between the same endpoint shapes. Specifically, for each pair, we linearly interpolate the global latent code using
\[
z_{\mathrm{interp}}(t) = (1-t)\, z_1 + t\, z_2,
\qquad
t \in \{0, 0.1, 0.2, \dots, 0.9, 1.0\}.
\]
We visualize these denser interpolation paths for DeepSDF, Curriculum DeepSDF, Curriculum Style DeepSDF, and \ours on Cars, Chairs, and Planes in~\cref{fig:supp_dense_interp_cars1,fig:supp_dense_interp_cars2,fig:supp_dense_interp_chairs,fig:supp_dense_interp_planes}.

The purpose of this supplementary analysis is not to argue that interpolation in the single global latent \(z\) yields disentangled control. On the contrary, these denser sequences make the entanglement more apparent. As stated in the main paper, interpolating a single global latent code causes coarse structure and fine detail to change simultaneously, rather than independently. As the interpolation factor increases, large-scale attributes such as overall silhouette, proportions, and major structural components evolve together with finer-scale geometric details, so the intermediate shapes exhibit averaged geometry rather than a clean preservation of one level of detail with selective transfer of another. 

This behavior is consistently visible in the car and plane examples. In both categories, the intermediate shapes do not preserve the coarse form of one endpoint while borrowing only local details from the other. Instead, low-frequency and high-frequency characteristics transition jointly along the path, confirming that global-latent interpolation remains an entangled operation. This observation holds for all compared methods, including \ours when interpolation is performed solely in \(z\). 

At the same time, the denser chair examples reveal an additional qualitative difference in interpolation quality. Although StyleFields still does not disentangle coarse and fine geometry under global-latent interpolation, it produces substantially cleaner and more structurally plausible intermediate chairs than the baselines. In particular, the transition remains more coherent across seat, backrest, and leg structure, whereas DeepSDF and Curriculum DeepSDF exhibit stronger geometric degradation and topological instability in the intermediate steps, and Curriculum Style DeepSDF, while more stable, is still less consistent than \ours. Thus, for Chairs, the supplementary results show two facts simultaneously: interpolation in \(z\) remains entangled, but the geometry generated by \ours along that entangled path is markedly more stable and realistic.

For Planes, the trajectories are visually smoother across all methods than in Chairs, but they still support the same conclusion regarding entanglement. Changes in wing configuration, engines, and finer structural cues occur together along the interpolation path, rather than being isolated by scale. Hence, denser interpolation reinforces the central point of the main paper: independent coarse-to-fine control is not achieved by interpolating the single global latent \(z\), but instead by selectively manipulating the depth-wise style codes in \(w^+\). 

\section{Ablation Studies}

We focus the supplementary ablation on the most central architectural variable introduced by StyleFields: the number of depth-wise style groups in \(W^+\). Since the main paper already compares against external baselines, the goal here is to isolate whether the gains of StyleFields arise from the proposed multi-style factorization itself, rather than from a generic increase in decoder capacity. To this end, Table~\ref{tab:supp_ablation_wplus_groups} varies the number of modulation groups \(|\mathcal J|\), where each style code \(w_j^+\) is assigned to a different depth range of the decoder, and reports the corresponding total number of learnable parameters.

\paragraph{Effect of the number of \texorpdfstring{$w^+$}{w+} groups.}
The quantitative trends show that the number of style groups matters substantially. When \(|\mathcal J|=1\), the model is effectively reduced to a single-style variant, and the results remain close to the single-code regime, especially on Cars and Planes. Increasing the number of groups from \(1\) to \(4\) improves reconstruction quality consistently, with the clearest gains appearing on Chairs and Planes. Cars also improve, although the margin is smaller, indicating that the benefit of the multi-style decomposition is present across all categories but is especially pronounced for shapes with larger structural variation.

The parameter counts show that these gains are not explained simply by uncontrolled model growth. As \(|\mathcal J|\) increases from \(1\) to \(5\), the total parameter count rises only moderately from \(3.2\)M to \(4.5\)M, yet the reconstruction improvements are substantial up to \(|\mathcal J|=4\). Moreover, increasing the number of groups beyond this point does not yield further benefit. Compared with \(|\mathcal J|=4\), the \(|\mathcal J|=5\) model is worse on all three categories in both CD and IoU. This indicates that performance saturates at four groups, suggesting that a compact multi-style factorization is sufficient to capture the intended coarse-to-fine decomposition.

An additional observation is that the inference-time parameterization matters. For the main configuration \((|\mathcal J|=4)\), reconstructing with \(W^+\) instead of a single global \(z\) further improves performance on all three categories without changing the model size. This confirms that the benefit of \ours is not only architectural, but also tied to the structured multi-code optimization space exposed by the proposed formulation.

Overall, Table~\ref{tab:supp_ablation_wplus_groups} supports three conclusions. First, the gains of \ours are not due merely to decoder growth: reconstruction quality improves markedly as the latent control is partitioned into a small number of depth-wise style groups. Second, a compact decomposition is sufficient in practice, with \(|\mathcal J|=4\) providing the best overall trade-off between flexibility and stability. Third, the parameter overhead remains efficient. Relative to Curriculum DeepSDF, the main \ours model roughly doubles the parameter count (\(2.1\)M \(\rightarrow\) \(4.2\)M), but this increase yields not only stronger reconstruction performance, but also the structured \(w^+\) parameterization that enables cross-instance style mixing and scale-selective control. In this sense, the added parameters do more than improve accuracy alone: they provide the representational structure required for the manipulation capabilities that are central to \ours.

\begin{table}[t]
\centering
\caption{Ablation on the number of depth-wise style groups in \(\mathcal J\) across Cars, Chairs, and Planes. Here, \(|\mathcal J|\) denotes the number of modulation groups used to partition the decoder into coarse-to-fine style-controlled ranges. We also report the corresponding number of learnable style parameters introduced by each setting. Lower CD is better and higher IoU is better.}
\label{tab:supp_ablation_wplus_groups}
\resizebox{\linewidth}{!}{%
\begin{tabular}{l c cc cc cc}
\toprule
& & \multicolumn{2}{c}{Cars} & \multicolumn{2}{c}{Chairs} & \multicolumn{2}{c}{Planes} \\
\cmidrule(lr){3-4} \cmidrule(lr){5-6} \cmidrule(lr){7-8}
Method & \# Params & CD (\(\downarrow\)) & IoU (\(\uparrow\)) & CD (\(\downarrow\)) & IoU (\(\uparrow\)) & CD (\(\downarrow\)) & IoU (\(\uparrow\)) \\
\midrule
DeepSDF  & 1.7M    & \texttt{$1.6\times10^{-4}$} & 97.27 & \texttt{$8.8\times10^{-4}$} & 89.74 & \texttt{$8.0\times10^{-4}$} & 89.28 \\
Curriculum DeepSDF & 2.1M & \texttt{$1.6\times10^{-4}$} & 97.19 & \texttt{$1.2\times10^{-3}$} & 89.33 & \texttt{$9.2\times10^{-4}$} & 90.57 \\
Curriculum + Style DeepSDF Reconstruct with $Z$ and $|\mathcal J|=4$  & 4.2M & \texttt{$1.3\times10^{-4}$} & 97.87 & \texttt{$5.7\times10^{-4}$} & 90.12 & \texttt{$7.8\times10^{-4}$} & 91.26 \\
\ours: Reconstruct with $Z$ with $|\mathcal J|=1$ & 3.2M & $1.6\times10^{-4}$ & 97.25 & $8.6\times10^{-4}$ & 90.05 & $8.1\times10^{-4}$ & 89.15 \\
\ours: Reconstruct with $Z$ with $|\mathcal J|=2$ & 3.6M & $1.5\times10^{-4}$ & 97.27 & $6.4\times10^{-4}$ & 91.17 & $7.9\times10^{-4}$ & 91.32 \\
\ours: Reconstruct with $Z$ with $|\mathcal J|=3$ & 3.9M & $1.5\times10^{-4}$ & 97.34 & $2.9\times10^{-4}$ & 93.76 & $7.2\times10^{-4}$ & 92.46 \\
\ours: Reconstruct with $Z$ with $|\mathcal J|=4$ & 4.2M & $1.3\times10^{-4}$ & 97.36 & $1.8\times10^{-4}$ & 94.39 & $6.6\times10^{-4}$ & 92.61 \\
\ours: Reconstruct with $W+$ with $|\mathcal J|=4$   & 4.2M  & \texttt{$\boldsymbol{1.2\times10^{-4}}$} & \textbf{98.43} & \texttt{$\boldsymbol{1.6\times10^{-4}}$} & \textbf{95.31} & \texttt{$\boldsymbol{6.4\times10^{-4}}$} & \textbf{93.66} \\
\ours: Reconstruct with $Z$ with $|\mathcal J|=5$ & 4.5M & $1.4\times10^{-4}$ & 97.34 & $1.9\times10^{-4}$ & 94.09 & $6.8\times10^{-4}$ & 92.55 \\
\bottomrule
\end{tabular}%
}
\end{table}

\section{Style Mixing}

We provide additional cross-instance style-mixing examples for Chairs and Planes to complement the car results shown in the main paper. As in the main paper, we construct hybrid shapes by selectively substituting subsets of the depth-wise style codes of Shape~1 with the corresponding codes from Shape~2. This experiment directly tests whether the representation organizes geometry across depth in a manner that supports selective exchange of coarse and fine attributes across instances. 

\begin{figure}[t]
    \centering
    \setlength{\tabcolsep}{2pt}
    \renewcommand{\arraystretch}{1.0}

    \resizebox{0.8\linewidth}{!}{%
    \begin{tabular}{ccccc}
        {\scriptsize Shape 1} &
        {\scriptsize A} &
        {\scriptsize B} &
        {\scriptsize C} &
        {\scriptsize Shape 2} \\[4pt]

        \includegraphics[width=0.18\linewidth]{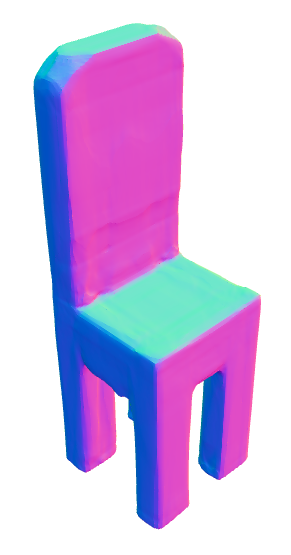} &
        \includegraphics[width=0.18\linewidth]{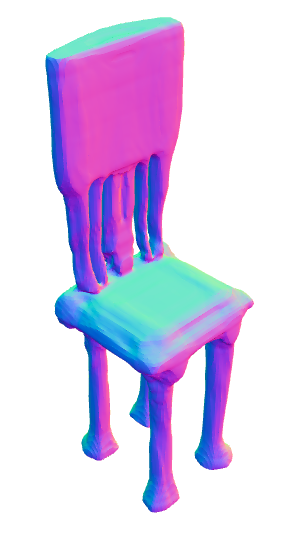} &
        \includegraphics[width=0.18\linewidth]{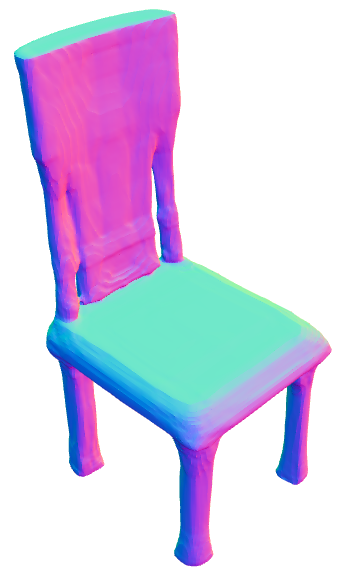} &
        \includegraphics[width=0.18\linewidth]{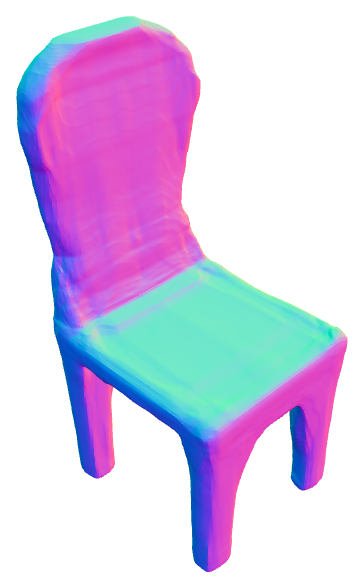} &
        \includegraphics[width=0.18\linewidth]{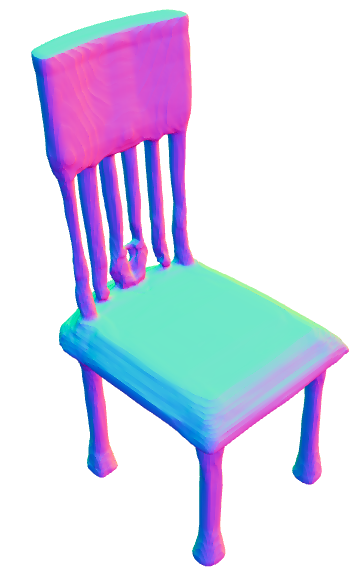} \\

        \includegraphics[width=0.18\linewidth]{figures/style_mixing/chairs_1/0} &
        \includegraphics[width=0.18\linewidth]{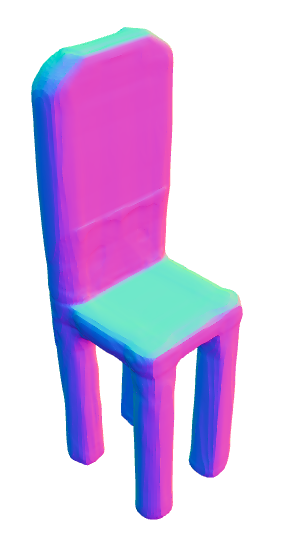} &
        \includegraphics[width=0.18\linewidth]{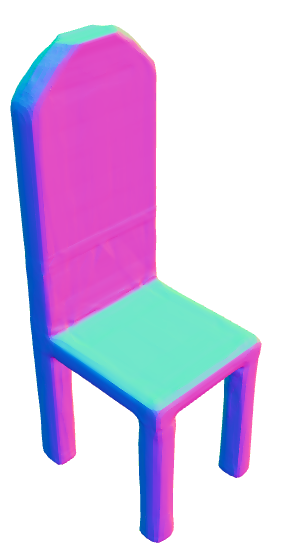} &
        \includegraphics[width=0.18\linewidth]{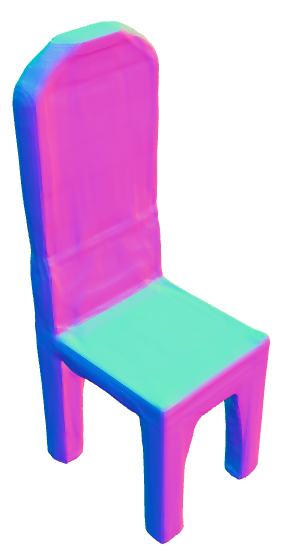} &
        \includegraphics[width=0.18\linewidth]{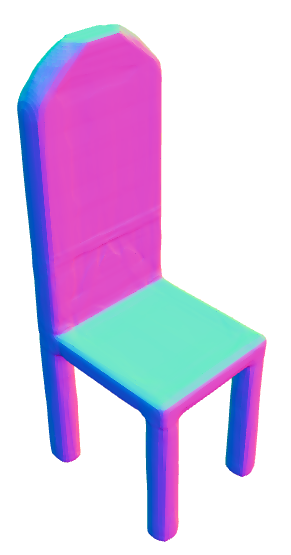} \\

        \includegraphics[width=0.18\linewidth]{figures/style_mixing/chairs_1/0} &
        \includegraphics[width=0.18\linewidth]{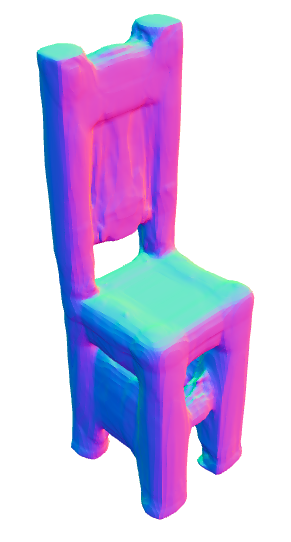} &
        \includegraphics[width=0.18\linewidth]{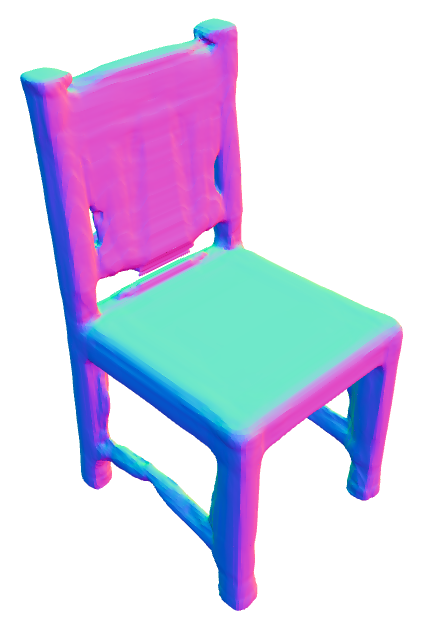} &
        \includegraphics[width=0.18\linewidth]{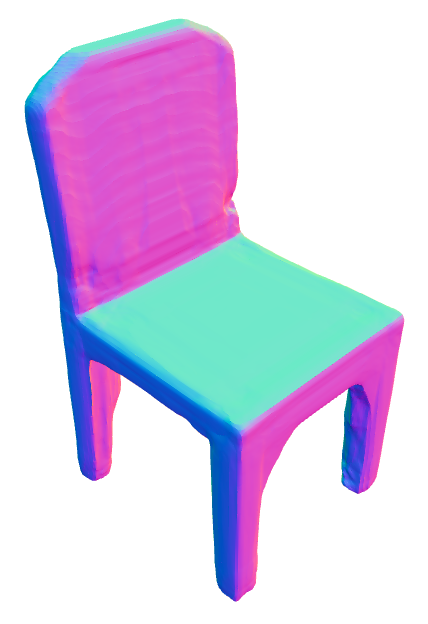} &
        \includegraphics[width=0.18\linewidth]{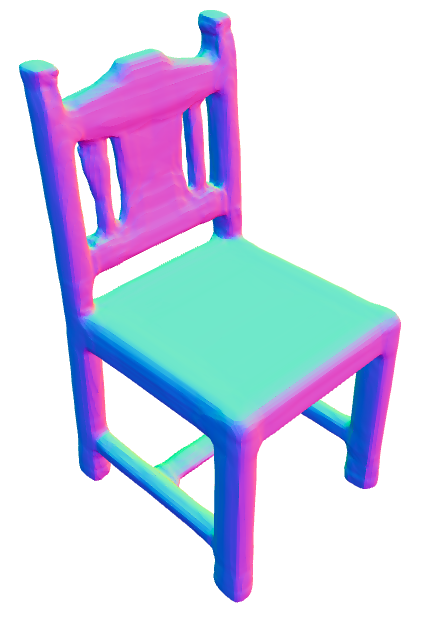} \\

        \includegraphics[width=0.18\linewidth]{figures/style_mixing/chairs_1/0} &
        \includegraphics[width=0.18\linewidth]{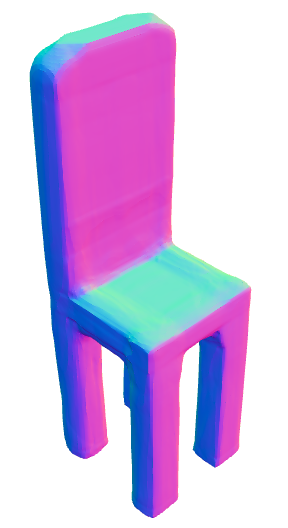} &
        \includegraphics[width=0.18\linewidth]{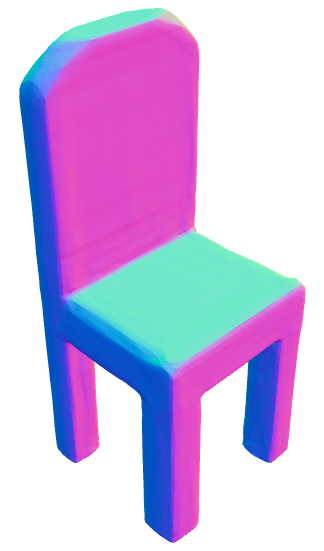} &
        \includegraphics[width=0.18\linewidth]{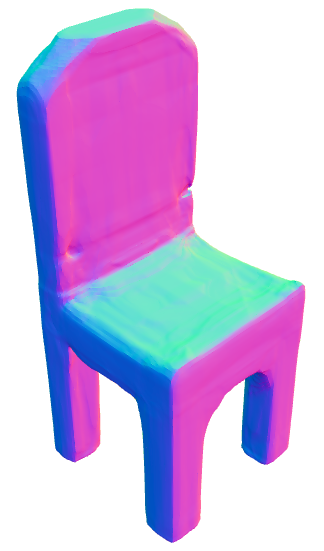} &
        \includegraphics[width=0.18\linewidth]{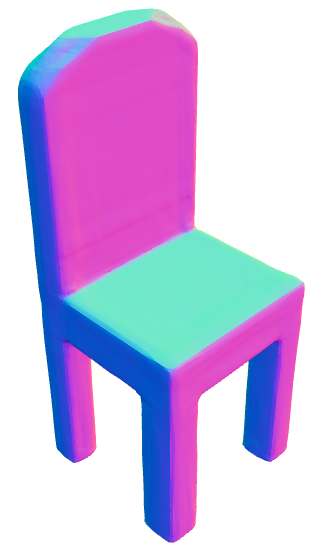} \\

    \end{tabular}%
    }

    \caption{Additional StyleFields cross-instance style mixing between two chair shapes (source on the left, target on the right). Each column shows a hybrid produced by selectively substituting style codes of Shape~1 with codes from Shape~2. (A) Coarse styles \((w^+_0, w^+_1)\) from Shape~1 and fine styles \((w^+_2, w^+_3)\) from Shape~2. (B) Mid-level styles \((w^+_1, w^+_2)\) from Shape~1 and coarsest and finest styles \((w^+_0, w^+_3)\) from Shape~2. (C) Fine styles \((w^+_2, w^+_3)\) from Shape~1 and coarse styles \((w^+_0, w^+_1)\) from Shape~2.}
    \label{fig:style_mixing_chair1}
\end{figure}

\begin{figure}[t]
    \centering
    \setlength{\tabcolsep}{2pt}
    \renewcommand{\arraystretch}{1.0}

    \resizebox{0.8\linewidth}{!}{%
    \begin{tabular}{ccccc}
        {\scriptsize Shape 1} &
        {\scriptsize A} &
        {\scriptsize B} &
        {\scriptsize C} &
        {\scriptsize Shape 2} \\[4pt]
        \includegraphics[width=0.18\linewidth]{figures/style_mixing/chairs_1/0} &
        \includegraphics[width=0.18\linewidth]{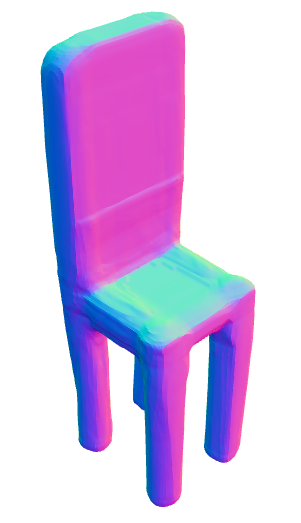} &
        \includegraphics[width=0.18\linewidth]{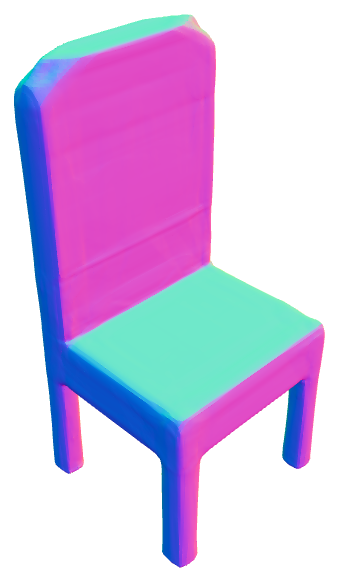} &
        \includegraphics[width=0.18\linewidth]{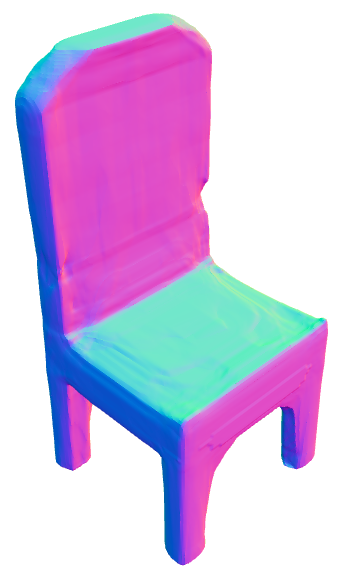} &
        \includegraphics[width=0.18\linewidth]{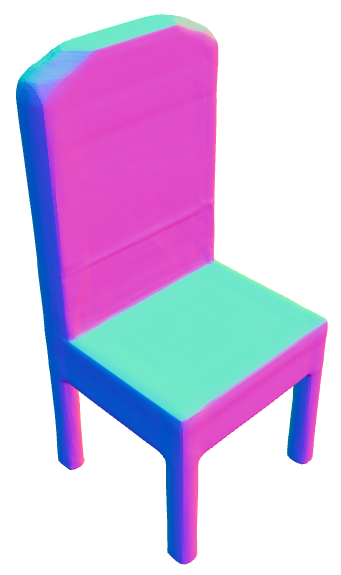} \\

        \includegraphics[width=0.18\linewidth]{figures/style_mixing/chairs_1/0} &
        \includegraphics[width=0.18\linewidth]{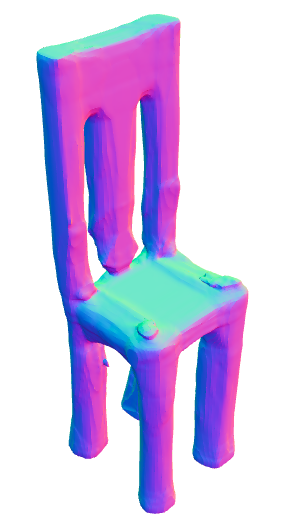} &
        \includegraphics[width=0.18\linewidth]{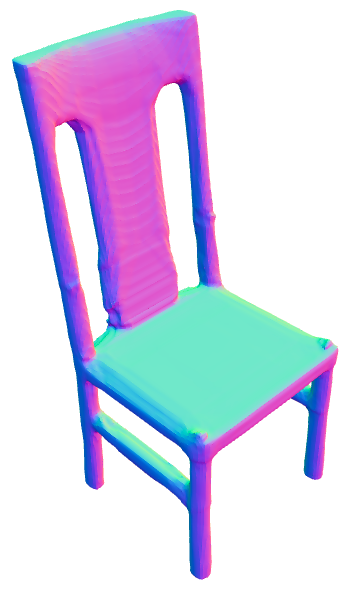} &
        \includegraphics[width=0.18\linewidth]{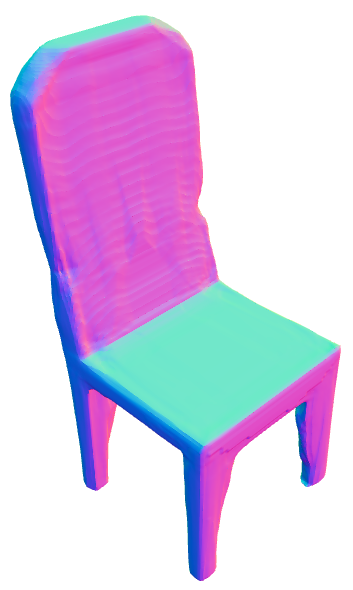} &
        \includegraphics[width=0.18\linewidth]{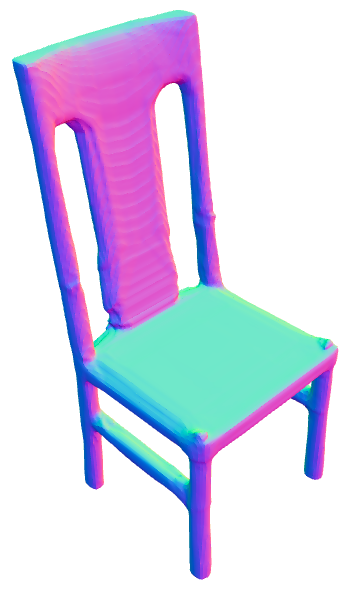} \\

        \includegraphics[width=0.18\linewidth]{figures/style_mixing/chairs_1/0} &
        \includegraphics[width=0.18\linewidth]{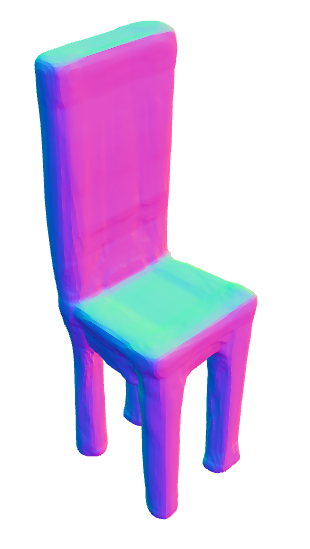} &
        \includegraphics[width=0.18\linewidth]{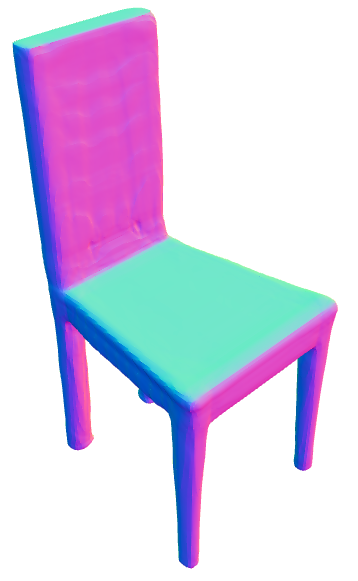} &
        \includegraphics[width=0.18\linewidth]{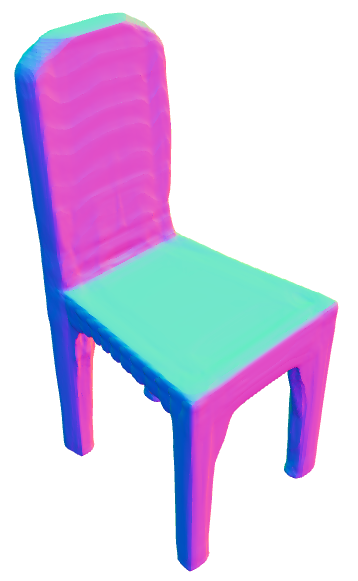} &
        \includegraphics[width=0.18\linewidth]{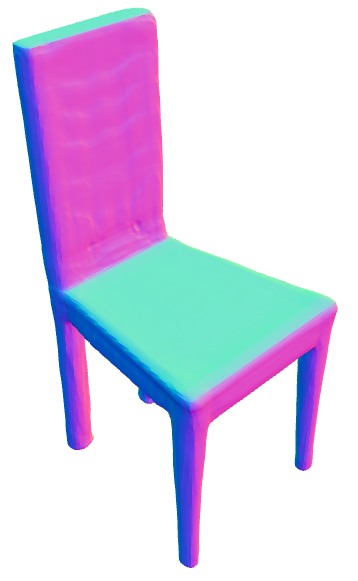} \\

        \includegraphics[width=0.18\linewidth]{figures/style_mixing/chairs_1/0} &
        \includegraphics[width=0.18\linewidth]{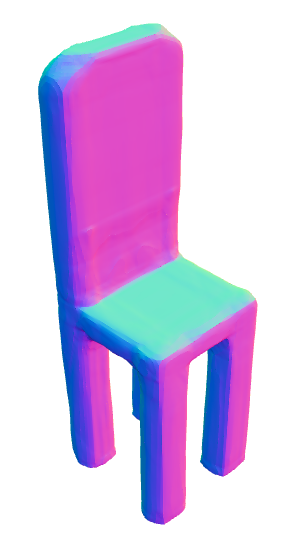} &
        \includegraphics[width=0.18\linewidth]{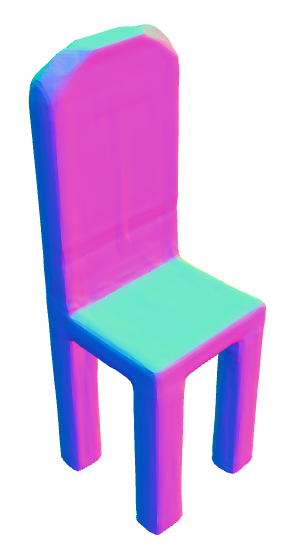} &
        \includegraphics[width=0.18\linewidth]{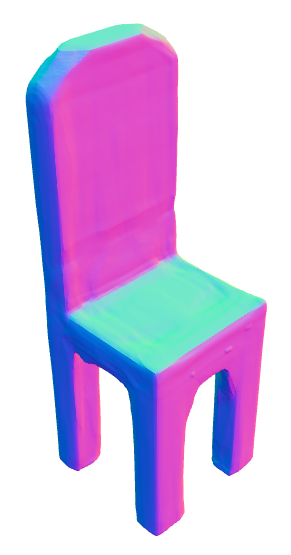} &
        \includegraphics[width=0.18\linewidth]{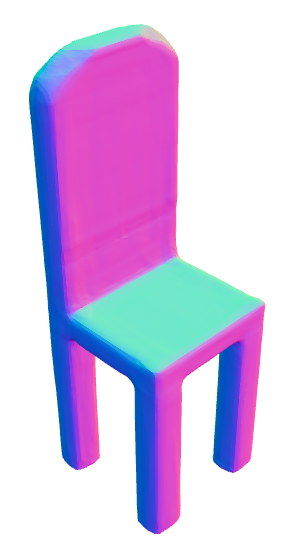} \\

    \end{tabular}%
    }

    \caption{More cross-instance style mixing examples on Chairs. The same substitution patterns as in Fig.~\ref{fig:style_mixing_chair1} are used. Replacing shallow style groups leads to larger changes in global chair structure, while preserving them retains more of the source shape’s broad configuration and yields more localized modifications.}
    \label{fig:style_mixing_chair2}
\end{figure}

\begin{figure}[t]
    \centering
    \setlength{\tabcolsep}{2pt}
    \renewcommand{\arraystretch}{1.0}

    \resizebox{0.8\linewidth}{!}{%
    \begin{tabular}{ccccc}
        {\scriptsize Shape 1} &
        {\scriptsize A} &
        {\scriptsize B} &
        {\scriptsize C} &
        {\scriptsize Shape 2} \\[4pt]

        \includegraphics[width=0.18\linewidth]{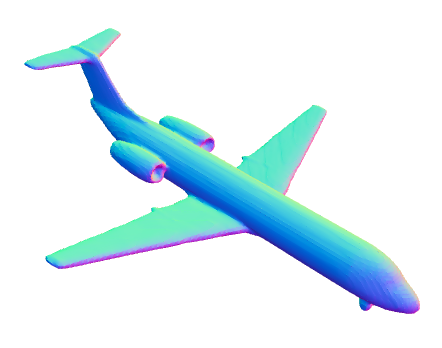} &
        \includegraphics[width=0.18\linewidth]{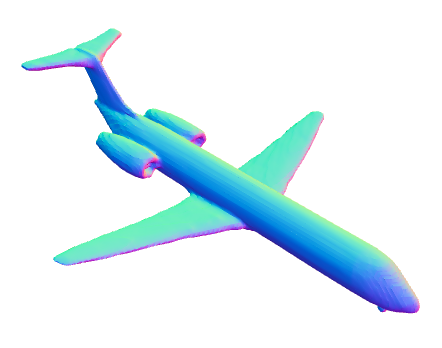} &
        \includegraphics[width=0.18\linewidth]{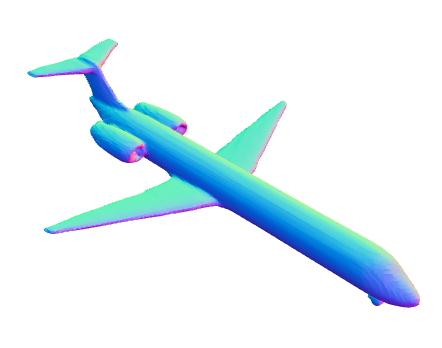} &
        \includegraphics[width=0.18\linewidth]{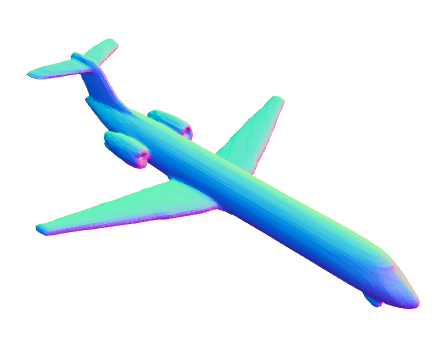} &
        \includegraphics[width=0.18\linewidth]{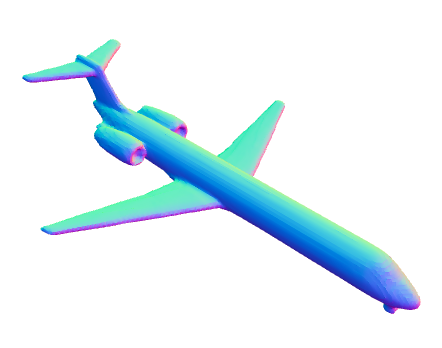} \\

        \includegraphics[width=0.18\linewidth]{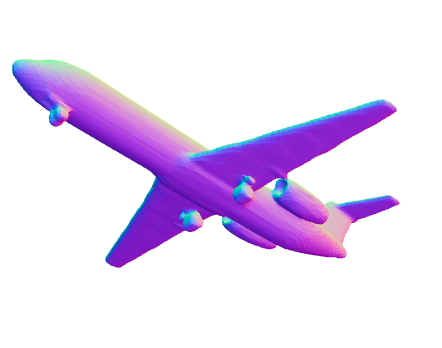} &
        \includegraphics[width=0.18\linewidth]{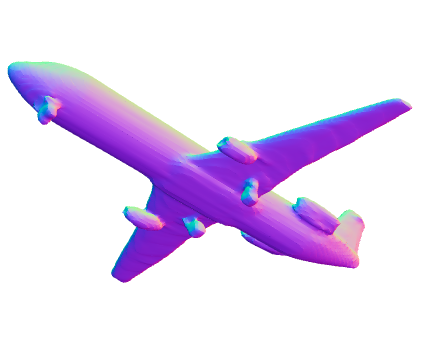} &
        \includegraphics[width=0.18\linewidth]{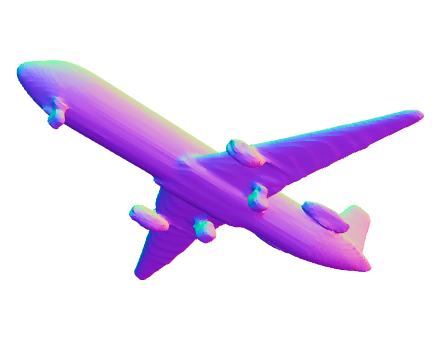} &
        \includegraphics[width=0.18\linewidth]{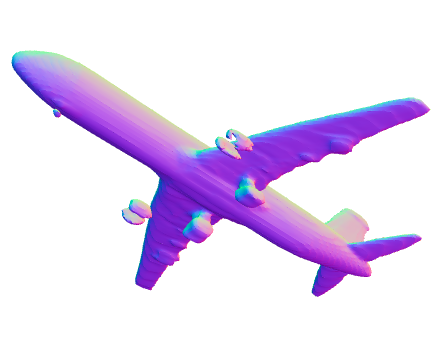} &
        \includegraphics[width=0.18\linewidth]{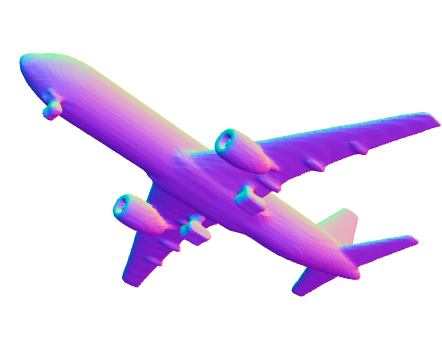} \\

        \includegraphics[width=0.18\linewidth]{figures/style_mixing/planes/0} &
        \includegraphics[width=0.18\linewidth]{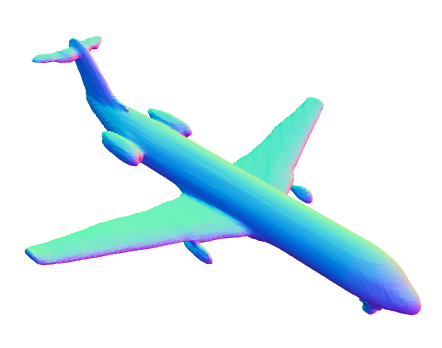} &
        \includegraphics[width=0.18\linewidth]{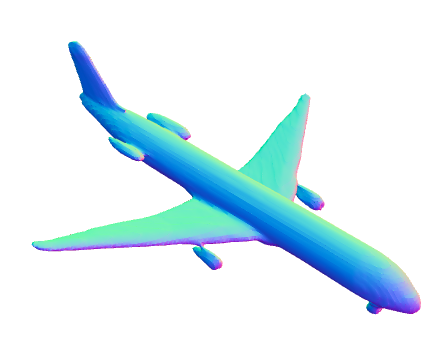} &
        \includegraphics[width=0.18\linewidth]{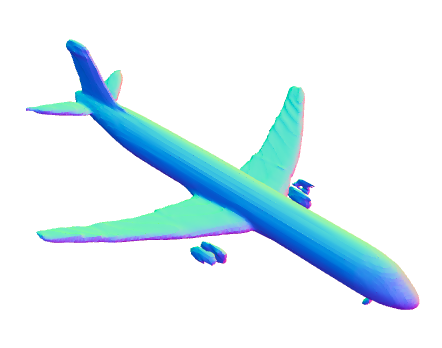} &
        \includegraphics[width=0.18\linewidth]{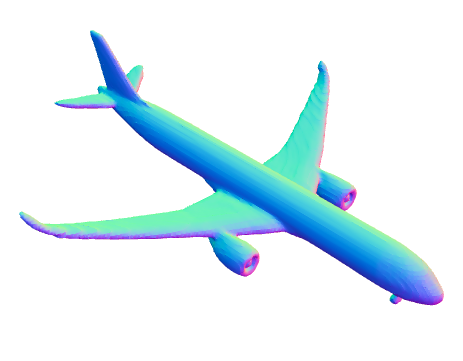} \\

        \includegraphics[width=0.18\linewidth]{figures/style_mixing/planes/0} &
        \includegraphics[width=0.18\linewidth]{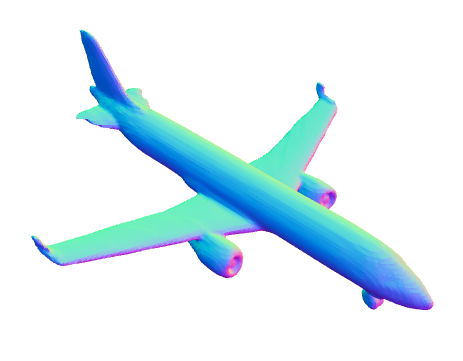} &
        \includegraphics[width=0.18\linewidth]{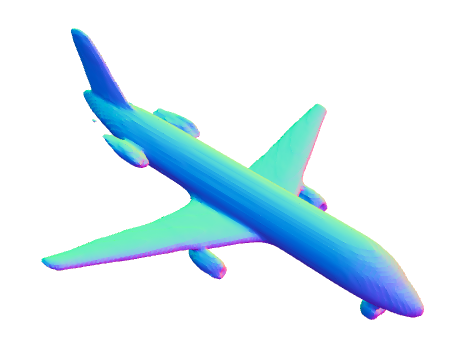} &
        \includegraphics[width=0.18\linewidth]{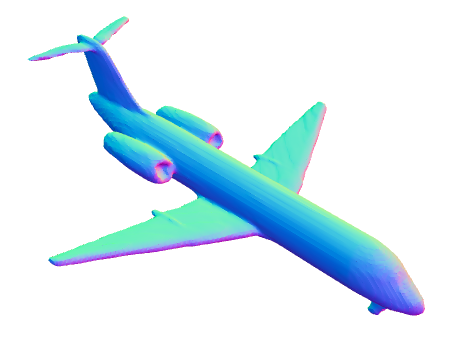} &
        \includegraphics[width=0.18\linewidth]{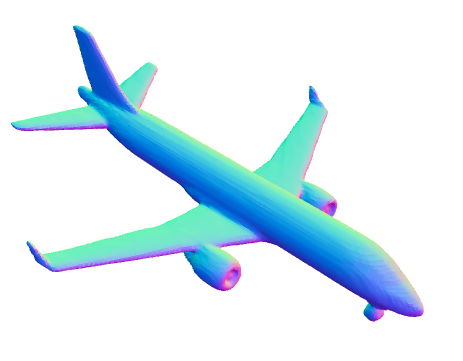} \\

    \end{tabular}%
    }

    \caption{Additional StyleFields cross-instance style mixing between two airplane shapes (source on the left, target on the right). Each column shows a hybrid produced by selectively substituting style codes of Shape~1 with codes from Shape~2. (A) Coarse styles \((w^+_0, w^+_1)\) from Shape~1 and fine styles \((w^+_2, w^+_3)\) from Shape~2. (B) Mid-level styles \((w^+_1, w^+_2)\) from Shape~1 and coarsest and finest styles \((w^+_0, w^+_3)\) from Shape~2. (C) Fine styles \((w^+_2, w^+_3)\) from Shape~1 and coarse styles \((w^+_0, w^+_1)\) from Shape~2.}
    \label{fig:style_mixing_plane1}
\end{figure}

\cref{fig:style_mixing_chair1,fig:style_mixing_chair2,fig:style_mixing_plane1} show the style mixing for Chairs and Planes. Each row corresponds to a different source--target pair, and each hybrid column uses the same substitution pattern as in the car experiment. Column~A keeps the coarse styles \((w_0^+, w_1^+)\) from Shape~1 and replaces the fine styles \((w_2^+, w_3^+)\) with those of Shape~2. Column~B keeps the mid-level styles \((w_1^+, w_2^+)\) from Shape~1 and replaces the coarsest and finest styles \((w_0^+, w_3^+)\) with those of Shape~2. Column~C keeps the fine styles \((w_2^+, w_3^+)\) from Shape~1 and replaces the coarse styles \((w_0^+, w_1^+)\) with those of Shape~2. This is the same style-mixing protocol used in the main paper for cars. 

The supplementary examples are consistent with the qualitative behavior reported in the main paper. For Chairs, substituting coarse styles produces pronounced changes in global structure, including the overall backrest profile, seat proportions, leg configuration, and the presence or absence of major structural elements such as slats or rounded backs. By contrast, when coarse styles are preserved and only deeper styles are taken from the other shape, the hybrids retain much more of the source chair’s broad configuration while exhibiting more localized changes. For Planes, the same coarse-to-fine pattern appears in terms of fuselage profile, wing sweep, tail configuration, and engine placement: replacing shallow style groups changes the overall airframe more strongly, whereas preserving them tends to maintain the global aircraft form more faithfully. These results support the interpretation that the depth-wise style codes expose a practically useful geometric organization across scales rather than a single entangled latent space. 

At the same time, the hybrids are not expected to decompose geometry into perfectly isolated factors. Instead, the examples demonstrate the more realistic and more relevant property claimed in the paper: \ours enables structured, scale-selective manipulation in which coarse attributes can be transferred across instances without fully destroying local character, and fine-level substitutions can modify shape appearance without forcing a complete change in global form. The chair and plane examples therefore extend the main paper’s car results and show that the same behavior generalizes across categories with very different geometric statistics. 

\section{Implementation Details}

For the multi-scale supervision used in \ours, we employ dataset-specific smoothing schedules for the auxiliary SDF targets. For Cars, we use Gaussian smoothing levels \(\sigma \in \{12, 5, 2, 1\}\). For Chairs, we use \(\sigma \in \{5, 3, 2, 1\}\). For Planes, we use \(\sigma \in \{2, 1.5, 1, 0.5\}\). These schedules are used to provide progressively sharper geometric supervision across decoder depth, consistent with the coarse-to-fine training strategy of the proposed model.

\section{Drag Optimization}

We provide additional details for the drag-optimization pipeline used in the main paper. Starting from a reconstructed car, we optimize the latent variables of the pretrained shape model while keeping all network weights fixed. The optimized variables are either: (i) the single global latent \(z\), (ii) only the early style codes \((w_0^+, w_1^+)\), or (iii) only the late style codes \((w_2^+, w_3^+)\). All non-optimized latent variables are kept fixed at their initial values throughout the optimization. The decoded shape is represented as an SDF and is converted at each iteration into a differentiable surface mesh using MeshSDF, so that gradients from the aerodynamic objective can flow back through the surface-extraction stage to the optimized latent variables. 

The aerodynamic objective is evaluated by a learned mesh-based surrogate. Concretely, the current mesh is passed through a graph convolutional network with GraphSAGE layers, which predicts the surface pressure at mesh vertices. The predicted pressure field is then integrated over the car surface to compute the pressure drag. Following the standard drag formulation, if \(p(x)\) denotes the predicted pressure and \(n(x)\) the outward surface normal at point \(x\), the pressure drag is obtained from the surface integral of the pressure component aligned with the longitudinal axis of the car, and is then normalized to produce the drag coefficient \(C_d\). In our setup we optimize this pressure-based drag objective directly, which provides a differentiable approximation of aerodynamic performance while avoiding repeated CFD solves inside the optimization loop. 

The full optimization objective is
\[
\mathcal{L}_{\mathrm{drag}} = C_d\!\big(\mathcal{M}(f_\theta)\big) + \mathcal{L}_{\mathrm{reg}},
\]
where \(\mathcal{M}(f_\theta)\) denotes the differentiably extracted mesh from the current SDF and \(\mathcal{L}_{\mathrm{reg}}\) keeps the edited shape close to the original reconstruction. In our implementation, this regularizer contains two components. First, we use an \(L_2\) penalty on the deviation of the optimized latent variables from their initial values, which discourages large latent drift:
\[
\mathcal{L}_{\mathrm{init}} = \|\psi - \psi_{\mathrm{init}}\|_2^2,
\]
where \(\psi\) denotes the variables being optimized in the current regime (\(z\), early \(w^+\), or late \(w^+\)). Second, we use a \(k\)-nearest-neighbor regularization term in latent space, which encourages the optimized codes to remain close to the training distribution and reduces the chance of producing unrealistic off-manifold shapes. The regularizer is therefore
\[
\mathcal{L}_{\mathrm{reg}} = \lambda_{\mathrm{init}} \mathcal{L}_{\mathrm{init}} + \lambda_{\mathrm{knn}} \mathcal{L}_{\mathrm{knn}}.
\]
This combination stabilizes optimization and helps preserve plausible geometry while still allowing meaningful drag-reducing edits. 

Each iteration of the optimization proceeds as follows. First, the current latent variables are decoded into an SDF on a 3D grid. Second, a differentiable mesh is extracted from this SDF. Third, the mesh is passed through the GraphSAGE surrogate to obtain the predicted pressure field and the corresponding drag coefficient. Fourth, the gradient of \( \mathcal{L}_{\mathrm{drag}} \) is backpropagated through both the surrogate and the differentiable meshing stage to the optimized latent variables, which are then updated with gradient descent. Repeating this procedure yields a sequence of shapes with progressively reduced predicted drag. In the supplementary folder, we additionally provide GIFs of these optimization trajectories so that the temporal evolution of the edits can be inspected directly, beyond the static examples shown in the main paper. 

This expanded procedure also clarifies the qualitative behavior reported in the main paper. Optimizing the full global latent \(z\) gives the optimizer access to all geometric scales simultaneously, which is why it often produces larger, more holistic shape drift. By contrast, restricting optimization to the early style groups primarily affects broad low-frequency structure, while restricting it to the late style groups concentrates the updates on smaller, more localized surface changes. The drag-optimization results in the paper are therefore not only a downstream application, but also an additional validation that the learned \(w^+\) parameterization exposes practically useful control over different geometric scales.

\end{document}